\documentclass[preprint,authoryear,3p]{elsarticle}

\usepackage[english]{babel}
\usepackage[latin1]{inputenc}
\usepackage{natbib}
\usepackage{amssymb}
\usepackage{amsmath}  
\usepackage{color}    
\usepackage{moreverb} 
\usepackage{rotating}
\usepackage{tabularx}

\usepackage{pdfpages}
\usepackage{morefloats}
\usepackage{graphics}
\usepackage{bbm}
\RequirePackage[latin1]{inputenc}
\usepackage{epsfig}
\usepackage{mathrsfs}
\usepackage{multirow}
\usepackage{amsthm}
\usepackage{graphicx} 

\newtheorem{df}{Definition}
\newtheorem{thm}{Theorem}

\newproof{pf}{Proof}

\usepackage{latexsym}

\usepackage{bold-extra}
\usepackage{stmaryrd}

\newcommand{\abs}[1]{\mid #1 \mid}
\newcommand{\sentences}[1]{\left\llbracket #1 \right\rrbracket}

\newcommand{\absd}[1]{\left\Vert #1 \right\Vert}

\newcommand{\dinfty}[0]{\pmb{\pmb{\infty}}}

\journal{Journal of Computer and System Sciences}
\begin{document}

\begin{frontmatter}

\title{Absolute convergence and error thresholds in non-active adaptive
  sampling}

\author[UVigo]{Manuel Vilares Ferro\corref{cor}}
\cortext[cor]{Corresponding author: tel. +34 988 387280, fax +34 988 387001.}
\ead{vilares@uvigo.es}
\author[UVigo]{V\'{\i}ctor M. Darriba Bilbao}
\ead{darriba@uvigo.es}
\author[UDC]{Jes\'us Vilares Ferro}
\ead{jvilares@udc.es}

\address[UVigo]{Department of Computer Science, University of Vigo \\ Campus As Lagoas s/n, 32004 -- Ourense, Spain}
\address[UDC]{Department of Computer Science, University of A Coru\~na \\ Campus de Elvi\~na, 15071 -- A Coru\~na, Spain}

\begin{abstract}
  Non-active adaptive sampling is a way of building machine learning
  models from a training data base which are supposed to dynamically
  and automatically derive guaranteed sample size. In this context and
  regardless of the strategy used in both scheduling and generating of
  weak predictors, a proposal for calculating absolute convergence and
  error thresholds is described. We not only make it possible to
  establish when the quality of the model no longer increases, but
  also supplies a proximity condition to estimate in absolute terms
  how close it is to achieving such a goal, thus supporting decision
  making for fine-tuning learning parameters in model selection.
  The technique proves its correctness and completeness with respect
  to our working hypotheses, in addition to strengthening the
  robustness of the sampling scheme. Tests meet our expectations and
  illustrate the proposal in the domain of natural language
  processing, taking the generation of part-of-speech taggers as case
  study.
\end{abstract}


\end{frontmatter}

\section{Introduction}
\label{section-introduction}

A recurrent issue in \textit{machine learning} ({\sc ml}) relates the
determination of optimal sampling data sets, the aim being to reduce
both training costs and time without making the modelling process less
reliable. In this sense, the operating principle for adaptive sampling
is simple and involves beginning with an initial number of examples
and then iteratively learning the model, evaluating it and acquiring
additional observations if necessary. Accordingly, there are two
questions to be considered: it is necessary to determine the training
data to be acquired at each cycle, and also to define a halting
condition to terminate the loop once a certain degree of performance
has been achieved by the learner. Both tasks confer the character of
research issues to the formalization of \textit{scheduling} and
\textit{stopping criteria}~\citep{John96staticversus}, respectively.
The former has been researched for decades in terms of
fixed~\citep{John96staticversus,Provost:1999:EPS:312129.312188} or
adaptive~\citep{Provost:1999:EPS:312129.312188} sequencing, and it is
not our objective. As regards the halting criteria, they are
independent of the scheduling and mostly start from the hypothesis
that learning curves are well-behaved, including an initial steeply
sloping portion, a more gently sloping middle one and a final balanced
zone~\citep{Meek:2002:LSM:944790.944798}. Accordingly, the purpose is
to identify the moment in which such a curve reaches a plateau, namely
when adding more data instances does not improve the accuracy,
although this often does not strictly verify. Instead, extra learning
efforts almost always result in modest increases. This justifies the
interest in having a \textit{proximity condition}, understood as a
measure of the degree of convergence attained from a given iteration,
rather than a stopping one. In short, this will make it possible to
select the level of reliability in predicting a learner's performance,
both in terms of accuracy and computational costs. We will thus have a
powerful and flexible decision support tool in the field of model
selection, capable of adapting to the user's needs in terms of the
evaluation quality of both the learning strategy and its
parameterization.


A major challenge is then to avoid the overvaluation of learning
perturbations, in such a way that the training does not stop
prematurely due to temporary increases in accuracy. Namely, we are
interested in proving the \textit{correctness} of a proximity
criterion with respect to the working hypotheses, but also in
improving its capacity to mitigate the impact of such fluctuations
without compromising it, i.e. its \textit{robustness}. Given that we
are looking for a practical formula, it is finally necessary to ensure
its applicability, which relies on proving the \textit{completeness}
of the approach. These properties focus the attention of this work,
the structure of which we briefly describe. Firstly,
Section~\ref{section-state-of-the-art} examines the methodologies
serving as an inspiration to solve the question posed, as well as our
contributions. Next, Section~\ref{section-formal-framework} reviews
the mathematical basis supporting the proposal, whose model we
introduce in Section~\ref{section-abstract-model}. In
Section~\ref{section-testing-frame}, we describe the testing frame for
the experiments illustrated in
Section~\ref{section-experiments}. Finally,
Section~\ref{section-conclusions} presents the final conclusions.

\section{The state of the art}
\label{section-state-of-the-art}

Below is a brief review on how correctness, robustness and completeness
have been addressed over time in the definition of halting conditions
in adaptive sampling, thus allowing to contextualize our contribution
in that respect.

\subsection{Working on correctness, robustness and completeness}

Regarding correctness, most adaptive samplers assume a set of
hypotheses guaranteeing concurrence, such as access to independent and
identically distributed
observations~\citep{Schutze:2006:PTP:1183614.1183709,KATRIN08.335}. The
learning curve is then monotonic and, since it is bounded, training
converges on a supremum. At this point, the conditions for halting are
addressed from two viewpoints, depending on whether predictive
accuracy is the only factor to take into account~\citep{FreyFisher99}
or just another one in an optimization scenario stated in
\textit{decision theory}~\citep{Howard66}. In this latter context,
performance is understood as the search for a proper cost/benefit
trade-off and authors resort to statistically based strategies by
applying the principle of \textit{maximum expected
  utility}~\citep{Meek:2002:LSM:944790.944798} ({\sc meu}). This
implies taking all effectiveness considerations into account, which
depends on the degree of control exercised by the user on the learning
process. In its absence, namely using non-active techniques as we do,
the final cost is the sum of data acquisition, error and model
induction charges~\citep{WeissTian08}. Nonetheless, at best,
heuristics are used to calculate the first two and there is thus no
way of guaranteeing the location of a global
optimum~\citep{Last:2009:IDM:1557019.1557076}, which often results in
assuming fixed budgets~\citep{Kapoor05}. Alternatively, procedures
exclusively based on accuracy estimates try to identify the plateau of
the learning curve in terms of functional convergence. Among the most
popular ones are \textit{local detection} and \textit{learning curve
  estimation}~\citep{John96staticversus}, or \textit{linear regression
  with local sampling}~\citep{Provost:1999:EPS:312129.312188}, all of
them based on heuristics. Again, we cannot talk here about proximity
criteria, only of stopping conditions. More recently, this issue has
been corrected~\citep{VilaresDarribaRibadas16}, although the proposal
is still far from our objective because the proximity is expressed in
terms of the net contribution of each iteration to the learning
process, which provides not absolute but relative estimates.

Turning to robustness, one common idea is to generate different
versions (\textit{weak predictors}) of the partial learning curves by
changing the training data distribution repeatedly, and integrating
the hypotheses thus obtained. That way, \textit{bagging}\footnote{For
  \textit{bootstrap aggregating}.}
procedures~\citep{Breiman:1996:BP:231986.231989} build the predictors
in parallel to combine them by voting
(\textit{classification})~\citep{Leung:2003:ECV:956750.956825} or
averaging
(\textit{regression})~\citep{Leite:2007:IPB:1782254.1782263}. On the
contrary, \textit{boosting}
algorithms~\citep{Schapire:1990:SWL:83637.83645} do it sequentially,
which allows the adapting of such a distribution from the results
observed in previous predictors. This gives rise to
\textit{arcing}\footnote{For \textit{adaptive resampling and
    combining}.}  strategies~\citep{Freund+Schapire:1996}, where
increasing weight is placed on the more frequently misclassified
observations. Since these are the troublesome points, focusing on them
may do better than the neutral bagging
approach~\citep{Bauer:1999:ECV:599591.599607},
justifying~\citep{Garcia-Pedrajas:2014:BIS:2657522.2658088} its
popularity. Another well-known method is the \textit{$k$-fold cross
  validation}~\citep{Clark2010}, where the sample is randomly
partitioned into $k$ equal sized subsamples. For each fold, a model is
trained on the other $k-1$ ones and tested on it, which gives an
advantage to working with small data sets. The performance reported is
the average of the values computed. Whatever the format, such as
online proposals, all these build on the observations available, a
major constraint for making estimations beyond the last one. One
simple way to alleviate this problem is by using
\textit{anchors}~\citep{VilaresDarribaRibadas16}, i.e. extra examples
placed at the point of infinity to generate the weak predictor in each
cycle. As any one of such curves is the result of a fitting action,
the sum total of its \textit{residuals}, namely the differences
between the observed values and the fitted ones, is null. This gives
the anchor the chance to neutralize irregularities by choosing an
appropriate value.

Finally, completeness of the halting conditions has received no
attention before to the best of our knowledge, probably because so far
no additional assumptions on the sampling premises were necessary to
provide a practical solution.

\subsection{Our contribution}

It revolves around foundations, reliability and applicability to
provide correctness, robustness and completeness in a context for
which the ease of use is a priority. The former is established from a
set of working hypotheses widely recognized in {\sc ml} and a previous
outcome, whose interest was only formal to date, on learning
convergence in adaptive sampling~\citep{VilaresDarribaRibadas16}. To
that end, the adaptation to the premises of the theoretical result
must be guaranteed. The solution is based on the concept of
\textit{anchoring}, also proposed by those authors but only as a
robustness mechanism, which requires the development of a specific
family of such techniques.

We choose the domain of \textit{natural language processing} ({\sc
  nlp}) as case study, and more precisely the modelling of
\textit{part-of-speech} ({\sc pos}) \textit{taggers}, the classifiers
that mark a word in a text (corpus) as corresponding to a particular
{\sc pos}\footnote{A {\sc pos} is a class of words which have
  similar grammatical properties. Words that are assigned to the same
  {\sc pos} generally display similar behaviour in terms of syntax,
  i.e. they play analogous roles within the grammatical structure of
  sentences. The same applies in terms of morphology, in that they
  undergo inflection for similar properties. Common {\sc pos} labels
  are lexical categories (noun, verb, adjective, adverb, pronoun,
  preposition, conjunction, interjection, numeral, article,
  determiner, ...), the number or the gender.}, based
on its definition and context. The reasons are the significant
resource and time costs of generating training data, the complexity of
the relations to be learned and the fact that {\sc pos} tagging is
prior to any other {\sc nlp} task, so errors at this stage can lower
its performance~\citep{Song:2012:CSP:2390524.2390661}. That highlights
the scale of the challenge, but also justifies its interest and
popularity as experimentation field for new {\sc ml} facilities,
particularly around sampling
technology~\citep{Bloodgood:2009:MSA:1596374.1596384,Lewis:1994:SAT:188490.188495,Reichart:2010:TLC:1870568.1870579,Schmid:2008:ECP:1599081.1599179,Vlachos:2008:SCA:1349893.1350099},
which is the case here.

\section{The formal framework}
\label{section-formal-framework}

We introduce the concepts underlying the proposal, most of them
taken from~\cite{VilaresDarribaRibadas16}, denoting the real numbers
by $\mathbb{R}$ and the natural ones by $\mathbb{N}$, assuming that
$0 \not\in \mathbb{N}$. A prior question to clarify, because the
generation of {\sc ml}-based {\sc pos} taggers serves as an
illustration guide, is the accuracy notion usually accepted in that
kind of model. We define it as the number of correctly tagged tokens
divided by the total ones, expressed as a
percentage~\citep{VanHalteren1999} and calculated following some
generally admitted usages: all tokens in the testing data set are
counted, including punctuation marks, and it is supposed that only one
tag \textit{per} token is provided.

\subsection{The working hypotheses}

\begin{figure*}[htbp]
\begin{center}
\begin{tabular}{cc}
\hspace*{-.4cm}
\includegraphics[width=0.475\textwidth]{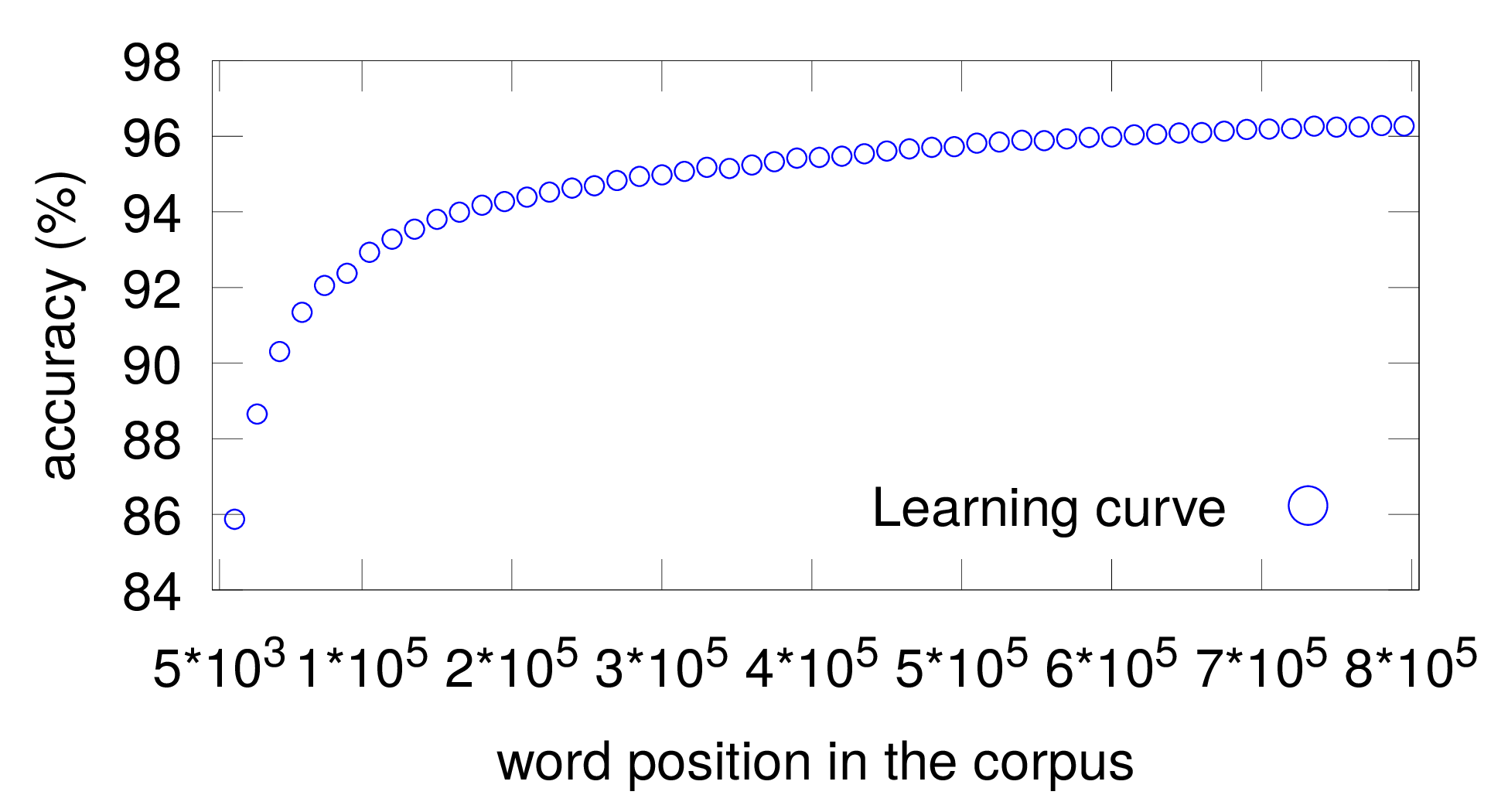} 
&
\hspace*{-.5cm}
\includegraphics[width=0.475\textwidth]{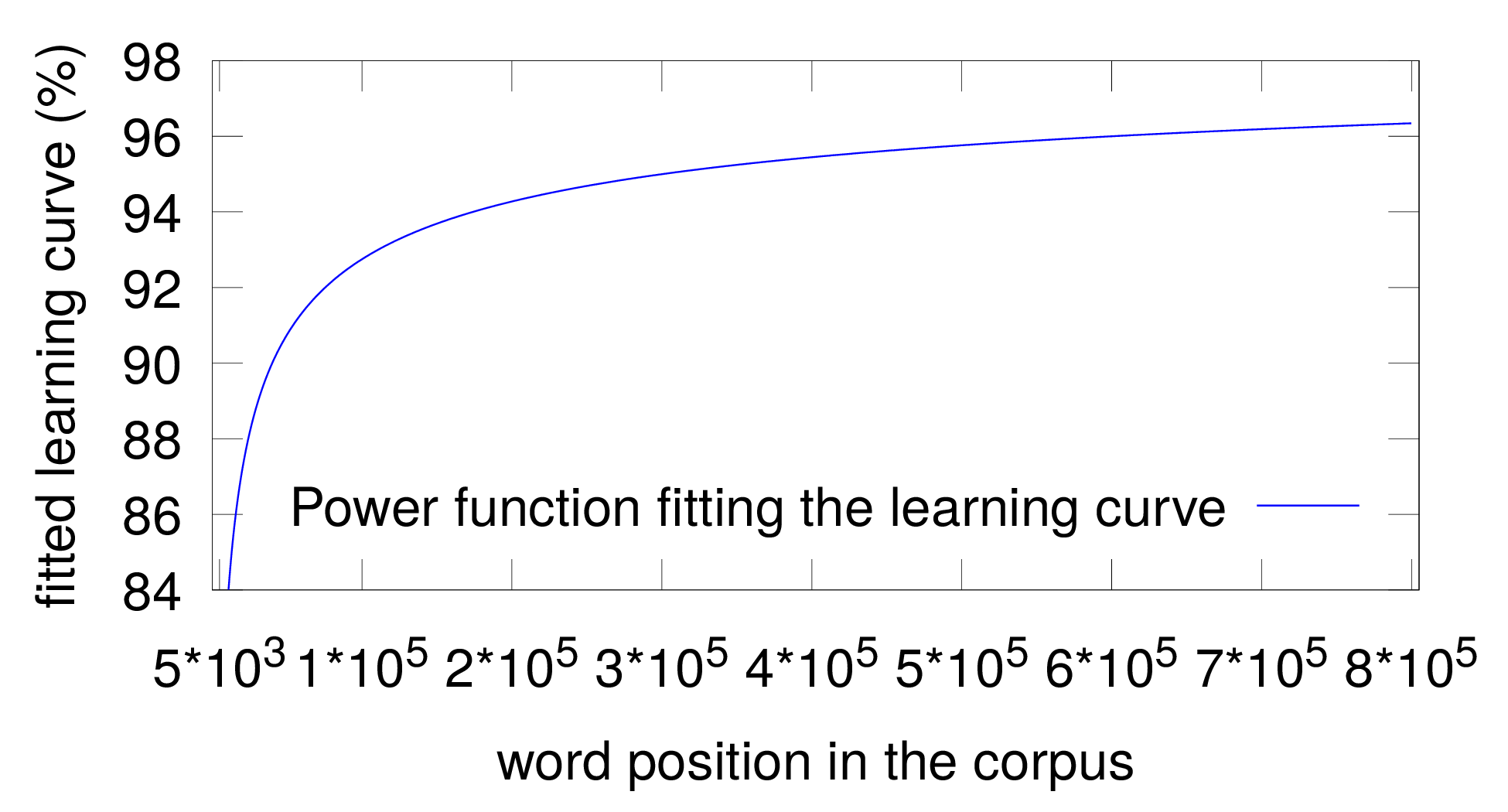}
\end{tabular}
\caption{Learning curve for fn{\sc tbl} on {\sc f}rown corpus, and an
  accuracy pattern fitting it.}
\label{fig-accuracy-fnTBL-Frown-5000-800000}
\end{center}
\end{figure*}

We start with a sequence of observations calculated from cases
incrementally taken from a training data base, meeting some conditions
to ensure a predictable progression of the estimates over a virtually
infinite interval. So, they are assumed to be independently and
identically
distributed~\citep{Domingo:2002:ASM:593433.593526,Schutze:2006:PTP:1183614.1183709,KATRIN08.335}. We
then accept that a learning curve is a positive definite and strictly
increasing function on $\mathbb{N}$, where numbers are the position of
the case in the training data base, and upper bounded by 100. This
results~\citep{Apostol} in a concave graph with horizontal
asymptote. Such hypotheses make up an idealized working frame to
support correctness, while real learners may deviate from it, thus
justifying the study of robustness. These deviations impact both the
concavity and increase of those curves, as shown in the left-most
diagram of Fig.~\ref{fig-accuracy-fnTBL-Frown-5000-800000} for the
{\it fast transformation-based learning} (fn{\sc tbl})
tagger~\citep{Ngai2001} on the \textit{Freiburg-Brown} ({\sc f}rown)
corpus of American English~\citep{Mair2007}.


\subsection{The notational support}

Having identified the context of the problem, it is necessary to
formalize the data structures we are going to work with, such as the
collection of instances whose convergence is intended to be measured.

\begin{df}
\label{def-learning-scheme} {\em (Learning scheme)}
Let ${\mathcal D}$ be a training data base,
$\mathcal K \subsetneq \mathcal D$ a set of initial items from
$\mathcal D$, and $\sigma:\mathbb{N} \rightarrow \mathbb{N}$ a
function. We define a {\em learning scheme} for $\mathcal D$ with {\em
  kernel} $\mathcal K$ and {\em step} $\sigma$, as a triple
$\mathcal{D}^{\mathcal{K}}_{\sigma}=[\mathcal{K},\sigma,\{\mathcal
D_i\}_{i \in \mathbb{N}}]$, such that
$\{\mathcal D_i\}_{i \in \mathbb{N}}$ is a cover of $\mathcal{D}$
verifying:
\begin{equation}
{\mathcal D}_1 := {\mathcal K} \mbox{ and }
{\mathcal D}_i := {\mathcal
  D}_{i-1} \cup {\mathcal I}_{i}, \; \mathcal I_i
\subset {\mathcal D} \setminus {\mathcal
  D}_{i-1}, \; \absd{{\mathcal I}_{i}}=\sigma(i), \; \forall i \geq 2
\end{equation}
\noindent with $\absd{{\mathcal I}_{i}}$ the cardinality of
${\mathcal I}_{i}$. We refer to $\mathcal{D}_i$ as the {\em individual
  of level} $i$ {\em for} $\mathcal{D}^{\mathcal{K}}_{\sigma}$.
\end{df}

A learning scheme relates a level $i$ with the position
$\absd{\mathcal D_i}$ in the training data base, determining the
sequence of observations
$\{[x_i, {\mathcal A}_{\dinfty{}}[{\mathcal D}](x_i)], \; x_i :=
\absd{\mathcal D_i} \}_{i \in \mathbb{N}}$, where
${\mathcal A}_{\dinfty{}}[{\mathcal D}](x_i)$ is the accuracy achieved
on such an instance by the learner. Thus, a level determines an
iteration in the adaptive sampling whose learning curve is
${\mathcal A}_{\dinfty{}}[{\mathcal D}]$, whilst ${\mathcal K}$
delimits a portion of ${\mathcal D}$ we believe to be enough to
initiate consistent evaluations of the training. For its part,
$\sigma$ identifies the sampling scheduling. As we want stable
estimates, partial learning curves are extrapolated according to a
functional pattern that verifies the working hypotheses, but are also
infinitely differentiable. This supplies graphs without disruptions
due to instantaneous jumps while ensuring their regularity.

\begin{df}
\label{def-accuracy-pattern-fitting} {\em (Accuracy pattern)}
Let $C^\infty_{(0,\infty)}$ be the C-infinity functions in
$\mathbb{R}^{+}$, we say that $\pi: \mathbb{R}^{{+}^{n}} \rightarrow
C^\infty_{(0,\infty)}$ is an {\em accuracy pattern} iff $\pi(a_1,
\dots, a_n)$ is positive definite, upper bounded, concave and strictly
increasing.
\end{df}

An example is the \textit{power law family} of curves $\pi(a,b,c)(x)
:=-a * x^{-b}+c$, hereafter used as the running one. Its upper bound
is the horizontal asymptote value \(\lim \limits_{x \rightarrow
  \infty} \pi(a,b,c)(x) = c\), and
\begin{equation}
\pi(a,b,c)'(x)=a * b * x^{-(b+1)} > 0 \hspace*{.75cm} 
\pi(a,b,c)''(x)=-a * b
* (b+1) * x^{-(b+2)} < 0 
\end{equation}
\noindent which guarantees increase and concavity in $\mathbb{R}^{+}$,
respectively. This is illustrated in the right-most diagram of
Fig.~\ref{fig-accuracy-fnTBL-Frown-5000-800000}, whose goal is to fit
the learning curve represented on the left-hand side. Here, the values
$a=542.5451$, $b=0.3838$ and $c=99.2876$ are provided by the
\textit{trust region method}~\citep{Branch1999}, a regression technique
minimizing the summed square of \textit{residuals}, namely the
differences between the observed values and the fitted
ones. Furthermore, as the aim is to determine the degree of 
convergence attained by the learning process, we need to
evaluate the progression of accuracy through the sequence of weak
predictors being computed.

\begin{df}
\label{def-trace} {\em (Learning trend)}
Let $\mathcal{D}^{\mathcal{K}}_{\sigma}$ be a learning scheme, $\pi$
an accuracy pattern and $\ell \in \mathbb{N}, \; \ell \geq 3$ a
position in the training data base $\mathcal{D}$. We define the {\em
learning trend of level} $\ell$ {\em for} ${\mathcal
D}^{\mathcal{K}}_{\sigma}$ {\em using} $\pi$, as a curve ${\mathcal
A}_{\ell}^\pi[{\mathcal D}^{\mathcal{K}}_{\sigma}] \in \pi$, fitting
the observations $\{[x_i, {\mathcal A}_{\dinfty{}}[{\mathcal
D}](x_i)], \; x_i := \absd{\mathcal D_i}
\}_{i=1}^{\ell}$. A sequence of learning trends ${\mathcal
  A}^\pi[{\mathcal D}^{\mathcal {K}}_{\sigma}] :=\{{\mathcal
  A}_{\ell}^\pi[{\mathcal D}^{\mathcal{K}}_{\sigma}]\}_{\ell \in
  \mathbb{N}}$ is called a {\em learning trace}. We refer to
$\{\alpha_\ell\}_{\ell \in \mathbb{N}}$ as the {\em asymptotic
  backbone} of ${\mathcal A}^\pi[{\mathcal D}^{\mathcal
    {K}}_{\sigma}]$, where $y = \alpha_\ell := \lim \limits_{x
  \rightarrow \infty} {\mathcal A}_{\ell}^\pi[{\mathcal
    D}^{\mathcal{K}}_{\sigma}](x)$ is the asymptote of ${\mathcal
  A}_\ell^\pi[{\mathcal D}^{\mathcal {K}}_{\sigma}]$.
\end{df}

\begin{figure*}[htbp]
\begin{center}
\begin{tabular}{cc}
\hspace*{-.4cm}
\includegraphics[width=0.475\textwidth]{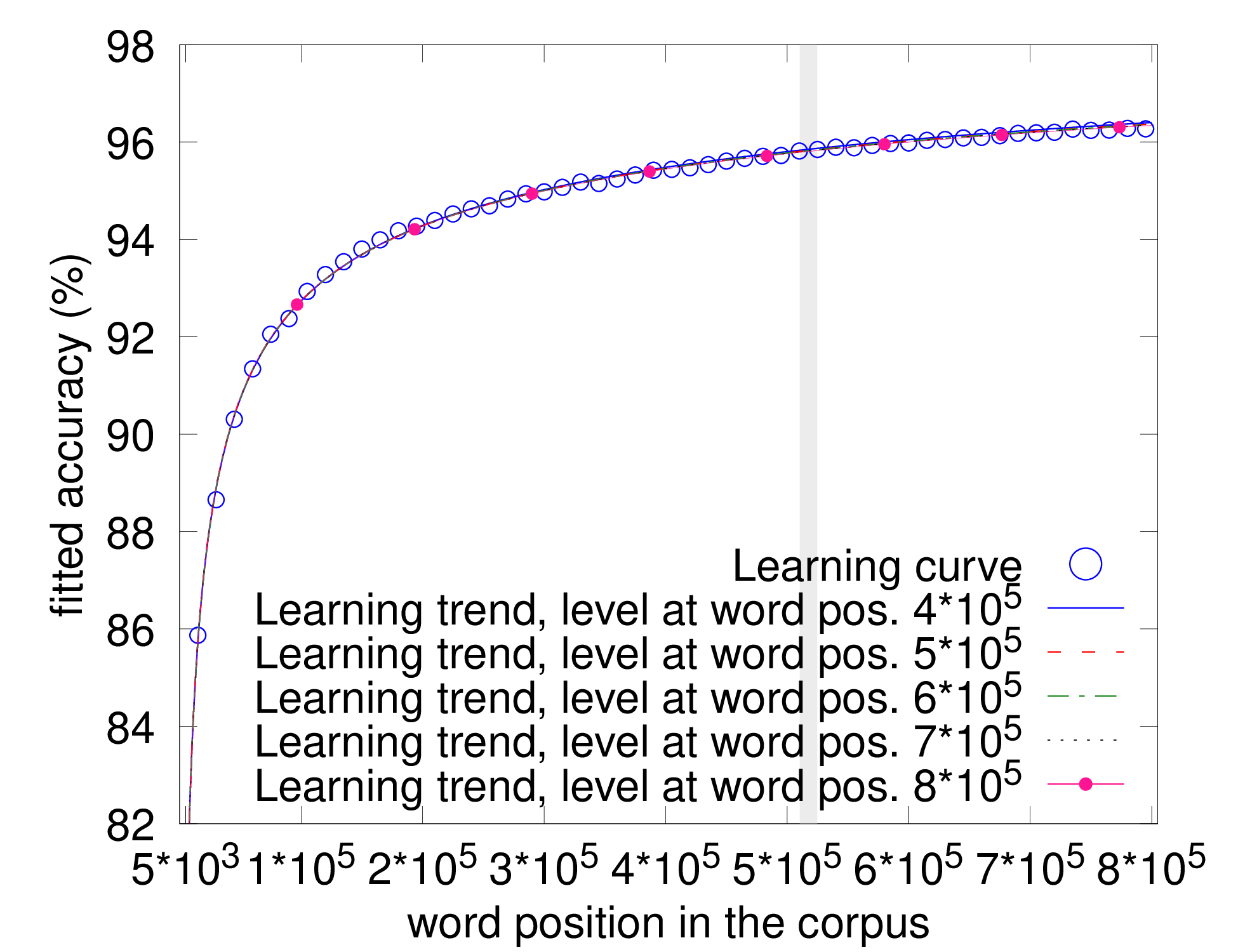}
& 
\hspace*{-.5cm}
\includegraphics[width=0.475\textwidth]{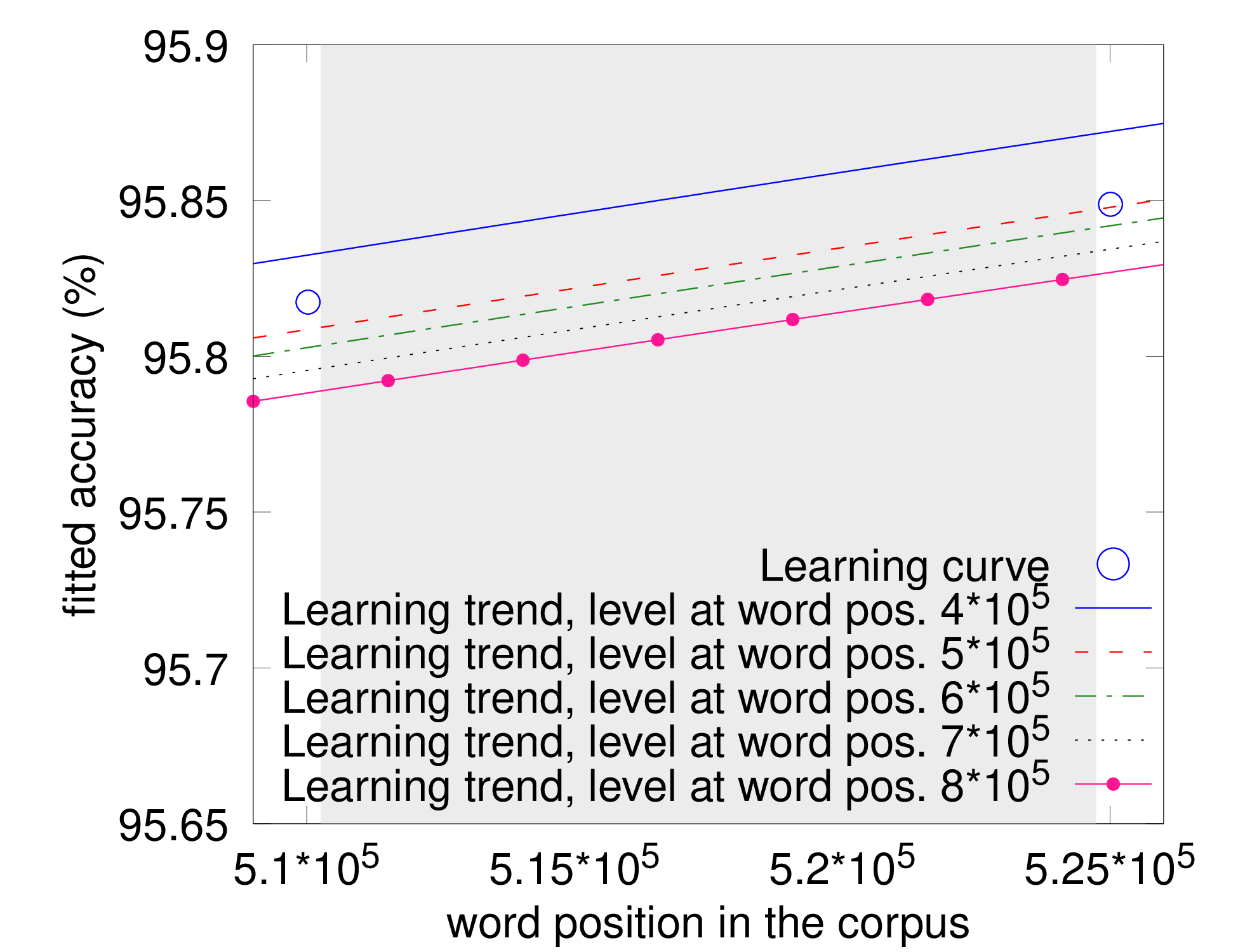}
\end{tabular}
\caption{Learning trace for fn{\sc tbl} on {\sc f}rown, with details
  in zoom.}
\label{fig-trace-and-level-sequence-nlls-fnTBL-Frown-5000-800000}
\end{center}
\end{figure*}

A learning trend
${\mathcal A}_{\ell}^\pi[{\mathcal D}^{\mathcal{K}}_{\sigma}]$
requires a level $\ell \geq 3$, because we need at least three
observations to generate a curve. Its value
${\mathcal A}_{\ell}^\pi[{\mathcal D}^{\mathcal{K}}_{\sigma}](x_i)$
represents the prediction for accuracy on a case $x_i$, using a model
generated from the first $\ell$ cycles of the learner. Accordingly,
the asymptotic term $\alpha_\ell$ is nothing other than the estimate
for the highest accuracy attainable. This way, a learning trace gives
a comprehensive picture of the increase in accuracy due to new
observations, as well as future expectations. Continuing with the
tagger fn{\sc tbl} and the corpus {\sc f}rown,
Fig.~\ref{fig-trace-and-level-sequence-nlls-fnTBL-Frown-5000-800000}
illustrates a portion of the learning trace with kernel and uniform
step function $5*10^3$, including a zoom view.

\section{The abstract model}
\label{section-abstract-model}

Learning traces lay the foundations for estimating absolute
convergence and error thresholds in adaptive
sampling~\citep{VilaresDarribaRibadas16}, thus giving coverage to the
correctness we are looking for, while only a relative solution is
described in practice. In short, assessing is done from the gain of
accuracy between consecutive iterations, which is not enough for our
purposes.  To overcome this limit, we turn to the concept of
\textit{anchoring}, originally introduced to improve robustness, but
which is now also useful to ensure completeness. The problem
formulates in terms of the uniform approximation of a learning curve
${\mathcal A}_{\dinfty{}}[{\mathcal D}]$ by means of the limit
function ${\mathcal A}_{\infty}^\pi[{\mathcal
    D}^{\mathcal{K}}_{\sigma}]$ for a learning trace ${\mathcal
  A}^\pi[{\mathcal D}^{\mathcal {K}}_{\sigma}] :=\{{\mathcal
  A}_{i}^\pi[{\mathcal D}^{\mathcal{K}}_{\sigma}]\}_{i \in
  \mathbb{N}}$ incrementally built from sampling. We start with a
brief reminder of the key results on robustness and correctness. The
reader can focus on the less formal aspects, to later address in
detail completeness as main contribution.


\subsection{Robustness}

Real learning conditions may diverge slightly from the ideal ones in
the working hypotheses on which correctness is stated. In this sense,
robustness is studied in the context of a more flexible set of
\textit{testing hypotheses}. These capture the notion of irregular
observation by assuming that learning curves are positive definite and
upper bounded by 100, but only quasi-strictly increasing and
concave. We then differentiate the alterations according to their
position in relation to the {\em working level} ({\sc wl}evel),
i.e. the cycle from which they would have a small enough impact to
work on their softening. As this depends on unpredictable factors such
as the magnitude, distribution and the very existence of these
disorders, a heuristic is necessary to identify it. Considering that
the model stabilizes as the training advances and that the monotony of
the asymptotic backbone is at the basis of the correctness for any
halting condition, a way of doing it is to categorize the variations
induced in the latter. This allows to estimate {\sc wl}evel as the
level providing the first fluctuation below a given ceiling and, once
passed, the \textit{prediction level} ({\sc pl}evel) marking the
beginning for learning trends which could feasibly predict the
learning curve, namely not exceeding its maximum (100).

\begin{df}
\label{def-level-of-work-trace}{\em (Working and prediction levels)}
Let ${\mathcal A}^\pi[{\mathcal D}^{\mathcal {K}}_{\sigma}]$ be a
learning trace with asymptotic backbone
$\{\alpha_i\}_{i \in \mathbb{N}}$, $\nu \in (0, 1)$,
$\varsigma \in \mathbb{N}$ and $\lambda \in \mathbb{N} \cup
\{0\}$. The {\em working level} {\em (}{\sc wl}{\em evel}{\em)} {\em for}
${\mathcal A}^\pi[{\mathcal D}^{\mathcal {K}}_{\sigma}]$ {\em with
  verticality threshold} $\nu$, {\em slowdown} $\varsigma$ {\em and
  look-ahead} $\lambda$, is the smallest
$\omega(\nu,\varsigma,\lambda) \in \mathbb{N}$ verifying
\begin{equation}
\label{equation-permissible-verticality-trace}
\frac{\sqrt[\varsigma]{\nu}}{1 - \nu} \geq \frac{\abs{\alpha_{i+1} -
    \alpha_{i}}}{x_{i+1} - x_i}, \; x_i :=
\absd{{\mathcal D}_{i}}, \; \forall i \mbox{ such that } 
\omega(\nu,\varsigma,\lambda) \leq i \leq
\omega(\nu,\varsigma,\lambda) + \lambda
\end{equation}
\noindent while the smallest $\wp(\nu,\varsigma,\lambda) \geq
\omega(\nu,\varsigma,\lambda)$ with
$\alpha_{\wp(\nu,\varsigma,\lambda)} \leq 100$ is the {\em prediction
  level} {\em (}{\sc pl}{\em evel}{\em )}. Unless they are necessary for
understanding, we shall omit the parameters, referring to {\em {\sc
    wl}evel} by $\omega$ {\em (}resp. {\sc pl}{\em evel} by $\wp${\em
  )}.
\end{df}

The {\sc wl}evel is the first level for which the normalized absolute
value of the slope of the line joining consecutive points on the
asymptotic backbone is less than the verticality threshold $\nu$,
which is corrected by a factor $1/\varsigma$ in order to slow down the
normalization pace for $\nu$. In effect, since the absolute value for a slope
is defined in the interval $[0,\infty)$, the normalizing function to
be applied can be expressed as follows:
\begin{equation}
\label{eq-normalizing-function}
  \begin{array}{lccc}
      \mbox{\sc n}: & [0, \infty) & \longrightarrow & [0, 1) \\
         & x           & \leadsto        & \frac{x}{x-1}
  \end{array}
\end{equation}
\noindent That way, given two consecutive points $(x_i,\alpha_i)$ and
$(x_{i+1},\alpha_{i+1})$ in the asymptotic backbone, the absolute
slope to be considered and the original condition we are looking for
are then, respectively:
\begin{equation}
\label{eq-normalizing-absolute-slopes}
   \frac{\abs{\alpha_{i+1} - \alpha_i}}{x_{i+1} - x_i} \hspace*{2cm}
   \mbox{and} \hspace*{2cm}
   \mbox{\sc n}(\frac{\abs{\alpha_{i+1} - \alpha_i}}{x_{i+1} - x_i}) <
   \nu
\end{equation}
\noindent The latter condition can be easily transformed into the
following equivalent one:
\begin{equation}
  \label{equation-permissible-verticality-trace-without-slowdown}
\frac{\nu}{1 - \nu} \geq \frac{\abs{\alpha_{i+1} -
    \alpha_{i}}}{x_{i+1} - x_i}  
\end{equation}   
from which, including the slowdown factor $1/\varsigma$ and the
condition on the look-ahead $\lambda$, we derive
Eq.~\ref{equation-permissible-verticality-trace}. The use of
normalized slopes corrected by the slowdown parameter $\varsigma$
allows recursion to infinitely large values and to extremely small
decimal fractions to be avoided, thus facilitating the setting of the
threshold $\nu$.

\begin{figure*}[htbp]
\begin{center}
\includegraphics[width=0.85\textwidth]{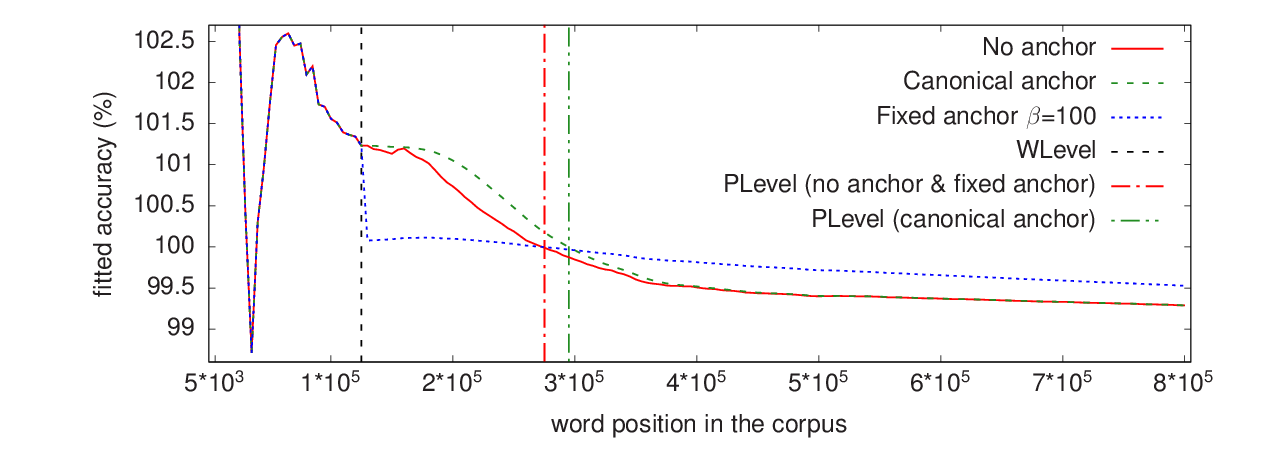} 
\end{center}
\caption{Working and prediction levels without and with canonical
  anchors for fn{\sc tbl} on {\sc f}rown, with details in zoom.}
\label{fig-lashes-trace-and-level-sequence-nlls-fnTBL-frown-5000-800000}
\end{figure*}

Intuitively, since slope values decrease together with the deviations in the
monotony studied, we can use them to categorize the latter, taking the
look-ahead $\lambda$ as verification window. We then place {\sc
  pl}evel on the first cycle with a learning trend below 100, which
would therefore be the first level with real predictive capacity,
since the previous ones would exceed this maximum accuracy value for
any model generated. In our example,
Fig.~\ref{fig-lashes-trace-and-level-sequence-nlls-fnTBL-frown-5000-800000}
shows the scale of such deviations before and after {\sc wl}evel, for
$\nu=2*10^{-5}$, $\varsigma=1$ and $\lambda=5$. Now the way is clear
to introduce \textit{anchoring} as a mechanism for robustness in
sampling.

\begin{df}
\label{def-anchoring-learning-trace} {\em (Anchoring learning trace)}
Let ${\mathcal A}^\pi[{\mathcal D}^{\mathcal {K}}_{\sigma}]$ be a
learning trace with {\em {\sc wl}evel} $\omega$, and 
$\{\hat{\mathcal A}_{\ell}(\infty)\}_{\ell > \omega} \subset
\mathbb{R}^+$. A {\em learning trend of level} $\ell > \omega$ {\em
  with anchor} $\hat{\mathcal A}_\ell(\infty)$ {\em for} ${\mathcal
  A}_{\dinfty{}}[{\mathcal D}]$ {\em using the
  accuracy pattern} $\pi$, is a curve $\hat{\mathcal
  A}_{\ell}^\pi[{\mathcal D}^{\mathcal{K}}_{\sigma}] \in \pi$ fitting
  the observations $\{[x_i, {\mathcal A}_{\dinfty{}}[{\mathcal
  D}](x_i)], \; x_i := \absd{\mathcal D_i}
\}_{i=1}^{\ell} \cup \; [\infty, \hat{\mathcal A}_{\ell}(\infty)]$. We denote by
$\hat\rho_\ell(i) := [{\mathcal A}_{\dinfty{}}[{\mathcal D}]
- \hat{\mathcal A}_{\ell}^\pi[{\mathcal
D}^{\mathcal{K}}_{\sigma}]](x_i)$ the {\em residual of} $\hat{\mathcal
A}_{\ell}^\pi[{\mathcal D}^{\mathcal{K}}_{\sigma}]$ {\em at the level}
$i$, by $\hat\rho_\ell(\infty) := \hat{\mathcal A}_{\ell}(\infty)
- \hat\alpha_\ell$ its {\em residual at the point of infinity} and by
$y = \hat\alpha_\ell$ its asymptote. When $\{\hat\alpha_\ell\}_{\ell
> \omega}$ is positive definite and converges monotonically to the
asymptotic value $\alpha_{\dinfty{}}$ of ${\mathcal
A}_{\dinfty{}}[{\mathcal D}]$, we say that
$\hat{\mathcal A}^\pi[{\mathcal D}^{\mathcal {K}}_{\sigma}]
:= \{\hat{\mathcal A}_{\ell}^\pi[{\mathcal
D}^{\mathcal{K}}_{\sigma}]\}_{\ell > \omega}$ is an {\em anchoring
learning trace} {\em of reference} $[{\mathcal A}^\pi[{\mathcal
D}^{\mathcal {K}}_{\sigma}], \omega]$.
\end{df}

These new learning trends differ from standard ones in the use of
fitting points at infinity, while in practice they are located as far
as the computer allows. The use of anchors to improve robustness
responds to the idea that extra observations facilitate the
realignment of the monotony for the asymptotic backbone, its residual
at the point of infinity being the maximum degree of smoothing
applicable at a given learning trend. This should be enough for small
irregularities, thus limiting the strategy to levels after {\sc
  wl}evel. A simple example is \textit{canonical anchoring}.

\begin{thm}
\label{th-canonical-anchoring-trace} {\em (Canonical anchoring)}
Let ${\mathcal A}^\pi[{\mathcal D}^{\mathcal {K}}_{\sigma}]$ be a
learning trace with asymptotic backbone
$\{\alpha_i\}_{i \in \mathbb{N}}$ and $\{\hat{\mathcal
A}_i(\infty)\}_{i > \omega}$ the sequence defined from its {\em {\sc
wl}evel} $\omega$ as
\begin{equation}
   \hat{\mathcal A}_{\omega+1}(\infty)
       := \alpha_{\omega} \hspace*{1cm} \hat{\mathcal
       A}_{i+1}(\infty)
       := \hat\alpha_{i}
       := \lim \limits_{x \rightarrow \infty} \hat{\mathcal
       A}_{i}^\pi[{\mathcal D}^{\mathcal{K}}_{\sigma}](x)
\end{equation}
\noindent with $\hat{\mathcal A}_{i}^\pi[{\mathcal
    D}^{\mathcal{K}}_{\sigma}]$ a curve fitting $\{[x_j, {\mathcal
    A}_{\dinfty{}}[{\mathcal D}](x_j)], \; x_j := \absd{\mathcal D_j}
\}_{j=1}^{i} \cup \; [\infty, \hat{\mathcal A}_{i}(\infty)]$, $\forall
i > \omega$. Then $\alpha_{\omega + i} \leq \hat\alpha_{\omega + i}$
{\em (resp.}  $\alpha_{\omega + i} \geq \hat\alpha_{\omega +
  i}${\em)}, $\forall i \in \mathbb{N}$, when $\{\alpha_i\}_{i \in
  \mathbb{N}}$ is decreasing {\em (resp.}  increasing{\em)}. Also,
$\{\hat{\mathcal A}_i^\pi[{\mathcal D}^{\mathcal {K}}_{\sigma}]\}_{i >
  \omega}$ is an anchoring learning trace of reference $[{\mathcal
    A}^\pi[{\mathcal D}^{\mathcal {K}}_{\sigma}],\omega]$, with
$\{\hat{\mathcal A}_i(\infty)\}_{i > \omega}$ its {\em canonical
  anchors}.
\end{thm}

\begin{pf}
To see in~\citep{VilaresDarribaRibadas16}. $\blacksquare$
\end{pf}

Since in each cycle the anchor takes the value from the asymptote of
the last learning trend, the technique described has a conservative
nature, which translates into a slower convergence process.
The effect of canonical anchoring in smoothing irregularities after
the {\sc wl}evel is illustrated vs. its absence, in our running
example by a dashed line, in
Fig.~\ref{fig-lashes-trace-and-level-sequence-nlls-fnTBL-frown-5000-800000}.


\subsection{Correctness}

It is addressed from the working hypotheses. That way, the uniform
convergence of learning traces has been demonstrated, and the topology
of the limit function described.

\begin{thm}
\label{th-uniform-convergence-trace} 
Let ${\mathcal A}^\pi[{\mathcal D}^{\mathcal {K}}_{\sigma}]$ be a
learning trace with or without anchors. Then, its asymptotic backbone
is monotonic and ${\mathcal A}_{\infty}^\pi[{\mathcal
D}^{\mathcal{K}}_{\sigma}] := {\lim \limits_{i \rightarrow \infty}}^u
{\mathcal A}_{i}^\pi[{\mathcal D}^{\mathcal{K}}_{\sigma}]$ exists, is
positive definite, increasing, continuous and upper bounded by 100 in
$(0, \infty)$.
\end{thm}

\begin{pf}
To see in~\citep{VilaresDarribaRibadas16}. $\blacksquare$
\end{pf}

This provides a way to estimate a learning curve ${\mathcal
A}_{\dinfty{}}[{\mathcal D}]$ by iteratively approximating the
function ${\mathcal A}_{\infty}^\pi[{\mathcal
D}^{\mathcal{K}}_{\sigma}]$, while a proximity criterion also
needs to measure the convergence (resp. error) threshold at each
stage. Namely, after fixing a level $i$ in a learning trace ${\mathcal
A}^\pi[{\mathcal D}^{\mathcal{K}}_{\sigma}]$, we have to calculate an
upper bound for the distance between ${\mathcal A}_{j}^\pi[{\mathcal
D}^{\mathcal{K}}_{\sigma}]$ and ${\mathcal A}_{\infty}^\pi[{\mathcal
D}^{\mathcal{K}}_{\sigma}]$ (resp. ${\mathcal A}_{\dinfty{}}[{\mathcal
D}]$) in the interval $[\absd{{\mathcal D}_j}, \infty), \; \forall
j \geq i$.

A previous result is needed. Let $\{(q_{i,x}^{i-1},q_{i,y}^{i-1})\}_{i
  \geq 4}$ (resp. $\{(p_{i,x}^{i-1},p_{i,y}^{i-1})\}_{i \geq 4}$) be
the sequence of the last (resp. first, if existing) points in
${\mathcal A}_{i}^\pi[{\mathcal D}^{\mathcal{K}}_{\sigma}] \cap
{\mathcal A}_{i-1}^\pi[{\mathcal D}^{\mathcal{K}}_{\sigma}]$. Then,
$\{q_{i,x}^{i-1}\}_{i \geq 4}$ and $\{q_{i,y}^{i-1}\}_{i \geq 4}$
(resp. $\{p_{i,x}^{i-1}\}_{i \geq 4}$ and $\{p_{i,y}^{i-1}\}_{i \geq
  4}$) are monotonic, except perhaps when there is a transition from
one (resp. two) to two (resp. one) intersection points at a level
$\imath$, or when we introduce/modify anchors. In that case,
$q_{\imath,y}^{\imath-1}$ and $q_{\imath+1,y}^{\imath}$
(resp. $p_{\imath,y}^{\imath-1}$ and $p_{\imath+1,y}^{\imath}$) may
momentarily invert their relative positions, and the same applies to
$q_{\imath,x}^{\imath-1}$ and $q_{\imath+1,x}^{\imath}$
(resp. $p_{\imath,x}^{\imath-1}$ and $p_{\imath+1,x}^{\imath}$).  We
then say that $\imath$ is a \textit{level of rupture for} ${\mathcal
  A}^\pi[{\mathcal D}^{\mathcal {K}}_{\sigma}]$. The order in
$\mathbb{N}$ is also extended to $\pmb{\mathbb{N}} := \mathbb{N} \cup
\{\infty, \dinfty{}\}$, in such a way that $\dinfty{} > \infty > i >
0, \; \forall i \in \mathbb{N}$.

\begin{thm}
  \label{th-correctness-trace} {\em (Correctness)} Let
  ${\mathcal A}^\pi[{\mathcal D}^{\mathcal {K}}_{\sigma}]$ be a
  learning trace with or without anchors, with {\sc wl}{\em evel} $\omega$, 
  and $y = \alpha_i$ the
  asymptote for ${\mathcal A}_i^\pi[{\mathcal D}^{\mathcal
      {K}}_{\sigma}], \; \forall i \in \pmb{\mathbb{N}} := \mathbb{N}
  \cup \{\infty, \dinfty{}\}$. Let also
  $(q_{i,x}^{i-1},q_{i,y}^{i-1})$ be the last point in ${\mathcal
    A}_i^\pi[{\mathcal D}^{\mathcal {K}}_{\sigma}] \cap {\mathcal
    A}_{i-1}^\pi[{\mathcal D}^{\mathcal {K}}_{\sigma}], \; \forall i
  \geq 4, \; i \neq \imath$, with $\imath$ level of rupture for
  ${\mathcal A}^\pi[{\mathcal D}^{\mathcal {K}}_{\sigma}]$. We then
  have, using for the occasion the notation ${\mathcal
    A}_{\dinfty{}}^\pi[{\mathcal D}^{\mathcal {K}}_{\sigma}]$ to refer
  ${\mathcal A}_{\dinfty{}}[{\mathcal D}]$, that
\begin{equation}
\label{equation-correctness-trace-decreasing}
\abs{[{\mathcal A}_k^\pi -
    {\mathcal A}_j^\pi][{\mathcal D}^{\mathcal {K}}_{\sigma}](x)} \leq
\varepsilon_i := \abs{q_{i,y}^{i-1} - \alpha_i}, \; \forall k,j \geq i
\geq 4, \; \forall x \in [q_{i,x}^{i-1}, \infty)
\end{equation}
\begin{equation}
\label{equation-correctness-trace-increasing}
\mbox{\em (resp. } \abs{[{\mathcal A}_k^\pi - {\mathcal
      A}_j^\pi][{\mathcal D}^{\mathcal {K}}_{\sigma}](x)} \leq
\varepsilon_i := \abs{q_{\dinfty{},y}^{i} - \alpha_{\dinfty{}}}, \;
\forall k,j \geq i \geq 1, \; \forall x \in [q_{\dinfty{},x}^i,
  \infty) \mbox{\em )}
\end{equation}
\noindent if $\{\alpha_i\}_{i > \omega + 1}$ is decreasing {\em (resp.}
increasing{\em )}, with $\{\varepsilon_i\}_{i > \omega +1}$ decreasing
and convergent to $0$
\end{thm}

\begin{pf}
To see in~\citep{VilaresDarribaRibadas16}. $\blacksquare$
\end{pf}

\begin{figure*}[htbp]
\begin{center}
\includegraphics[width=0.85\textwidth]{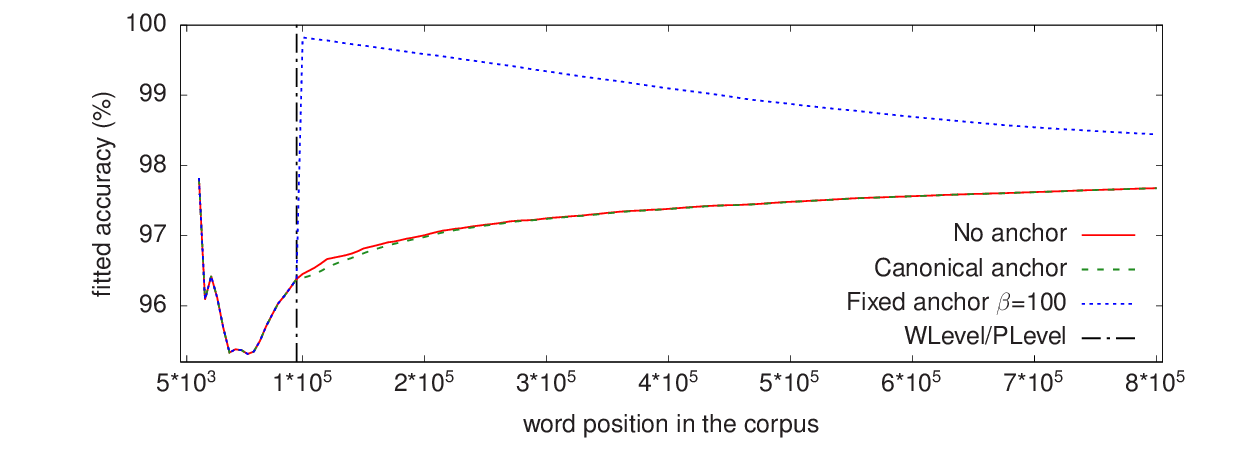} 
\end{center}
\caption{Asymptotic backbones without and with fixed anchors for
  {\sc m}ax{\sc e}nt on {\sc p}enn.}
\label{fig-comparative-asymptotic-backbones-from-increasing}
\end{figure*}

This result establishes the uniform convergence~\citep{Apostol} of the
learning trace ${\mathcal A}^\pi[{\mathcal D}^{\mathcal
    {K}}_{\sigma}]$ to the learning curve ${\mathcal
  A}_{\dinfty{}}[{\mathcal D}_{\sigma}]$. In particular, this implies
that the curve ${\mathcal A}_{\infty{}}^\pi[{\mathcal D}^{\mathcal
    {K}}_{\sigma}]$ we are iteratively approximating matches the
latter if the training process is long enough and, therefore,
$\alpha_{\infty} = \alpha_{\dinfty}$. Sadly, the result only has a
practical reading when the asymptotic backbone is decreasing, as it was
in Fig.~\ref{fig-trace-and-level-sequence-nlls-fnTBL-Frown-5000-800000}.
Otherwise, the bound depends on the final accuracy we want to estimate
($\alpha_{\dinfty{}}$), as with {\sc o}pen{\sc nlp} {\sc m}ax{\sc e}nt
(see {\em opennlp.apache.org/}) on the section of the \textit{Wall
  Street Journal} ({\sc wsj}) in the {\sc p}enn {\sc
  t}reebank~\citep{Marcus1999}. In this case, the asymptotic backbone
is increasing, as reflects the continuous line in
Fig.~\ref{fig-comparative-asymptotic-backbones-from-increasing}. This
gap must therefore be closed to guarantee the operability of the
approach.


\subsection{Completeness}
\label{subsection-completeness}

The bulk of our work focuses on this issue, through research on
anchoring as a tool to force the dynamics of convergence on learning
traces and obtain a decreasing asymptotic backbone, thus ensuring the
completeness sought. As a first step, we introduce a sufficient
condition to identify anchors verifying such a property.

\begin{thm}
\label{th-sufficient-condition-for-decreasing-backbone} 
Let ${\mathcal A}^\pi[{\mathcal D}^{\mathcal {K}}_{\sigma}]$ be a
learning trace with $\mbox{\em {\sc wl}evel}$ $\omega$ and $\mbox{\em
{\sc pl}evel}$ $\wp$, $y=\alpha_{\dinfty{}}$ the asymptote for the
learning curve ${\mathcal A}_{\dinfty{}}[{\mathcal D}]$ and
$\{\hat{\mathcal A}_i(\infty)\}_{i > \omega} \subset
\mathbb{R}^+$ a convergent sequence such that:
\begin{equation}
\label{equation-condition-anchors-lower-bounded}
\hat{\mathcal A}_i(\infty) \geq \alpha_{i}, \; \forall i > \wp
\end{equation}
\begin{equation}
\label{equation-condition-anchors-evolution}
\hat{\mathcal A}_i(\infty) -
\hat{\mathcal A}_{i+1}(\infty) \geq \hat\rho_{i}(\infty) -
\hat\rho_{i+1}(\infty), \; \forall i > \omega
\end{equation}
\noindent Let also $\hat{\mathcal A}_i^\pi[{\mathcal D}^{\mathcal
    {K}}_{\sigma}]$ be the learning trend with anchor $\hat{\mathcal
  A}_i(\infty)$ and asymptote $y = \hat\alpha_{i}$, and
  $\hat\rho_{i}(\infty) :=
\hat{\mathcal A}_i(\infty) -
\hat\alpha_{i}, \; \forall i > \omega$. Then, $\{\hat{\mathcal A}_i^\pi[{\mathcal
    D}^{\mathcal {K}}_{\sigma}]\}_{i > \omega}$ is an anchoring
learning trace of reference $[{\mathcal A}^\pi[{\mathcal D}^{\mathcal
{K}}_{\sigma}],\omega]$, such that $\{\hat\alpha_i\}_{i > \omega}$ is
decreasing.
\end{thm}

\begin{pf}
To see in Appendix~\ref{appendix-proofs}. $\blacksquare$
\end{pf}

So, a criterion to generate a learning trace with decreasing backbone
is to select a set of anchors never below the accuracy extrapolated at
each cycle to the total training data
base~(\ref{equation-condition-anchors-lower-bounded}), as long as the
learner does not override the readjustment applied by the
anchoring~(\ref{equation-condition-anchors-evolution}). Intuitively
simple, we will first study the practical utility of this idea for
the case of the previously introduced canonical anchors.


\begin{thm}
\label{th-canonical-anchoring-trace-decreasing}
Let $\hat{\mathcal A}^\pi[{\mathcal D}^{\mathcal {K}}_{\sigma}]$ be a
learning trace with canonical anchoring. We then have that if the
asymptotic backbone of the reference is decreasing, then the same
thing applies to that of the former. 
\end{thm}

\begin{pf}
To see in Appendix~\ref{appendix-proofs}. $\blacksquare$
\end{pf}

Unfortunately, this result does not settle the question at hand, i.e.
to guarantee decreasing asymptotic backbones by using anchors in order
to have a practical absolute measure of the convergence of learning
traces. As shown above, when using a canonical approach, this is
ensured only if the reference already verifies it.  Otherwise, the
resulting asymptotic backbone can also be increasing, as is shown
by the dashed line in
Fig.~\ref{fig-comparative-asymptotic-backbones-from-increasing}, and
another anchoring strategy is needed to respond to the challenge.


\subsubsection{Fixed anchoring}
\label{subsubsection-fixed-anchoring}

Learning trends are fitting curves on the set of observations
available at that time and, when using anchoring, also the value of
the latter associated to the point of infinity. This is the key to
making the proximity criterion in Theorem~\ref{th-correctness-trace}
fully operational, because the sum total of residuals on such curves
is null. So, to achieve decreasing asymptotic backbones it suffices to
fix anchors with negative or null residual, which is to say that they
must rise above all existing and future observations, for example
using values higher or equal than the maximum accuracy (100).
\begin{thm}
\label{th-fixed-anchoring-trace} {\em (Fixed anchoring)}
Let ${\mathcal A}^\pi[{\mathcal D}^{\mathcal {K}}_{\sigma}]$ be a
learning trace with $\mbox{\em {\sc wl}evel}$ $\omega$ and
$\hat{\mathcal A}_i^{\stackrel{\beta}{\pi}}[{\mathcal D}^{\mathcal
    {K}}_{\sigma}]$ the learning trend with anchor $\hat{\mathcal
  A}_i^\beta(\infty) := \beta \geq 100, \forall i > \omega$. Then,
$\hat{\mathcal A}^{\stackrel{\beta}{\pi}}[{\mathcal D}^{\mathcal
    {K}}_{\sigma}] := \{\hat{\mathcal
  A}_i^{\stackrel{\beta}{\pi}}[{\mathcal D}^{\mathcal
    {K}}_{\sigma}]\}_{i > \omega}$ is an anchoring learning trace of
reference $[{\mathcal A}^\pi[{\mathcal D}^{\mathcal
      {K}}_{\sigma}],\omega]$ with asymptotic backbone decreasing. We
call $\{\hat{\mathcal A}_i^\beta(\infty)\}_{i > \omega}$ the {\em
  fixed anchors of value} $\beta$ for $\hat{\mathcal
  A}^{\stackrel{\beta}{\pi}}[{\mathcal D}^{\mathcal {K}}_{\sigma}]$.
\end{thm}

\begin{pf}
To see in Appendix~\ref{appendix-proofs}. $\blacksquare$
\end{pf}

Fixed anchoring therefore guarantees the hypotheses under which we can
determine a computable estimation of the convergence and error
thresholds in absolute terms. Namely, it allows us to generate
learning traces with decreasing asymptotic backbones, regardless of
the training process considered. An example of this is shown in
Fig.~\ref{fig-comparative-asymptotic-backbones-from-increasing}, where
the monotony of the starting asymptotic backbone changes from
increasing to decreasing when using fixed anchors of value $\beta =
100$. In these conditions, the completeness of our abstract model
derives immediately.

\begin{thm}
\label{th-completeness-traces}
{\em (Completeness)} Let $\hat{\mathcal
  A}^{\stackrel{\beta}{\pi}}[{\mathcal D}^{\mathcal {K}}_{\sigma}]$ be
a learning trace of fixed anchoring with $\mbox{\em {\sc wl}evel}$
$\omega$ and $y = \alpha_i$ the asymptote for $\hat{\mathcal
  A}_i^{\stackrel{\beta}{\pi}}[{\mathcal D}^{\mathcal {K}}_{\sigma}],
\; \forall i \in \pmb{\mathbb{N}} := \mathbb{N}
  \cup \{\infty, \dinfty{}\}$. Let also be
$(q_{i,x}^{i-1},q_{i,y}^{i-1})$ the last point in $\hat{\mathcal
  A}_i^{\stackrel{\beta}{\pi}}[{\mathcal D}^{\mathcal {K}}_{\sigma}]
\cap \hat{\mathcal A}_{i-1}^{\stackrel{\beta}{\pi}}[{\mathcal
    D}^{\mathcal {K}}_{\sigma}], \; \forall i \geq 4, \; i \neq
\imath$, with $\imath$ level of rupture for $\hat{\mathcal
  A}^{\stackrel{\beta}{\pi}}[{\mathcal D}^{\mathcal
    {K}}_{\sigma}]$. We then have, using for the occasion the notation
$\hat{\mathcal A}_{\infty{}}^{\stackrel{\beta}{\pi}}[{\mathcal
    D}^{\mathcal {K}}_{\sigma}]$ {\em (}resp. $\hat{\mathcal
  A}_{\dinfty{}}^{\stackrel{\beta}{\pi}}[{\mathcal D}^{\mathcal
    {K}}_{\sigma}]${\em )} to refer ${\mathcal A}_\infty^\pi[{\mathcal
    D}^{\mathcal {K}}_{\sigma}]$ {\em (}resp. ${\mathcal
  A}_{\dinfty{}}[{\mathcal D}]${\em )}, that
\begin{equation}
\label{equation-completeness-traces}
\abs{[\hat{\mathcal A}_k^{\stackrel{\beta}{\pi}} - \hat{\mathcal
      A}_j^{\stackrel{\beta}{\pi}}][{\mathcal D}^{\mathcal
      {K}}_{\sigma}](x)} \leq \varepsilon_i := \abs{q_{i,y}^{i-1} -
  \alpha_i}, \; \forall k,j \geq i > \omega + 1, \; \forall x \in
    [q_{i,x}^{i-1}, \infty)
\end{equation}
\noindent with $\{\varepsilon_i\}_{i > \omega +1}$ decreasing and
convergent to $0$. We call the smallest
  $\iota \geq 4$ for which $\abs{q_{\iota,y}^{\iota-1} -
  \alpha_\iota} \leq \tau$, the {\em threshold level for} $\tau \in
\mathbb{R}^+$.

\end{thm}

\begin{pf}
To see in Appendix~\ref{appendix-proofs}. $\blacksquare$
\end{pf}

Contrary to what happened with canonical anchors, the fixed ones free
us from checking the decrease in the asymptotic backbone. Following
Theorem~\ref{th-correctness-trace}, this provides a practical and
extremely simple criterion for implementing a proximity condition
measuring absolute thresholds, henceforward referred to as
$\mathcal{H}_a$.

More in detail, given a learning trace with fixed anchoring
$\hat{\mathcal A}^{\stackrel{\beta}{\pi}}[{\mathcal D}^{\mathcal
    {K}}_{\sigma}]$ and a value $\tau \in \mathbb{R}^+$, we can assure
that, once the corresponding threshold level $\iota$ has been located:
\begin{equation}
\label{equation-absolute-proximity-condition-verification}
\abs{[\hat{\mathcal A}_\infty^{\stackrel{\beta}{\pi}} - 
    \hat{\mathcal A}_j^{\stackrel{\beta}{\pi}}][{\mathcal D}^{\mathcal {K}}_{\sigma}](x)} \leq
\varepsilon_\iota := \abs{q_{\iota,y}^{\iota-1} - \alpha_\iota} \leq \tau, \; \forall j \geq \iota
> \omega + 1, \; \forall x \in [q_{\iota,x}^{\iota-1}, \infty)
\end{equation}
Namely, all estimates in the interval $[q_{\iota,x}^{\iota-1}, \infty)
  \supseteq [\absd{\mathcal D_\iota}, \infty)$ for the learning trends
    computed from the $\iota$ level are at a distance from the curve
    ${\mathcal A}_\infty^\pi[{\mathcal D}^{\mathcal {K}}_{\sigma}]$ to
    which we converge (resp. the learning curve ${\mathcal
      A}_{\dinfty{}}[{\mathcal D}]$), which is less than the threshold
    $\tau$ set.

    As for canonical anchors, the fixed ones also contribute to a
    litte delay in the convergence, as can be seen in
    Fig.~\ref{fig-lashes-trace-and-level-sequence-nlls-fnTBL-frown-5000-800000}
    because their values are always higher than the asymptotes
    associated with the learning trends. One way to reduce this
    undesirable side effect is to provide the anchoring with
    mechanisms that allow it to adapt to the dynamics of the learning
    process.

\subsubsection{Endowing fixed anchoring with flexibility}
\label{subsubsection-endowing-fixed-anchoring-flexibility}

The goal is to define a configurable family of anchorings ensuring the
completeness of the proposal, thus allowing us to control the
performance via an appropriate setting. Our starting point is the
fixed anchor concept described above. Since the residuals at the point
of infinity are then negative, we can fine tune anchors from the
approximations for accuracy generated as learning progresses, without
compromising our objective.

\begin{thm}
\label{th-fixed-anchoring-trace-with-look-ahead} {\em (Fixed anchoring with
    look-ahead)} Let ${\mathcal A}^\pi[{\mathcal D}^{\mathcal
      {K}}_{\sigma}]$ be a learning trace with $\mbox{\em {\sc
      pl}evel}$ $\wp$, $\beta \geq 100$, $\ell \in \mathbb{N}$ and
  $\{\hat{\mathcal A}_i^{\beta,\ell}(\infty)\}_{i > \omega}$ the
  sequence defined from its $\mbox{\em {\sc wl}evel}$ $\omega$ by
\begin{equation}
\hat{\mathcal A}_i^{\beta,\ell}(\infty) := \beta, \; \forall \; \wp +
\ell + 1 > i > \omega
\hspace*{1cm}
\hat{\mathcal A}_i^{\beta,\ell}(\infty) := \hat\alpha_{\wp + \ell}^{\beta,\ell},
\; \forall i \geq \wp + \ell + 1
\end{equation}
\noindent Let also $\hat{\mathcal
  A}_i^{\stackrel{\beta,\ell}{\pi}}[{\mathcal D}^{\mathcal
    {K}}_{\sigma}]$ be the learning trend with anchor $\hat{\mathcal
  A}_i^{\beta,\ell}(\infty), \forall i > \omega$. Then, $\hat{\mathcal
  A}^{\stackrel{\beta,\ell}{\pi}}[{\mathcal D}^{\mathcal
    {K}}_{\sigma}] := \{\hat{\mathcal
  A}_i^{\stackrel{\beta,\ell}{\pi}}[{\mathcal D}^{\mathcal
    {K}}_{\sigma}]\}_{i > \omega}$ is an anchoring learning trace of
reference $[{\mathcal A}^\pi[{\mathcal D}^{\mathcal
      {K}}_{\sigma}],\omega]$ and asymptotic backbone
$\{\hat\alpha_i^{\beta,\ell}\}_{i > \omega}$ decreasing, and we call
$\{\hat{\mathcal A}_i^{\beta,\ell}(\infty)\}_{i > \omega}$ its set of {\em
  fixed anchors with look-ahead} $\ell$ {\em and value} $\beta$.
\end{thm}

\begin{pf}
To see in Appendix~\ref{appendix-proofs}. $\blacksquare$
\end{pf}

Intuitively, we are talking about a learning trace with conventional
fixed anchoring, in which the anchor is subject to revision once the
study of the levels in the interval $[\omega+1,\wp + \ell]$ has been
completed. Since $\omega \leq \wp$, this interval includes $[\wp+1,\wp
  + \ell]$, which in either case allows us to take advantage of the
knowledge provided by the first $\ell$ iterations from the {\sc
  pl}evel $\wp$. As these are the best performing training cycles --
together with the one associated with level $\wp$, in the case where
$\omega = \wp$ -- among those with real predictive capability,
convergence can be expected to accelerate significantly once the
anchor has been updated.

\begin{figure*}[htbp]
\begin{center}
\includegraphics[width=0.85\textwidth]{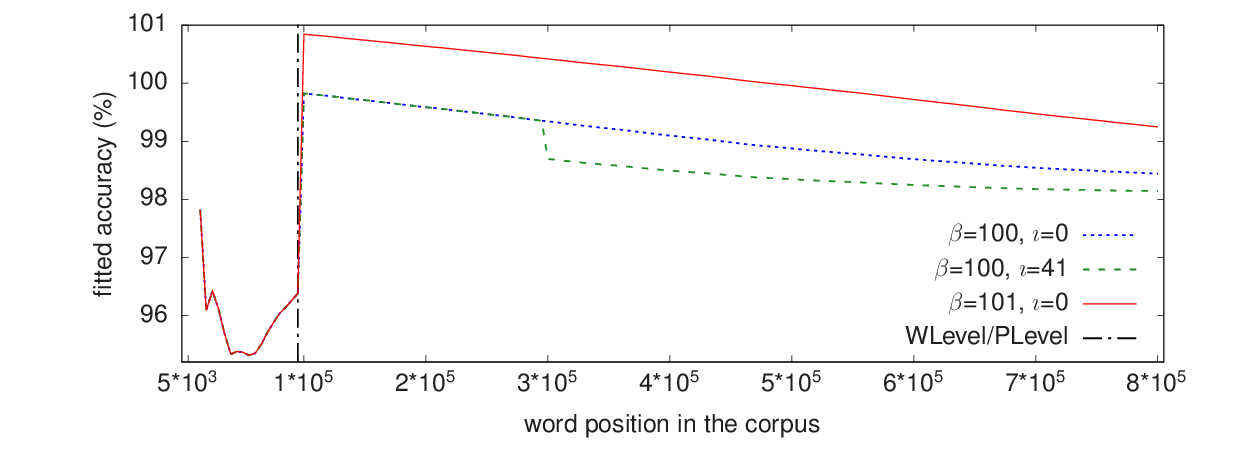}
\end{center}
\caption{Asymptotic backbones with fixed anchors for {\sc m}ax{\sc
    e}nt on {\sc p}enn (look-ahead $\imath$, value $\beta$).}
\label{fig-comparative-asymptotic-backbones}
\end{figure*}

To illustrate this, we look again at the learning process shown in
Fig.~\ref{fig-comparative-asymptotic-backbones-from-increasing}, to
compare in Fig.~\ref{fig-comparative-asymptotic-backbones} the
asymptotic backbones associated to some value/look-ahead combinations,
focusing on two use cases: different look-aheads ($\imath = 0$ and
$\imath = 41$) with the same value ($\beta = 100$) and the same
look-ahead ($\imath = 0$) with different values ($\beta = 100$ and
$\beta = 101$). In the former scenario, we check how a non-trivial
look-ahead ($\imath = 41$) causes the desired effect. Also, as might
be expected, the second one suggests that the closer the anchor to the
real accuracy, the faster the convergence. All the above underscores
the importance of an in-depth study on the impact of values and
look-aheads on accuracy prediction. The objective is to establish
whether these first impressions have a formal basis that
allows us to effectively categorize the anchoring strategies
described.

\begin{thm}
\label{th-anchoring-trace-comparison} {\em (Anchoring categorization)}
Let $\hat{\mathcal A}^\pi[{\mathcal D}^{\mathcal {K}}_{\sigma}]$,
$\hat{\mathcal A}^{\stackrel{\eta,\imath}{\pi}}[{\mathcal D}^{\mathcal
    {K}}_{\sigma}]$, $\hat{\mathcal
  A}^{\stackrel{\beta,\imath}{\pi}}[{\mathcal D}^{\mathcal
    {K}}_{\sigma}]$ and $\hat{\mathcal
  A}^{\stackrel{\beta,\jmath}{\pi}}[{\mathcal D}^{\mathcal
    {K}}_{\sigma}]$ be learning traces of reference $[{\mathcal
    A}^\pi[{\mathcal D}^{\mathcal {K}}_{\sigma}],\omega]$, generated
from canonical and fixed anchors with look-aheads $\imath, \jmath \in
\mathbb{N} \cup \{0\}$ for values $\eta, \beta \geq 100$
respectively. We then have that, in any case {\em (resp.} when
$\{\alpha_i\}_{i \in \mathbb{N}}$ decreasing{\em )}, it verifies that
$\forall i > \omega$
\begin{equation}
\label{eq-speed-comparison-for-standard-and-canonical-asymptotic-backbones}
\abs{\alpha_i - \alpha_{\dinfty{}}} \leq \abs{\hat\alpha_i -
  \alpha_{\dinfty{}}}
\end{equation}
\noindent and also $\forall \jmath > \imath, \eta > \beta, i \geq
\wp + \jmath$
\begin{equation}
\label{eq-speed-comparison-for-anchoring-asymptotic-backbones}
\abs{\hat\alpha_i^{\beta,\jmath} - \alpha_{\dinfty{}}} \leq
\abs{\hat\alpha_i^{\beta,\imath} - \alpha_{\dinfty{}}} \leq
\abs{\hat\alpha_i^{\eta,\imath} - \alpha_{\dinfty{}}}
\end{equation}
\begin{equation}
\label{eq-speed-comparison-for-decreasing-asymptotic-backbones}
(\mbox{resp. } \abs{\alpha_i - \alpha_{\dinfty{}}} \leq
\abs{\hat\alpha_i - \alpha_{\dinfty{}}} \leq
\abs{\hat\alpha_i^{\beta,\jmath} - \alpha_{\dinfty{}}} \leq
\abs{\hat\alpha_i^{\beta,\imath} - \alpha_{\dinfty{}}} \leq
\abs{\hat\alpha_i^{\eta,\imath} - \alpha_{\dinfty{}}})
\end{equation}
\noindent with $\{\hat\alpha_i\}_{i > \omega}$,
$\{\hat\alpha_i^{\eta,\imath}\}_{i > \omega}$,
$\{\hat\alpha_i^{\beta,\imath}\}_{i >
  \omega}$,
$\{\hat\alpha_i^{\beta,\jmath}\}_{i > \omega}$
and $\{\alpha_i\}_{i \in \mathbb{N}}$ their corresponding asymptotic
backbones, and $\alpha_{\dinfty{}}$ the asymptotic value for the
learning curve ${\mathcal A}_{\dinfty{}}[{\mathcal D}]$.
\end{thm}

\begin{pf}
To see in Appendix~\ref{appendix-proofs}. $\blacksquare$
\end{pf}

The result guides the choice of anchoring. So, the fastest way
to converge when the working hypotheses verify is to avoid fixed
anchors, except if the asymptotic backbone is not decreasing. When
this is quasi-decreasing because only the testing hypotheses are
guaranteed, the canonical strategy is the most adequate. Finally,
fixed anchoring with look-ahead is the alternative when no data about
the training are available. The convergence speed here is inversely
proportional to the value, which is why the best option is 100, the
minimum one. Once a value is selected, the look-ahead introduces an
extra factor to speed up the convergence according to its length, but
only from the time the anchor is updated. Because of this, our
objective could be reached before the latter is activated, in such a
way that a smaller look-ahead might be more effective. Namely, an
optimal choice depends on the convergence threshold -- that
matches, by Theorem~\ref{th-correctness-trace}, the error
one -- we are trying to identify, thus suggesting an iterative
approach for dealing with it. We then make the decision to depend on
the degree of convergence reached with respect to that threshold,
taking into account that the first reliable level for predictions is
{\sc pl}evel.

\begin{df}
\label{def-percentage-uncovered-threshold}
{\em (Percentage of uncovered threshold)} Let
$\hat{\mathcal A}^{\stackrel{\beta}{\pi}}[{\mathcal D}^{\mathcal
    {K}}_{\sigma}]$ be a learning trace with fixed anchoring and {\sc
  pl}{\em evel} $\wp$, and $\tau$ a threshold for a proximity
condition ${\mathcal H}$. We define its {\em percentage of uncovered
  threshold for} $\tau$ {\em on} ${\mathcal H}$
{\em at a level} $\ell > \wp + 1$ as
\begin{equation}
\label{eq-percentage-uncovered-threshold}
\mbox{\sc put}[\hat{\mathcal A}^{\stackrel{\beta}{\pi}}[{\mathcal
      D}^{\mathcal {K}}_{\sigma}],\tau,{\mathcal H}](\ell) :=
\left\{ \begin{array}{ll} 100*\frac{\abs{\alpha_{\dinfty{}} -
      \hat\alpha_\ell^\beta}_{\mathcal H} -
    \tau}{\abs{\alpha_{\dinfty{}} - \hat\alpha_{\wp +
        2}^\beta}_{\mathcal H} - \tau} & \mbox{if} \; \;
  \abs{\alpha_{\dinfty{}} - \hat\alpha_\ell^\beta}_{\mathcal H} \geq
  \tau \\ 0 & \mbox{otherwise}
        \end{array}
\right.
\end{equation}
\noindent with $\abs{\alpha_{\dinfty{}} -
  \hat\alpha_\ell^\beta}_{\mathcal H}$ the estimates of ${\mathcal H}$
for $\abs{\alpha_{\dinfty{}} - \hat\alpha_\ell^\beta}$, and 
$\hat\alpha_\ell^\beta$ {\em (}resp. $\alpha_{\dinfty{}}${\em )} the
asymptotic value for the learning trend {\em (}resp. learning
curve{\em )} $ \hat{\mathcal
  A}_{\ell}^{\stackrel{\beta}{\pi}}[{\mathcal D}^{\mathcal
    {K}}_{\sigma}]$ {\em (}resp. ${\mathcal A}_{\dinfty{}}[{\mathcal
    D}]${\em )}.
\end{df}

\begin{figure*}[htbp]
\begin{center}
\includegraphics[width=0.75\textwidth]{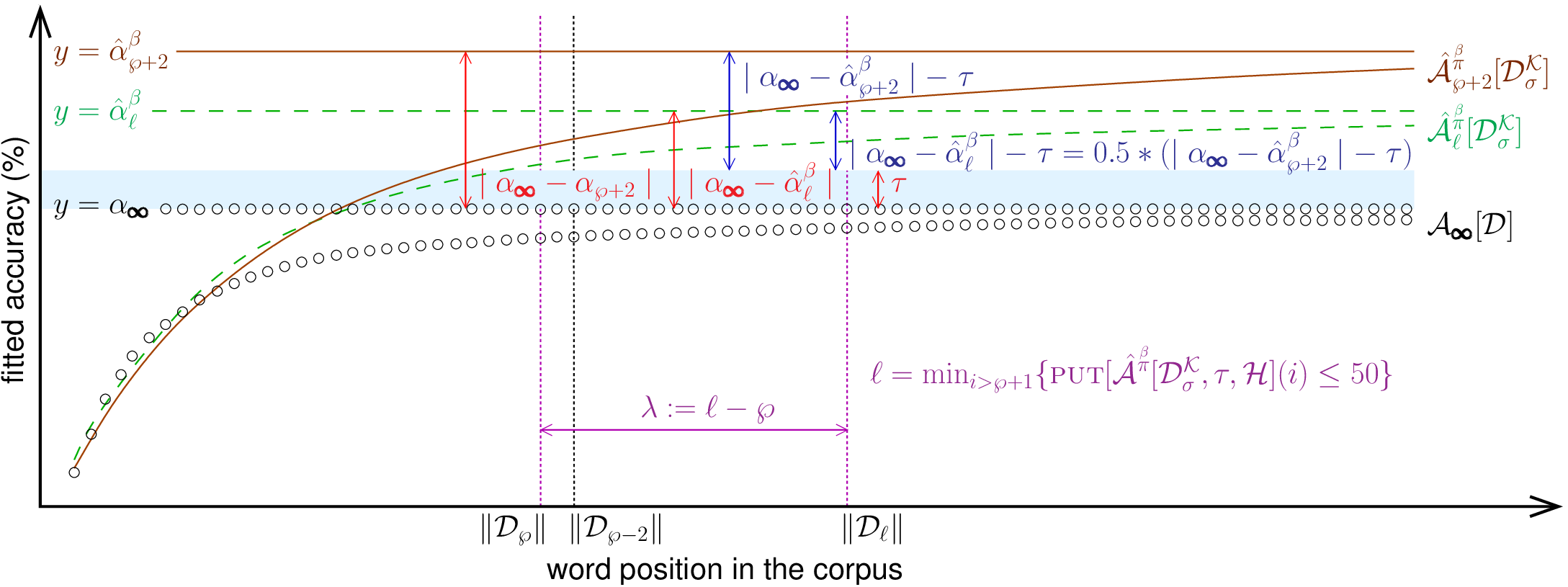}
\end{center}
\caption{Computing the minimal look-ahead $\lambda$ for the tentative
  {\sc put} value of 50 associated with a fixed anchoring.}
\label{fig-percentage-uncovered-threshold}
\end{figure*}

The {\sc put} takes values in the interval $[0,100]$ and is
decreasing in the level covered. Its geometric interpretation is shown in
Fig.~\ref{fig-percentage-uncovered-threshold}. Its
minimum is 0 and is reached when the estimated degree of approximation improves
or equals the threshold $\tau$ set by the user, in asymptotic
terms. That is, when the efficiency of the last asymptotic value as
fixed anchor can no longer be improved, which implies that the
current level is the last one at which a possible update of the fixed
anchor based on learning asymptotic values makes sense. As for the
maximum value, it is reached at level $\wp+2$, the first at which the
anchor could be updated and is $100$, unless at that point the
convergence is already effective.

Once we have captured the concept of {\sc put} and the user has
selected a particular value, we can assess for which particular level
a fixed anchoring should be updated. That is, we are in a position to
determine the look-ahead that matches the user's requirements.

\begin{df}
\label{def-minimal-look-ahead-for-a-tentative-PUT-value}
      {\em (Minimal look-ahead for a tentative {\sc put} value in
        fixed anchoring)} 
Let $\hat{\mathcal A}^{\stackrel{\beta}{\pi}}[{\mathcal D}^{\mathcal {K}}_{\sigma}]$ be a
learning trace with fixed anchoring and {\sc pl}{\em evel} $\wp$,
$\tau$ a threshold for a proximity condition ${\mathcal H}$ and $\zeta
\in [0,100]$ the {\sc put} value we want to reach before updating
the anchor. We then define
\begin{equation}
\lambda := \min_{\ell > \wp + 1}
\{\mbox{\sc put}[\hat{\mathcal A}^{\stackrel{\beta}{\pi}}[{\mathcal D}^{\mathcal
      {K}}_{\sigma}, \tau, {\mathcal H}](\ell) \leq \zeta\} - \wp
\end{equation}
as the {\em minimal look-ahead for the tentative} {\sc put} {\em value} $\zeta$.
\end{df}

Formally, the fact that {\sc put} is monotonic and bound guarantees
that the concept of minimal look-ahead is
well-defined~\citep{Apostol}. Regarding its geometric interpretation
and calculation process, both of them are schematized in
Fig.~\ref{fig-percentage-uncovered-threshold}. We start from the
asympotic values $\alpha_{\dinfty{}}$, $\hat\alpha_{\ell}^\beta$ and
$\hat\alpha_{\wp+2}^\beta$, which respectivelly correspond to the
learning curve, the last learning trend and the first one for which
the fixed anchor could be updated.  We can then estimate, according
the ${\mathcal H}$ criterion, the distance yet to be completed from
the current level (resp. the first level at which a look-ahead would
be applicable) to reach the required convergence threshold, by means
of the value $\abs{\alpha_{\dinfty{}} -
  \hat\alpha_\ell^\beta}_{\mathcal H} - \tau$
(resp. $\abs{\alpha_{\dinfty{}} - \hat\alpha_{\wp +2}^\beta}_{\mathcal
  H} - \tau$).

By way of illustration, assuming that we want the update of a fixed
anchoring to be activated once half of the convergence distance has
been covered, it is sufficient to identify the first level $\ell > \wp
+1$ at which the {\sc put} is less than or equal to $50$ and then
choose $\lambda := \ell - \wp$ as the minimal look-ahead for that
tentative value. Returning to the example in
Fig.~\ref{fig-comparative-asymptotic-backbones} when the value of the
fixed anchoring for the learning trace is $\beta=100$, this results in
a look-ahead of $41$, whose associated asymptotic backbone shows the
best results from among those represented.

\section{The testing frame}
\label{section-testing-frame}

The focus is now on evaluating the proposal, taking the generation of
{\sc ml}-based {\sc pos} taggers as a case study. This first involves
the design of a uniform framework, in the sense that its standards of
evidence do not favour any particular \textit{approximation
  technique}, namely any proximity condition nor particular setting.
Once a learner and a training data base are fixed, the aim is to
assess the impact of forcing the completeness of our argument, the
balance between its costs and benefits, and also its stability against
look-ahead variations.

To do this, we introduce the corresponding performance metric,
together with its monitoring architecture for data collection. The
latter captures the concept of testing round (\textit{run}), which
serves to normalize the conditions under which the experiments take
place. Runs only distinguishable by their approximation technique are
grouped around a baseline, in what we call a \textit{local testing
  frame}, thus providing the environment we are looking for.

\subsection{The monitoring structure}

After setting an {\sc ml} task represented by a learning trace
${\mathcal{A}}^{\pi}[\mathcal{D}^{\mathcal{K}}_{\sigma}]$, the goal is
to standardize the testing conditions, with a view to allowing for its
objective assessment.

\subsubsection{The testing rounds}

Our evaluation basis is the {\em run}, a tuple
$\hat{\mathcal{E}}^{\mathcal H}
=[\hat{\mathcal{A}}^{\pi}[\mathcal{D}^{\mathcal{K}}_{\sigma}],\tau,
  \mathcal{H}]$ characterized by a learning trace
$\hat{\mathcal{A}}^{\pi}[\mathcal{D}^{\mathcal{K}}_{\sigma}]$ of
reference ${\mathcal{A}}^{\pi}[\mathcal{D}^{\mathcal{K}}_{\sigma}]$
and anchoring $\, \hat{} \,$, a convergence or error threshold $\tau$ and
a proximity condition $\mathcal{H}$. We can then express the {\sc
  put} as a function on the runs, denoting is value on a given
testing round $\hat{\mathcal{E}}^{\mathcal H}$ as ${\mbox{\sc
    put}}[\hat{\mathcal{E}}^{\mathcal H}]$, while the notion of
\textit{prediction level} is naturally extended as the one of its
learning trace and denoting it by $\mbox{{\sc
    pl}evel}[\hat{\mathcal{E}}^{\mathcal H}]$. A value $\mbox{{\sc
    cl}evel}[\hat{\mathcal{E}}^{\mathcal H}]$ is associated as its
\textit{convergence level}, the one from which $\mathcal{H}$ verifies
for $\tau$ and the training stops. Given two runs
$\hat{\mathcal{E}}^{\mathcal H} = [\hat{\mathcal
    A}^{\pi}[\mathcal{D}^{\mathcal{K}}_{\sigma}],\tau, {\mathcal H}]$
and $\check{\mathcal{E}}^{{\mathcal H}'} = [\check{\mathcal
    A}^{\pi}[\mathcal{D}^{\mathcal{K}}_{\sigma}],\tau, {\mathcal
    H}']$, they are \textit{similar} when they are only
distinguishable by the anchoring strategy used, i.e. when ${\mathcal H} =
{\mathcal H}'$, and \textit{dissimilar} otherwise. As our proposal
requires decreasing asymptotic backbones, it is mandatory to use runs
meeting such a condition to give a comprehensive understanding of the
tests.

\subsubsection{The testing scenarios}

Our aim is to define run groupings to study the behaviour of an
approximation technique beyond the qualitative considerations in
Theorem~\ref{th-anchoring-trace-comparison}. As the idea is to do it
through a ratio with respect to a benchmark, it is necessary to compare
runs sharing the reference but not the anchoring or the proximity
condition. To this end, we introduce an order relation for these
latter ones.


\begin{thm}
\label{def-relation-halting-condition-faster}
Let ${\mathcal L}$ be a set of runs only distinguishable by a
proximity condition taken from $\Lambda$. Then, the relation 
\begin{equation}
\label{eq-relation-halting-condition-faster}
\forall {\mathcal H}, {\mathcal H}' \in \Lambda, \; {\mathcal H}
\succeq_{\mathcal L} {\mathcal H}' :\Leftrightarrow \mbox{{\sc
    cl}{\em evel}}[{\mathcal{E}}^{{\mathcal H}}] \leq \mbox{{\sc
    cl}{\em evel}}[{\mathcal{E}}^{{\mathcal H}'}], \forall
{\mathcal{E}}^{\mathcal H}, 
{\mathcal{E}}^{{\mathcal H}'} \in {\mathcal L}
\end{equation}
\noindent defines a partial order and we say that ${\mathcal H}$ {\em is faster than}
${\mathcal H}'$ {\em on} ${\mathcal L}$.
\end{thm}         

\begin{pf}
Trivial. $\blacksquare$
\end{pf}

Comparing runs also entails normalizing a threshold when it applies to
proximity conditions with different scales, as with the absolute
criterion just introduced and the relative one based on the layered
correctness~\citep{VilaresDarribaRibadas16}, which we refer to as
$\mathcal{H}_a$ and $\mathcal{H}_r$, respectively. Once that happens,
we first fix the relative threshold $\tau_r$ to be used with
$\mathcal{H}_r$. The corresponding absolute one $\tau_a$ concerning
$\mathcal{H}_a$ is then calculated, as
Theorem~\ref{th-correctness-trace} indicates for decreasing asymptotic
backbones, from the level at which $\mathcal{H}_r$ determines the
layered convergence. Such absolute thresholds are the ones referred to
in the runs, which can then be grouped for testing purposes.

\begin{df}
\label{def-local-testing-frame}{\em (Local testing frame)}
Let ${\mathcal A}^{\pi}[{\mathcal D}^{\mathcal {K}}_{\sigma}]$ be a
learning trace, $\tau$ a convergence or error threshold and $\Lambda$
{\em (}resp. $\Gamma${\em )} a set of proximity criteria {\em (}resp. anchoring
strategies{\em )}. We say that the collection
\begin{equation}
\label{eq-local-testing-frame}
\mathcal{L}^{\Lambda}_{\Gamma}[\mathcal{A}^{\pi}[\mathcal{D}^{\mathcal{K}}_{\sigma}],\tau] := \{[\hat{\mathcal{A}}^{\pi}[\mathcal{D}^{\mathcal{K}}_{\sigma}], \tau, \mathcal{H}], \mbox{
  such that } \; \hat{} \in \Gamma
\mbox{ and } \mathcal{H} \in \Lambda\}
\end{equation}
\noindent is a {\em local testing frame} iff exists ${\mathcal F} \in
\Lambda$ which is the fastest on it.
\end{df}

Intuitively, we are talking about sets of items only distinguishable
by the anchoring and/or proximity condition, i.e. by the approximation
technique considered. As $\emptyset \in \Gamma$, any local testing
frame
$\mathcal{L}^{\Lambda}_{\Gamma}[\mathcal{A}^{\pi}[\mathcal{D}^{\mathcal{K}}_{\sigma}],
  \tau]$ includes the anchor-free runs $\mathcal{E}^{\mathcal H} :=
[\mathcal{A}^{\pi}[\mathcal{D}^{\mathcal{K}}_{\sigma}], \tau,
  \mathcal{H}]$ whatever $\mathcal{H} \in \Lambda$. We baptize
${\mathcal E}^{\mathcal F} :=
[\mathcal{A}^{\pi}[\mathcal{D}^{\mathcal{K}}_{\sigma}], \tau,
  \mathcal{F}]$, the one using the fastest proximity criterion, as the
{\em baseline run}. Since the anchors decelerate the convergence,
their absence automatically increases its speed, depending on the
proximity criteria used. That way, from a computational viewpoint, the
baseline is the most efficient learning configuration in a local
testing frame.

\subsection{Performance metric}

According to the principle of \textit{maximum expected utility} ({\sc
  meu})~\citep{Meek:2002:LSM:944790.944798}, we interpret the
performance associated with a run as the search for a satisfactory
cost/benefit trade-off. In that regard, any estimate of such
performance requires the prior formalization of the concepts of cost
and benefit for a run within its local testing frame, i.e. within its
referential context. At this point, since we are interested in
studying the behaviour of different anchorages and/or proximity
conditions through a collection of local testing frames, it will 
be necessary to resort to measures relative to the baseline
runs.

\subsubsection{Cost of a run}

The effort of convergence for a run identifies with its {\sc cl}evel,
provided it may be expressed in terms of the number of iterations
needed to attain the degree of refinement required. We can do this by
considering the same computational reference and threshold in all
runs compared, as occurs within a local testing frame
$\mathcal{L}^{\Lambda}_{\Gamma}[\mathcal{A}^{\pi}[\mathcal{D}^{\mathcal{K}}_{\sigma}],
  \tau]$, assuming that the costs associated to the anchoring (resp.
proximity condition) itself are comparable for all $\; \hat{} \in
\Gamma$ (resp. ${\mathcal H} \in \Lambda$). This also provides a
simple way for normalizing the cost associated with a run, taking the
baseline as a benchmark.

\begin{df}
\label{def-relative-cost-run} {\em (Relative cost)}
Let $\hat{\mathcal E}^{\mathcal
  H} \in \mathcal{L}^{\Lambda}_{\Gamma}[{\mathcal
  A}^\pi[\mathcal{D}^{\mathcal{K}}_{\sigma}], \tau]$ be a run in a
  local testing frame of baseline ${\mathcal E}^{\mathcal
  F}$. We define its {\em relative cost} as
\begin{equation}
\label{eq-relative-cost-run}
\mbox{\sc rc}(\hat{\mathcal E}^{\mathcal H}) := \frac{\mbox{\em {\sc
      cl}evel}[{\mathcal E}^{\mathcal H}]}{\mbox{\em {\sc
      cl}evel}[\hat{\mathcal E}^{\mathcal F}]} \in [1,\infty)
\end{equation}
\end{df}



The {\sc rc} is positive and greater the greater the number of epochs
needed to converge, i.e. the faster the limit curve is approximated the
more its value is reduced, which allows
Theorem~\ref{th-anchoring-trace-comparison} to be interpreted in terms
of computational costs. That way, its minimum is 1 and is reached when
the cost is that of the baseline, thus justifying our interest in
{\sc rc}{\footnotesize s} as close as possible to the unit. However, a
low cost is not enough to conclude the advisability of using absolute
thresholds against relative ones. Unless the specifications explicitly
require one or the other, such a decision should be the consequence of
balancing costs and benefits.


\subsubsection{Performance of a run}

Understood as the balance between benefit and cost, the performance of
a run in the context of its local test frame is assimilable
to the ratio between the degree of accuracy achieved and the relative cost
accumulated during the learning process. With this aim in mind, we still
have to formalize the concept of \textit{accuracy}. If we refer to a
convergence (resp. error) threshold $\tau$, this should be higher the
better the fit of the latter to the difference between the curve
${\mathcal A}_{\infty}^\pi[{\mathcal D}^{\mathcal{K}}_{\sigma}]$ to
which we converge (resp. the learning curve
${\mathcal A}_{\dinfty{}}[{\mathcal D}]$) and the converging learning
trend, which is used to provide an estimate at which the proximity
condition ${\mathcal H}$ is verified, as indicated in
Theorem~\ref{th-completeness-traces}.

\begin{df}
\label{def-accuracy-run} {\em (Convergence and error accuracy)}
Let $\hat{\mathcal E}^{\mathcal H} \in
\mathcal{L}^{\Lambda}_{\Gamma}[\hat{\mathcal
    A}^\pi[\mathcal{D}^{\mathcal{K}}_{\sigma}], \tau]$ be a run with
        {\sc cl}{\em evel} $\ell$. We define its {\em convergence} {\em (}resp. {\em error)} {\em accuracy} as
\begin{equation}
\label{eq-accuracy-run}
\mbox{\sc a}^c(\hat{\mathcal E}^{\mathcal H}) \;\; (\mbox{resp. }
\mbox{\sc a}^e(\hat{\mathcal E}^{\mathcal H})) :=
\left\{ \begin{array}{ll} 0 & \mbox{if } \exists \; i \geq \iota,
  \; \delta^c_\ell(i) \;\; (\mbox{resp. }
  \delta^e_\ell(i)) > \tau \\ 100 \ast \frac{\max_{i \geq \iota}
    \{\abs{\delta^c_\ell(i)} \;\; (\mbox{resp. }
    \abs{\delta^e_\ell(i)})\}}{\tau} & \mbox{otherwise}
        \end{array}
\right.
\end{equation}
with $\iota$ the threshold level for $\tau$, and $\delta^c_\ell(i) \;\; \mbox{\em (resp. }\delta^e_\ell(i)\mbox{\em )}:=
\abs{[\hat{\mathcal A}_{\infty}^\pi[{\mathcal D}^{\mathcal{K}}_{\sigma}] \;\; \mbox{\em (resp. }{\mathcal
  A}_{\dinfty{}}[{\mathcal D}]\mbox{\em )} -
    \hat{\mathcal A}_{\ell}^\pi[{\mathcal
        D}^{\mathcal{K}}_{\sigma}]](\absd{\mathcal D_i})}$ the {\em
  divergence of} $\hat{\mathcal A}_{\ell}^\pi[{\mathcal
    D}^{\mathcal{K}}_{\sigma}]$ {\em with respect to} $\hat{\mathcal
  A}_{\infty}^\pi[{\mathcal D}^{\mathcal{K}}_{\sigma}]$ {\em (resp.} ${\mathcal
  A}_{\dinfty{}}[{\mathcal D}]${\em )} {\em at level}
$i$.

\end{df}

\begin{figure*}[htbp]
\begin{center}
\begin{tabular}{cc}
\hspace*{-.4cm}
\includegraphics[width=0.49\textwidth]{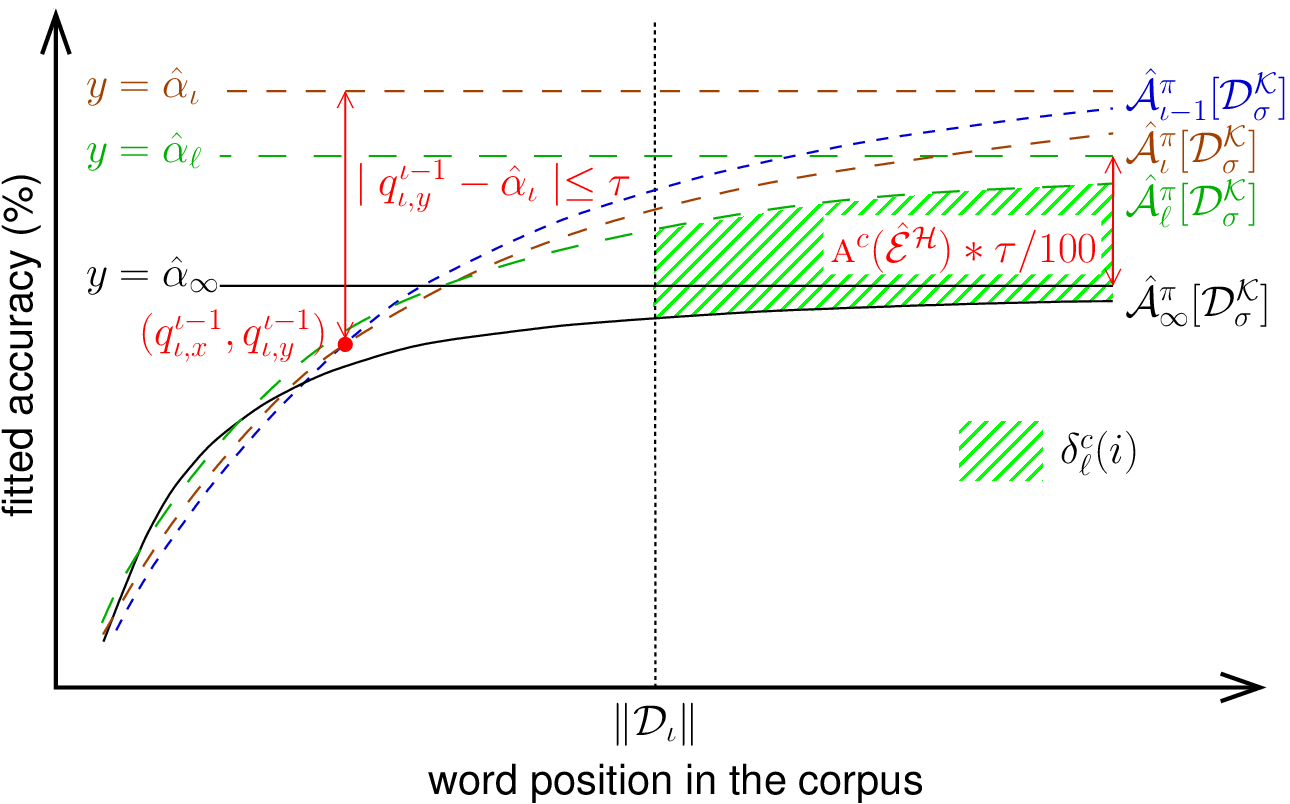}
& 
\hspace*{-.5cm}
\includegraphics[width=0.49\textwidth]{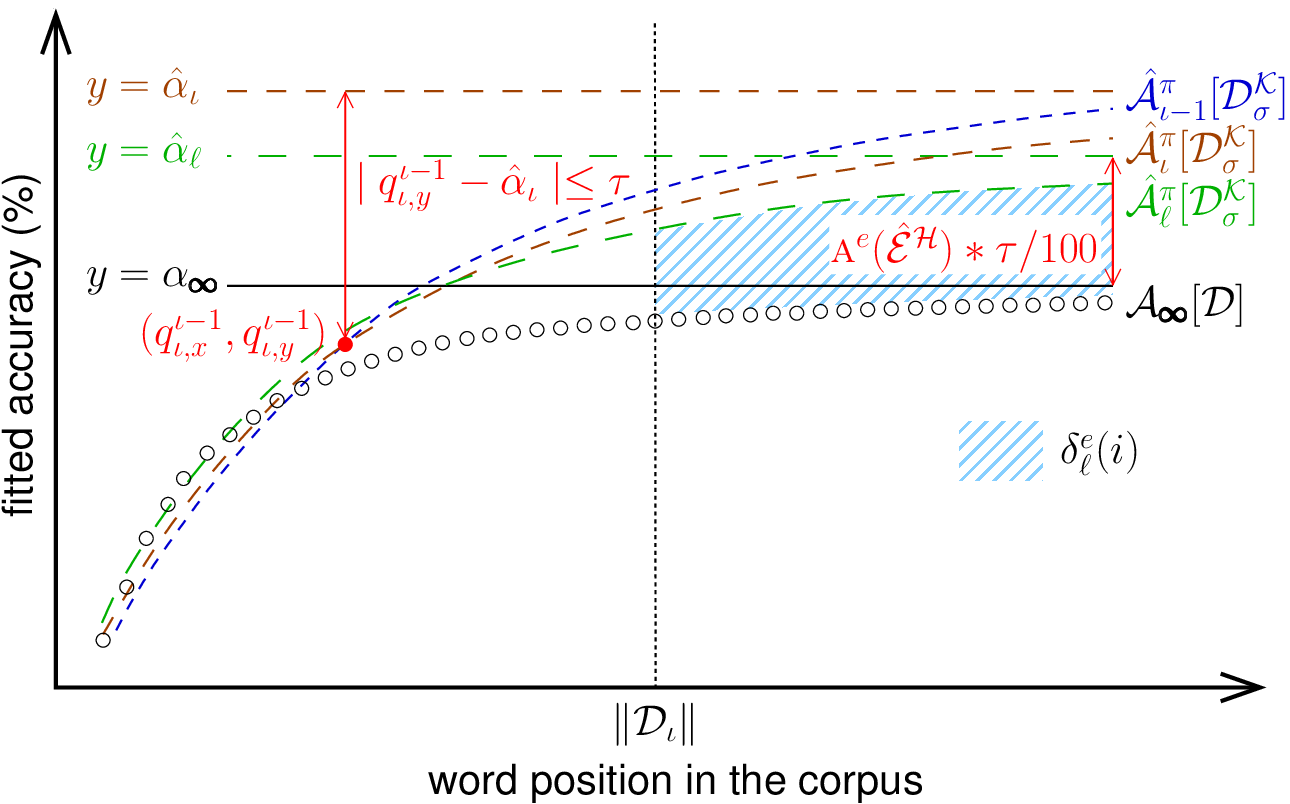}
\end{tabular}
\caption{Computing the convergence (resp. error) accuracy for a
  convergence (resp. error) threshold $\tau$.}
\label{fig-convergence-and-error-accuracy}
\end{center}
\end{figure*}

Thus defined as a percentage, the accuracy corresponds to the
intuitive concept, which justifies our choice for the name of these
metrics. Indeed, we calculate $\mbox{\sc a}^c(\hat{\mathcal
  E}^{\mathcal H})$ (resp. $\mbox{\sc a}^e(\hat{\mathcal E}^{\mathcal
  H})$) as the degree of precision achieved by run $\hat{\mathcal
  E}^{\mathcal H}$ at its {\sc cl}evel in the estimation of the
convergence (resp. error) threshold $\tau$, starting from the
iteration $\iota$ delimiting the interval for the completeness
condition described in Theorem~\ref{th-completeness-traces}. That way,
the value is zero when the threshold $\tau$ is not reached in that
interval. For ease of understanding, the calculation process is
illustrated in the left (resp. right) diagram of
Fig.~\ref{fig-convergence-and-error-accuracy} for the convergence
(resp. error) accuracy. It can be seen that the threshold level
$\iota$ from which we search for the maximum value for the divergence
$\delta_\ell^c$ (resp.  $\delta_\ell^e$) on the learning trend
$\hat{\mathcal A}_{\ell}^\pi[{\mathcal D}^{\mathcal{K}}_{\sigma}]$
associated with {\sc cl}evel $\ell$, and which in the case under
consideration would be reached at its asymptote. The figure also shows
the point $(q_{\iota,x}^{\iota-1},q_{\iota,y}^{\iota-1})$, which marks
the beginning of the domain of completeness for the threshold $\tau$.
However, to calculate this accuracy measure we need to know the curve
$\hat{\mathcal A}_{\infty}^\pi[{\mathcal D}^{\mathcal{K}}_{\sigma}]$
(resp. ${\mathcal A}_{\dinfty{}}[{\mathcal D}]$). To address this
issue, we assume a large enough set of observations provided by an
omniscient oracle for the learning curve ${\mathcal
  A}_{\dinfty{}}[{\mathcal D}]$ through a sequence of contiguous
individuals including the kernel ${\mathcal K}$. Henceforth, we refer
to this set as the \textit{horizon} of the learning trace ${\mathcal
  A}^\pi[\mathcal{D}^{\mathcal{K}}_{\sigma}]$. From this,
$\hat{\mathcal A}_{\infty}^\pi[{\mathcal D}^{\mathcal{K}}_{\sigma}]$
(resp. ${\mathcal A}_{\dinfty{}}[{\mathcal D}]$) is estimated by the
learning trend approximating such a set (resp. by such a set) of
observations, together with its asymptotic value $\hat\alpha_\infty$
(resp. $\alpha_{\dinfty{}}$). Note that, following
Theorem~\ref{th-correctness-trace}, $\alpha_{\dinfty{}} =
\hat\alpha_\infty$. According to this, the calculation of error
accuracy will be made assuming in each case the same asymptotic value
for the set of observations as for the approximation considered of
$\hat{\mathcal A}_{\infty}^\pi[{\mathcal D}^{\mathcal{K}}_{\sigma}]$.

In practice, although we are interested in high values, it should be
borne in mind that low accuracy does not necessarily imply a poor
approximation process. Since the computationally more efficient runs
are associated with larger convergence (resp. error) distances between
epochs, it is more likely that it is in those runs where the
divergences from the limit (resp. learning) curve will be smaller. In
other words, runs with small {\sc rc}{\footnotesize s} could
eventually reach low accuracies, thus justifying the need for
formalizing the concept of performance. As for accuracy, we
differentiate between convergence and error performances to refer to
the approximations of the limit and learning curves, respectively.



\begin{df}
  \label{def-relative-performance-run}
        {\em (Relative convergence and error performance)} Let
        $\hat{\mathcal E}^{\mathcal H} \in
        \mathcal{L}^{\Lambda}_{\Gamma}[{\mathcal
            A}^\pi[\mathcal{D}^{\mathcal{K}}_{\sigma}], \tau]$ be. We
        define its {\em relative convergence} {\em (}resp. {\em
          relative error)} {\em performance} as
\begin{equation}
\label{eq-relative-performance-run}
\mbox{\sc rp}^c(\hat{\mathcal E}^{\mathcal H}) \;\; (\mbox{resp. }\mbox{\sc rp}^e(\hat{\mathcal E}^{\mathcal H})) := \frac{\mbox{\sc a}^c(\hat{\mathcal E}^{\mathcal H}) \;\; (\mbox{resp. } \mbox{\sc a}^e(\hat{\mathcal E}^{\mathcal H}))}{\mbox{\sc rc}(\hat{\mathcal E}^{\mathcal H})} \in [0,100]
\end{equation}
\end{df}



Intuitively, the lower the {\sc rp} in either of its 
interpretations, the more we could argue that an alternative
approximation strategy should be considered. We are therefore
interested in {\sc rp}{\footnotesize s} as close as possible to 100.


\section{The experiments}
\label{section-experiments}

As said, they focus on learners for {\sc ml}-based tagger generation,
a demanding task in {\sc nlp}. We thus need to introduce the
linguistic resources and the testing space.

\subsection{The linguistics resources}

Corpora and {\sc pos} tagger generators are selected from the most
popular ones, as training data and learners respectively, the
former together with their tag-sets:

\begin{enumerate}
\item The section of the {\sc wsj} in the {\sc p}enn {\sc
  t}reebank~\citep{Marcus1999}, of over 1,170,000 words.


\item The Freiburg-Brown ({\sc f}rown) of American
  English~\citep{Hinrichs2010}, of over 1,165,000 words.
\end{enumerate}

\noindent {\sc p}enn is annotated with {\sc pos} tags and
syntactic structures. By stripping it of the latter, it can be used to
train {\sc pos} taggers. To ensure well-balanced
corpora, we have scrambled them at sentence level before testing.

We focus on models built from supervised learning, which make it
possible to work with predefined tag-sets, thereby facilitating both
the evaluation and the comprehension of the results:

\begin{enumerate}

\item In the category of stochastic methods and representing the
  \textit{hidden M\'arkov models} ({\sc hmm}s), we chose {\sc t}n{\sc
    t}~\citep{Brants2000}. We also include the {\sc t}ree{\sc
    t}agger~\citep{Schmid1994}, which uses decision trees to generate
  the {\sc hmm}, and {\sc m}orfette~\citep{Chrupala2008}, an averaged
  perceptron approach~\citep{Collins2002}. To illustrate the
  \textit{maximum entropy models} ({\sc mem}s), {\sc
    mxpost}~\citep{Ratnaparki1996} and {\sc o}pen{\sc nlp} {\sc
    m}ax{\sc e}nt~\citep{Toutanova2003}. Finally, the \textsc{s}tanford
  {\sc pos} tagger~\citep{Toutanova2003} combines features of {\sc
    hmm}s and {\sc mem}s using a \textit{conditional M\'arkov model}.

\item Under the heading of other approaches we consider fn{\sc
    tbl}~\citep{Ngai2001}, an update of {\sc
    b}rill~\citep{Brill1995a}, as example of transformation-based
  learning. As memory-based method we take the
  \textit{memory-based tagger} ({\sc mbt})~\citep{Daelemans1996a},
  while {\sc svmt}ool~\citep{Gimenez2004} illustrates the behaviour
  with respect to support vector machines. We also use a
  perceptron-based training strategy with look-ahead, {\sc
    lapos}~\citep{Tsuruoka2011}.
\end{enumerate}
\noindent This all ensures a good coverage of the linguistic resources
for testing our proposal.

\subsection{The testing space}

We consider a collection ${\mathcal L}$ of local testing frames, with
an entry for each combination of corpus and tagger. For each of these
structures ${\mathcal
  L}^{\Lambda}_{\Gamma}[\mathcal{A}^{\pi}[\mathcal{D}^{\mathcal{K}}_{\sigma}],\tau]$,
the size of the kernel $\mathcal{K}$ and the step function $\sigma$
are fixed to $5*10^3$, while a power law family parameterizable by the
trust region method~\citep{Branch1999} is chosen as accuracy pattern
$\pi$. The {\sc wl}evel of the runs is calculated from the values
proposed in~\citep{VilaresDarribaRibadas16}: $\nu=2*10^{-5}$,
$\varsigma=1$ and $\lambda=5$. With respect to $\Gamma$, it includes
both canonical and fixed anchoring. The proximity conditions are taken
from $\Lambda = \{\mathcal{H}_a,\mathcal{H}_r\}$, as defined
above. Since the asymptotic backbone of any of the runs concerned is
decreasing, the applicability of $\mathcal{H}_a$ is guaranteed and
these local testing frames are well-defined. Whatever the learning trace
$\mathcal{A}^{\pi}[\mathcal{D}^{\mathcal{K}}_{\sigma}]$, we will
consider as reference for testing purposes a horizon of $160$ real
observations provided by an omniscient oracle.


Having defined the testing structure, we address three aspects
supporting the significance of the trials. The first relates to the
exploitation of the training resources. Thus, as phrases are the
smallest grammatical units with specific sense, samples should be
aligned to the sentential distribution of the text. The second
concerns the utility of the generated models, which depends on both
the quality metrics being well-defined within the scope of the corpora
and the reduction of variability phenomena. Finally, we tackle model
optimization, i.e. the anchoring setting in each run.


\subsubsection{Sampling fitting to sentence level and stability}


We then need to adapt the learning schema. Given a corpus $\mathcal
D$ with kernel $\mathcal K$ and a step function $\sigma$, we build the
individuals $\{\mathcal D_i\}_{i \in \mathbb{N}}$ with ${\mathcal D}_i
:= \sentences{{\mathcal W}_i}$ such that
\begin{equation}
\begin{array}{l}
{\mathcal W}_1 := {\mathcal K} \mbox{ and } {\mathcal W}_i :=
{\mathcal W}_{i-1} \cup {\mathcal I}_{i}, \; \mathcal I_i \subset
{\mathcal D} \setminus {\mathcal W}_{i-1}, \; \absd{{\mathcal I}_{i}}
:= \sigma(i), \; \forall i \geq 2
\end{array}
\end{equation}
\noindent where $\sentences{{\mathcal W}_{i}}$ denotes the minimal set
of sentences including the words in ${\mathcal W}_{i}$. This has no
impact on our foundations and allows us to reap the maximum from
training. Following~\citep{Daelemans1996a,Giesbrecht2009},
$10$-fold cross validation confers stability on our measures.

%

\subsubsection{Parameter tuning}

As the speed of convergence relies on the anchoring used, fine-tuning
is required to select a configuration close to the most efficient one
and provide credibility to the tests when the strategy is
parameterizable, which is what happens with fixed anchors. Given a
local testing frame
$\mathcal{L}^{\Lambda}_{\Gamma}[{\mathcal
  A}^\pi[\mathcal{D}^{\mathcal{K}}_{\sigma}],\tau]$, the way to do it
is by choosing the optimal look-ahead for a value 100, namely the one
with lowest {\sc rc}. Moreover, as we will see, it is only necessary
to focus on runs using the absolute proximity condition
${\mathcal H}_a$.

To this end, the potential look-aheads are studied in order of
increasing size, which is the same as saying that they are explored
according to their corresponding {\sc put}, in decreasing order. For a
uniform and complete monitoring of the procedure, its codomain [0,100]
is covered with step 10. We then compare the {\sc rc} of the runs
using the minimal look-aheads corresponding to such a sequence of
tentative {\sc put}{\footnotesize s}.

Given $\zeta$ a tentative {\sc put} for a run
$[{\mathcal A}^{\pi}[{\mathcal D}^{\mathcal {K}}_{\sigma}],\tau,
{\mathcal H}_a]$, a candidate
$[\hat{\mathcal A}^{\stackrel{100,\iota}{\pi}}[{\mathcal D}^{\mathcal
  {K}}_{\sigma}], \tau, {\mathcal H}_a]$ is generated from its minimal
look-ahead $\iota$ and the {\sc rc} calculated. The process is repeated
for the next {\sc put} until we locate the lowest {\sc rc}. Hence, it
is enough to identify the run associated to the turning point in such
a {\sc rc} sequence because, with respect to an increasing look-ahead,
the performance is also increasing until reaching its maximum and then
begins a decreasing trend. On this basis and to reduce the impact of
irregularities, we chose that from which the window of increasing {\sc
  rc}{\footnotesize s} is the largest one. So, it is hoped that the
look-ahead is optimal with an error margin of 10\% regarding the {\sc
  put} metric.

\subsection{The testing strategy}

We do so according to the goals of our testing frame. That way, the
cost of ensuring the completeness for $\mathcal{H}_a$ via fixed anchors
is given by comparing within ${\mathcal L}$ the {\sc rc} then
applicable and the one estimated for similar runs when no anchoring or
alternative canonical technique are used. These runs through
${\mathcal L}$ are hereafter referred to as ${\mathcal L}_{\mbox{\sc rc}}$ and do
not include the baseline ones.

To balance costs and benefits of ${\mathcal H}_a$ against ${\mathcal
  H}_r$, we contrast the {\sc rp} in its two interpretations --
$\mbox{\sc rp}^c$ and $\mbox{\sc rp}^e$ -- when applying the former on
a run in ${\mathcal L}$ with fixed anchors, and when no or canonical
anchors is used for its dissimilar ones. We exclude the dissimilar run
with fixed anchor because Theorem~\ref{th-anchoring-trace-comparison}
makes it clear that the only practical interest of this anchoring
technique is to provide completeness for ${\mathcal H}_a$. Such runs
through ${\mathcal L}$ are hereafter referred to as ${\mathcal
  L}_{\mbox{\sc rp}}$, including the baseline ones.

\begin{table*}[htbp]
\begin{center}
\caption{{\sc rc} monitoring in ${\mathcal L}_{\mbox{\sc rc}}$}
\label{table-runs-RC}
\begin{scriptsize}
\begin{tabular}{@{\hspace{0pt}}l@{\hspace{3pt}}l@{\hspace{2pt}}l@{\hspace{2pt}}
                c@{\hspace{0pt}}c@{\hspace{3pt}}  
                c@{\hspace{0pt}}c@{\hspace{3pt}}  
                c@{\hspace{0pt}}c@{\hspace{3pt}}  
                c@{\hspace{0pt}}c@{\hspace{3pt}}  
                c@{\hspace{0pt}}c@{\hspace{3pt}}  
                c@{\hspace{0pt}}c@{\hspace{3pt}}  
                c@{\hspace{0pt}}c@{\hspace{3pt}}  
                c@{\hspace{0pt}}c@{\hspace{3pt}}  
                c@{\hspace{0pt}}c@{\hspace{3pt}}  
                c@{\hspace{0pt}}c@{\hspace{3pt}}  
                c@{\hspace{0pt}}c@{\hspace{3pt}}  
                c@{\hspace{0pt}}}                 

\hline
& & & \boldmath$\tau$ & & \multicolumn{5}{c}{{\bf No anchor} \boldmath$+\;{\mathcal H}_a$} & & \multicolumn{5}{c}{{\bf Canonical} \boldmath$+\;{\mathcal H}_a$} & & \multicolumn{9}{c}{{\bf Fixed} \boldmath$+\;{\mathcal H}_a$} \rule{0pt}{2.5ex} \\

\cline{1-2} \cline{4-4} \cline{6-10} \cline{12-16} \cline{18-26}

& & & & & \bf{\scshape pl}evel & & \bf{\scshape cl}evel & & \bf{\scshape rc} & & \bf{\scshape pl}evel & & \bf{\scshape cl}evel & & \bf{\scshape rc} & & \bf{\scshape put} & & \boldmath$\ell_o$ & & \bf{\scshape pl}evel & & \bf{\scshape cl}evel & & \bf{\scshape rc} \rule{0pt}{2.5ex} \\

\cline{1-2} \cline{4-4} \cline{6-6} \cline{8-8} \cline{10-10} \cline{12-12} \cline{14-14} \cline{16-16} \cline{18-18} \cline{20-20} \cline{22-22} \cline{24-24} \cline{26-26}

\multirow{9}{*}{\begin{sideways}{\bf \textsc{f}rown}\end{sideways}} & fn\textsc{tbl} & & 1.50 & & \bf\em 55 & & \bf\em 58 & & \bf\em 1.00 & &  59 & &  77 & &  1.33 & &  89.38 & &  6 & &  55 & &  100 & &  1.72\rule{0pt}{2.75ex} \\
& \textsc{lapos} & & 1.27 & & \bf\em 18 & & \bf\em 46 & & \bf\em 1.00 & &  18 & &  71 & &  1.54 & &  64.86 & &  18 & &  18 & &  80 & &  1.74\rule{0pt}{2.25ex} \\
& \textsc{m}ax\textsc{e}nt & & 1.70 & & \bf\em 32 & & \bf\em 49 & & \bf\em 1.00 & &  32 & &  83 & &  1.69 & &  16.83 & &  52 & &  32 & &  94 & &  1.92\rule{0pt}{2.25ex} \\
& \textsc{mbt} & & 1.95 & & \bf\em 43 & & \bf\em 51 & & \bf\em 1.00 & &  51 & &  86 & &  1.69 & &  29.40 & &  43 & &  43 & &  98 & &  1.92\rule{0pt}{2.25ex} \\
& \textsc{m}orfette & & 1.43 & & \bf\em 20 & & \bf\em 48 & & \bf\em 1.00 & &  20 & &  75 & &  1.56 & &  69.71 & &  19 & &  20 & &  95 & &  1.98\rule{0pt}{2.25ex} \\
& \textsc{mxpost} & & 2.84 & & \bf\em 22 & & \bf\em 30 & & \bf\em 1.00 & &  22 & &  31 & &  1.03 & &  75.63 & &  11 & &  22 & &  59 & &  1.97\rule{0pt}{2.25ex} \\
& \textsc{s}tanford & & 1.91 & & \bf\em 24 & & \bf\em 36 & & \bf\em 1.00 & &  29 & &  72 & &  2.00 & &  78.29 & &  12 & &  24 & &  82 & &  2.28\rule{0pt}{2.25ex} \\
& \textsc{svmt}ool & & 1.41 & & \bf\em 41 & & \bf\em 52 & & \bf\em 1.00 & &  46 & &  89 & &  1.71 & &  76.02 & &  10 & &  41 & &  92 & &  1.77\rule{0pt}{2.25ex} \\
& \textsc{t}n\textsc{t} & & 1.51 & & \bf\em 19 & & \bf\em 41 & & \bf\em 1.00 & &  19 & &  73 & &  1.78 & &  45.41 & &  32 & &  19 & &  86 & &  2.10\rule{0pt}{2.25ex} \\
\cline{1-2} \cline{4-4} \cline{6-6} \cline{8-8} \cline{10-10} \cline{12-12} \cline{14-14} \cline{16-16} \cline{18-18} \cline{20-20} \cline{22-22} \cline{24-24} \cline{26-26}
\multirow{3}{*}{\begin{sideways}{\bf \textsc{p}enn}\end{sideways}} & \textsc{mbt} & & 1.66 & & \bf\em 15 & & \bf\em 39 & & \bf\em 1.00 & &  15 & &  85 & &  2.18 & &  39.50 & &  44 & &  15 & &  98 & &  2.51\rule{0pt}{2.75ex} \\
& \textsc{mxpost} & & 1.40 & & \bf\em 17 & & \bf\em 28 & & \bf\em 1.00 & &  17 & &  50 & &  1.79 & &  18.73 & &  37 & &  17 & &  57 & &  2.04\rule{0pt}{2.25ex} \\
& \textsc{svmt}ool & & 1.25 & & \bf\em 26 & & \bf\em 31 & & \bf\em 1.00 & &  26 & &  66 & &  2.13 & &  29.69 & &  47 & &  26 & &  87 & &  2.81\rule{0pt}{2.25ex} \\

\hline
\end{tabular}
\end{scriptsize}
\end{center}
\end{table*}

\begin{table*}[htbp]
\begin{center}
\caption{$\mbox{\sc rp}^c$ monitoring in ${\mathcal L}_{\mbox{\sc rp}}$}
\label{table-runs-convergence-RP}

\begin{scriptsize}
\begin{tabular}{@{\hspace{0pt}}l@{\hspace{3pt}}l@{\hspace{2pt}}l@{\hspace{2pt}}
                c@{\hspace{0pt}}c@{\hspace{3pt}}  
                c@{\hspace{0pt}}c@{\hspace{3pt}}  
                c@{\hspace{0pt}}c@{\hspace{3pt}}  
                c@{\hspace{0pt}}c@{\hspace{3pt}}  
                c@{\hspace{0pt}}c@{\hspace{3pt}}  
                c@{\hspace{0pt}}c@{\hspace{3pt}}  
                c@{\hspace{0pt}}c@{\hspace{3pt}}  
                c@{\hspace{0pt}}c@{\hspace{3pt}}  
                c@{\hspace{0pt}}c@{\hspace{3pt}}  
                c@{\hspace{0pt}}c@{\hspace{3pt}}  
                c@{\hspace{0pt}}c@{\hspace{3pt}}  
                c@{\hspace{0pt}}c@{\hspace{3pt}}  
                c@{\hspace{0pt}}c@{\hspace{3pt}}  
                c@{\hspace{0pt}}c@{\hspace{3pt}}  
                c@{\hspace{0pt}}c@{\hspace{3pt}}  
                c@{\hspace{0pt}}c@{\hspace{3pt}}  
                c@{\hspace{0pt}}c@{\hspace{3pt}}  
                c@{\hspace{0pt}}}                 

\hline
& & & \boldmath$\tau$ & & \multicolumn{9}{c}{{\bf No anchor} \boldmath$+\;{\mathcal H}_r$} & & \multicolumn{9}{c}{{\bf Canonical} \boldmath$+\;{\mathcal H}_r$} & & \multicolumn{13}{c}{{\bf Fixed} \boldmath$+\;{\mathcal H}_a$} \rule{0pt}{2.5ex} \\

\cline{1-2} \cline{4-4} \cline{6-14} \cline{16-24} \cline{26-38}

& & & & & \bf{\scshape pl}evel & & \bf{\scshape cl}evel & & \boldmath$\mbox{\bf{\scshape a}}^c$ & & \bf{\scshape rc} & & \boldmath$\mbox{\bf{\scshape rp}}^c$ & & \bf{\scshape pl}evel & & \bf{\scshape cl}evel & & \boldmath$\mbox{\bf{\scshape a}}^c$ & & \bf{\scshape rc} & & \boldmath$\mbox{\bf{\scshape rp}}^c$ & & \bf{\scshape put} & & \boldmath$\ell_o$ & & \bf{\scshape pl}evel & & \bf{\scshape cl}evel & & \boldmath$\mbox{\bf{\scshape a}}^c$ & & \bf{\scshape rc} & & \boldmath$\mbox{\bf{\scshape rp}}^c$ \rule{0pt}{2.5ex} \\

\cline{1-2} \cline{4-4} \cline{6-6} \cline{8-8} \cline{10-10} \cline{12-12} \cline{14-14} \cline{16-16} \cline{18-18} \cline{20-20} \cline{22-22} \cline{24-24} \cline{26-26} \cline{28-28} \cline{30-30} \cline{32-32} \cline{34-34} \cline{36-36} \cline{38-38}  

\multirow{9}{*}{\begin{sideways}{\bf \textsc{f}rown}\end{sideways}} & fn\textsc{tbl} & & 4.26 & & \em 55 & & \em 58 & & \em 14.39 & & \em 1.00 & & \em 14.39 & & \bf 59 & & \bf 60 & & \bf 16.42 & & \bf 1.03 & & \bf 15.88 & &  89.38 & &  6 & &  55 & &  100 & &  \underline{4.24} & &  1.72 & &  \underline{2.46}\rule{0pt}{2.75ex} \\
& \textsc{lapos} & & 3.47 & & \em 18 & & \em 46 & & \em 6.15 & & \em 1.00 & & \em 6.15 & & \bf 18 & & \bf 46 & & \bf 6.88 & & \bf 1.00 & & \bf 6.88 & &  64.86 & &  18 & &  18 & &  80 & &  8.38 & &  1.74 & &  4.82\rule{0pt}{2.25ex} \\
& \textsc{m}ax\textsc{e}nt & & 4.42 & & \em 32 & & \em 49 & & \em 4.03 & & \em 1.00 & & \em 4.03 & & \bf 32 & & \bf 50 & & \bf 6.55 & & \bf 1.02 & & \bf 6.42 & &  16.83 & &  52 & &  32 & &  94 & &  \underline{3.02} & &  1.92 & &  \underline{1.57}\rule{0pt}{2.25ex} \\
& \textsc{mbt} & & 4.78 & & \em 43 & & \em 51 & & \em 3.02 & & \em 1.00 & & \em 3.02 & & \bf 51 & & \bf 52 & & \bf 10.66 & & \bf 1.02 & & \bf 10.45 & &  29.40 & &  43 & &  43 & &  98 & &  \underline{1.11} & &  1.92 & &  \underline{0.58}\rule{0pt}{2.25ex} \\
& \textsc{m}orfette & & 3.67 & & \em 20 & & \em 48 & & \em 6.02 & & \em 1.00 & & \em 6.02 & & \bf 20 & & \bf 48 & & \bf 6.83 & & \bf 1.00 & & \bf 6.83 & &  69.71 & &  19 & &  20 & &  95 & &  6.98 & &  1.98 & &  3.53\rule{0pt}{2.25ex} \\
& \textsc{mxpost} & & 6.28 & & \em 22 & & \em 30 & & \em \underline{1.81} & & \em 1.00 & & \em \underline{1.81} & &  22 & &  30 & &  \underline{0.85} & &  1.00 & &  \underline{0.85} & & \bf 75.63 & & \bf 11 & & \bf 22 & & \bf 59 & & \bf 6.73 & & \bf 1.97 & & \bf 3.42\rule{0pt}{2.25ex} \\
& \textsc{s}tanford & & 4.31 & & \em 24 & & \em 36 & & \em 7.88 & & \em 1.00 & & \em 7.88 & & \bf 29 & & \bf 38 & & \bf 16.01 & & \bf 1.06 & & \bf 15.16 & &  78.29 & &  12 & &  24 & &  82 & &  5.33 & &  2.28 & &  2.34\rule{0pt}{2.25ex} \\
& \textsc{svmt}ool & & 3.82 & & \em 41 & & \em 52 & & \em 13.89 & & \em 1.00 & & \em 13.89 & & \bf 46 & & \bf 53 & & \bf 16.50 & & \bf 1.02 & & \bf 16.19 & &  76.02 & &  10 & &  41 & &  92 & &  7.09 & &  1.77 & &  4.01\rule{0pt}{2.25ex} \\
& \textsc{t}n\textsc{t} & & 3.69 & & \em 19 & & \em 41 & & \em 5.45 & & \em 1.00 & & \em 5.45 & & \bf 19 & & \bf 43 & & \bf 9.65 & & \bf 1.05 & & \bf 9.20 & &  45.41 & &  32 & &  19 & &  86 & &  5.77 & &  2.10 & &  2.75\rule{0pt}{2.25ex} \\
\cline{1-2} \cline{4-4} \cline{6-6} \cline{8-8} \cline{10-10} \cline{12-12} \cline{14-14} \cline{16-16} \cline{18-18} \cline{20-20} \cline{22-22} \cline{24-24} \cline{26-26} \cline{28-28} \cline{30-30} \cline{32-32} \cline{34-34} \cline{36-36} \cline{38-38}
\multirow{3}{*}{\begin{sideways}{\bf \textsc{p}enn}\end{sideways}} & \textsc{mbt} & & 3.72 & & \bf\em 15 & & \bf\em 39 & & \bf\em 18.89 & & \bf\em 1.00 & & \bf\em 18.89 & &  15 & &  39 & &  \underline{5.96} & &  1.00 & &  \underline{5.96} & &  39.50 & &  44 & &  15 & &  98 & &  12.85 & &  2.51 & &  5.11\rule{0pt}{2.75ex} \\
& \textsc{mxpost} & & 3.34 & & \em 17 & & \em 28 & & \em \underline{2.22} & & \em 1.00 & & \em \underline{2.22} & &  17 & &  29 & &  \underline{4.40} & &  1.04 & &  \underline{4.25} & & \bf 18.73 & & \bf 37 & & \bf 17 & & \bf 57 & & \bf 14.70 & & \bf 2.04 & & \bf 7.22\rule{0pt}{2.25ex} \\
& \textsc{svmt}ool & & 2.64 & & \em 26 & & \em 31 & & \em 23.09 & & \em 1.00 & & \em 23.09 & & \bf 26 & & \bf 35 & & \bf 28.46 & & \bf 1.13 & & \bf 25.20 & &  29.69 & &  47 & &  26 & &  87 & &  22.94 & &  2.81 & &  8.18\rule{0pt}{2.25ex} \\

\hline
\end{tabular}
\end{scriptsize}
\end{center}
\end{table*}


\begin{table*}[htbp]

\begin{center}
\caption{$\mbox{\sc rp}^e$ monitoring in ${\mathcal L}_{\mbox{\sc rp}}$}
\label{table-runs-error-RP}

\begin{scriptsize}
\begin{tabular}{@{\hspace{0pt}}l@{\hspace{3pt}}l@{\hspace{2pt}}l@{\hspace{2pt}}
                c@{\hspace{0pt}}c@{\hspace{3pt}}  
                c@{\hspace{0pt}}c@{\hspace{3pt}}  
                c@{\hspace{0pt}}c@{\hspace{3pt}}  
                c@{\hspace{0pt}}c@{\hspace{3pt}}  
                c@{\hspace{0pt}}c@{\hspace{3pt}}  
                c@{\hspace{0pt}}c@{\hspace{3pt}}  
                c@{\hspace{0pt}}c@{\hspace{3pt}}  
                c@{\hspace{0pt}}c@{\hspace{3pt}}  
                c@{\hspace{0pt}}c@{\hspace{3pt}}  
                c@{\hspace{0pt}}c@{\hspace{3pt}}  
                c@{\hspace{0pt}}c@{\hspace{3pt}}  
                c@{\hspace{0pt}}c@{\hspace{3pt}}  
                c@{\hspace{0pt}}c@{\hspace{3pt}}  
                c@{\hspace{0pt}}c@{\hspace{3pt}}  
                c@{\hspace{0pt}}c@{\hspace{3pt}}  
                c@{\hspace{0pt}}c@{\hspace{3pt}}  
                c@{\hspace{0pt}}c@{\hspace{3pt}}  
                c@{\hspace{0pt}}}                 

\hline
& & & \boldmath$\tau$ & & \multicolumn{9}{c}{{\bf No anchor} \boldmath$+\;{\mathcal H}_r$} & & \multicolumn{9}{c}{{\bf Canonical} \boldmath$+\;{\mathcal H}_r$} & & \multicolumn{13}{c}{{\bf Fixed} \boldmath$+\;{\mathcal H}_a$} \rule{0pt}{2.5ex} \\

\cline{1-2} \cline{4-4} \cline{6-14} \cline{16-24} \cline{26-38}

& & & & & \bf{\scshape pl}evel & & \bf{\scshape cl}evel & & \boldmath$\mbox{\bf{\scshape a}}^e$ & & \bf{\scshape rc} & & \boldmath$\mbox{\bf{\scshape rp}}^e$ & & \bf{\scshape pl}evel & & \bf{\scshape cl}evel & & \boldmath$\mbox{\bf{\scshape a}}^e$ & & \bf{\scshape rc} & & \boldmath$\mbox{\bf{\scshape rp}}^e$ & & \bf{\scshape put} & & \boldmath$\ell_o$ & & \bf{\scshape pl}evel & & \bf{\scshape cl}evel & & \boldmath$\mbox{\bf{\scshape a}}^e$ & & \bf{\scshape rc} & & \boldmath$\mbox{\bf{\scshape rp}}^e$ \rule{0pt}{2.5ex} \\

\cline{1-2} \cline{4-4} \cline{6-6} \cline{8-8} \cline{10-10} \cline{12-12} \cline{14-14} \cline{16-16} \cline{18-18} \cline{20-20} \cline{22-22} \cline{24-24} \cline{26-26} \cline{28-28} \cline{30-30} \cline{32-32} \cline{34-34} \cline{36-36} \cline{38-38}  

\multirow{9}{*}{\begin{sideways}{\bf \textsc{f}rown}\end{sideways}} & fn\textsc{tbl} & & 4.26 & & \em 55 & & \em 58 & & \em 14.39 & & \em 1.00 & & \em 14.39 & & \bf 59 & & \bf 60 & & \bf 16.42 & & \bf 1.03 & & \bf 15.88 & &  89.38 & &  6 & &  55 & &  100 & &  \underline{5.09} & &  1.72 & &  \underline{2.95}\rule{0pt}{2.75ex} \\
& \textsc{lapos} & & 3.47 & & \em 18 & & \em 46 & & \em 6.15 & & \em 1.00 & & \em 6.15 & & \bf 18 & & \bf 46 & & \bf 6.88 & & \bf 1.00 & & \bf 6.88 & &  64.86 & &  18 & &  18 & &  80 & &  8.38 & &  1.74 & &  4.82\rule{0pt}{2.25ex} \\
& \textsc{m}ax\textsc{e}nt & & 4.42 & & \em 32 & & \em 49 & & \em 4.03 & & \em 1.00 & & \em 4.03 & & \bf 32 & & \bf 50 & & \bf 6.55 & & \bf 1.02 & & \bf 6.42 & &  16.83 & &  52 & &  32 & &  94 & &  \underline{5.23} & &  1.92 & &  \underline{2.73}\rule{0pt}{2.25ex} \\
& \textsc{mbt} & & 4.78 & & \em 43 & & \em 51 & & \em 3.02 & & \em 1.00 & & \em 3.02 & & \bf 51 & & \bf 52 & & \bf 10.66 & & \bf 1.02 & & \bf 10.45 & &  29.40 & &  43 & &  43 & &  98 & &  \underline{1.46} & &  1.92 & &  \underline{0.76}\rule{0pt}{2.25ex} \\
& \textsc{m}orfette & & 3.67 & & \em 20 & & \em 48 & & \em 6.02 & & \em 1.00 & & \em 6.02 & & \bf 20 & & \bf 48 & & \bf 6.83 & & \bf 1.00 & & \bf 6.83 & &  69.71 & &  19 & &  20 & &  95 & &  6.98 & &  1.98 & &  3.53\rule{0pt}{2.25ex} \\
& \textsc{mxpost} & & 6.28 & & \em 22 & & \em 30 & & \em \underline{1.91} & & \em 1.00 & & \em \underline{1.91} & &  22 & &  30 & &  \underline{2.62} & &  1.00 & &  \underline{2.62} & & \bf 75.63 & & \bf 11 & & \bf 22 & & \bf 59 & & \bf 6.73 & & \bf 1.97 & & \bf 3.42\rule{0pt}{2.25ex} \\
& \textsc{s}tanford & & 4.31 & & \em 24 & & \em 36 & & \em 7.88 & & \em 1.00 & & \em 7.88 & & \bf 29 & & \bf 38 & & \bf 16.01 & & \bf 1.06 & & \bf 15.16 & &  78.29 & &  12 & &  24 & &  82 & &  5.33 & &  2.28 & &  2.34\rule{0pt}{2.25ex} \\
& \textsc{svmt}ool & & 3.82 & & \em 41 & & \em 52 & & \em 13.89 & & \em 1.00 & & \em 13.89 & & \bf 46 & & \bf 53 & & \bf 16.50 & & \bf 1.02 & & \bf 16.19 & &  76.02 & &  10 & &  41 & &  92 & &  7.09 & &  1.77 & &  4.01\rule{0pt}{2.25ex} \\
& \textsc{t}n\textsc{t} & & 3.69 & & \em 19 & & \em 41 & & \em 5.45 & & \em 1.00 & & \em 5.45 & & \bf 19 & & \bf 43 & & \bf 9.65 & & \bf 1.05 & & \bf 9.20 & &  45.41 & &  32 & &  19 & &  86 & &  5.77 & &  2.10 & &  2.75\rule{0pt}{2.25ex} \\
\cline{1-2} \cline{4-4} \cline{6-6} \cline{8-8} \cline{10-10} \cline{12-12} \cline{14-14} \cline{16-16} \cline{18-18} \cline{20-20} \cline{22-22} \cline{24-24} \cline{26-26} \cline{28-28} \cline{30-30} \cline{32-32} \cline{34-34} \cline{36-36} \cline{38-38}
\multirow{3}{*}{\begin{sideways}{\bf \textsc{p}enn}\end{sideways}} & \textsc{mbt} & & 3.72 & & \bf\em 15 & & \bf\em 39 & & \bf\em 18.89 & & \bf\em 1.00 & & \bf\em 18.89 & &  15 & &  39 & &  \underline{7.80} & &  1.00 & &  \underline{7.80} & &  39.50 & &  44 & &  15 & &  98 & &  12.85 & &  2.51 & &  5.11\rule{0pt}{2.75ex} \\
& \textsc{mxpost} & & 3.34 & & \em 17 & & \em 28 & & \em \underline{5.05} & & \em 1.00 & & \em \underline{5.05} & &  17 & &  29 & &  \underline{4.95} & &  1.04 & &  \underline{4.78} & & \bf 18.73 & & \bf 37 & & \bf 17 & & \bf 57 & & \bf 14.70 & & \bf 2.04 & & \bf 7.22\rule{0pt}{2.25ex} \\
& \textsc{svmt}ool & & 2.64 & & \em 26 & & \em 31 & & \em 23.09 & & \em 1.00 & & \em 23.09 & & \bf 26 & & \bf 35 & & \bf 28.46 & & \bf 1.13 & & \bf 25.20 & &  29.69 & &  47 & &  26 & &  87 & &  22.94 & &  2.81 & &  8.18\rule{0pt}{2.25ex} \\

\hline
\end{tabular}
\end{scriptsize}
\end{center}
\end{table*}

Finally, in order to assess the stability of using ${\mathcal H}_a$
via fixed anchoring against look-ahead variations, we shall simply
extend the set $\Gamma$ of anchorings in each local testing frame
$\mathcal{L}^{\Lambda}_{\Gamma}[{\mathcal
    A}^\pi[\mathcal{D}^{\mathcal{K}}_{\sigma}],\tau]$, to include a
selection of representative look-aheads. The runs involved in this
study through ${\mathcal L}$ will be referred to as ${\mathcal
  L}_{\mbox{\sc rp}_\ell}$.


\subsection{The analysis of the results}

The detail of the monitoring is compiled separately for ${\mathcal
  L}_{\mbox{\sc rc}}$ (resp. ${\mathcal L}_{\mbox{\sc rp}}$ and
${\mathcal L}_{\mbox{\sc rp}_\ell}$) in Table~\ref{table-runs-RC}
(resp. Tables~\ref{table-runs-convergence-RP}-\ref{table-runs-error-RP}
and~\ref{table-runs-augmented-convergence-RP}-\ref{table-runs-augmented-error-RP})
along with its {\sc pl}evel. We also include the {\sc cl}evel for each
run, as well as its {\sc rc} (resp. $\mbox{\sc a}^c$ and $\mbox{\sc
  rp}^c$, and $\mbox{\sc a}^e$ and $\mbox{\sc rp}^e$) value, which is
better the closer it comes to 1 (resp. to 100). When using a fixed
anchoring with value $\beta = 100$, the look-ahead is also visualized,
together with the associated {\sc put} in the case of the optimal
value $\ell_o$. These values are expressed to two decimal places
because of space limitations, using bold (resp. italic) fonts to mark
the best results among all the (resp. baseline) runs, while all the
calculations have been done to six decimal places of precision. Absent
from these tables are the local testing frames whose baselines show
increasing asymptotic backbones, as with fn{\sc tbl}, {\sc lapos},
{\sc m}ax{\sc e}nt, {\sc m}orfette, {\sc s}tanford, {\sc t}n{\sc t}
and {\sc t}ree{\sc t}agger trained on {\sc p}enn. We also discard
those with any run converging beyond the boundaries of the training
corpora, as with {\sc t}ree{\sc t}agger on {\sc f}rown.

%
%


\subsubsection{Evaluating the cost of using ${\mathcal H}_a$
               via fixed anchoring}

\begin{figure*}[htbp]
\begin{center}
\begin{tabular}{cc}
\hspace*{-.5cm}
\includegraphics[width=0.52\textwidth]{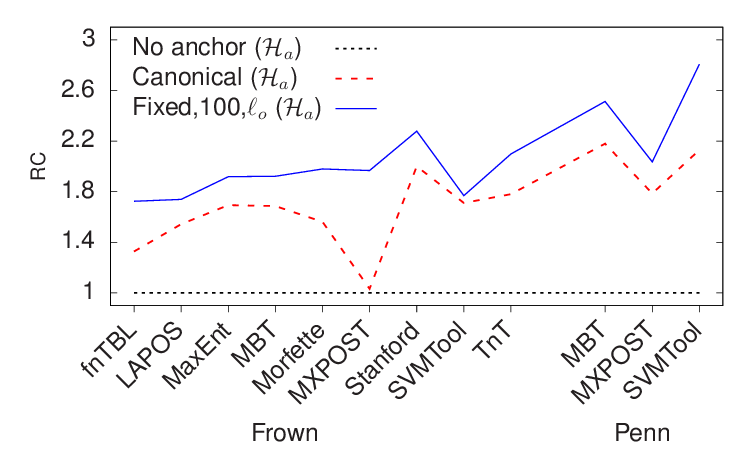} 
&
\hspace*{-.8cm}
\includegraphics[width=0.52\textwidth]{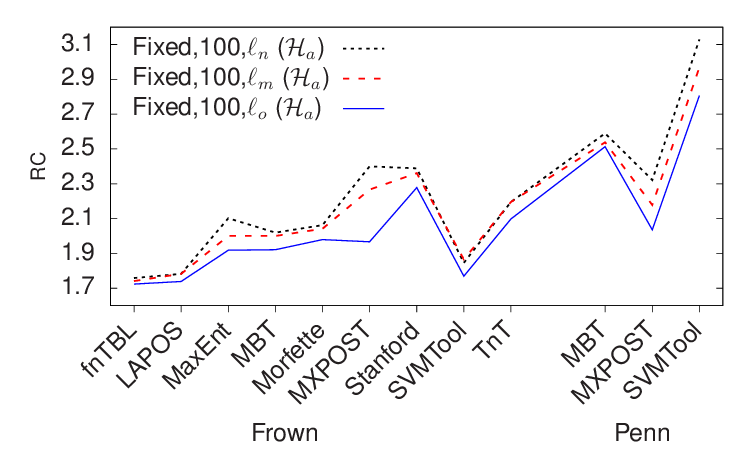}
\end{tabular}
\caption{{\sc rc} values for runs in ${\mathcal L}_{\mbox{\sc rc}}$ and 
  ${\mathcal L}_{\mbox{\sc rp}_\ell}$ (look-ahead $\imath$, value
  $\beta=100$).}
\label{fig-runs-RC}
\end{center}
\end{figure*}

The benchmark measure is now {\sc rc}, whose values are compiled in
Table~\ref{table-runs-RC} and illustrated in the left-most diagram of
Fig.~\ref{fig-runs-RC} for the set ${\mathcal L}_{\mbox{\sc rc}}$ of
significant runs for this issue. Note that the sole aim of the
estimates for the canonical technique is to illustrate the smaller
impact of using ${\mathcal H}_a$ with a less invasive anchoring where
possible, because in no way does the latter guarantee the completeness
of such a proximity criterion.

In greater detail, {\sc rc}{\footnotesize s} range from 1 for the
baselines -- anchor-free runs -- to 2.81 for {\sc svmt}ool on {\sc
  p}enn if fixed anchors are considered. In percentages, 77.78\% of
these values are less than 2, in an interval $[1, \infty)$ of possible
  costs. Analyzing each anchoring approach, this ratio grows to 100\%
  for the baselines, dropping to 75\% for those applying a canonical
  one and to 58.33\% for the fixed strategy with optimal
  look-ahead. The best score is for anchor-free runs in all local
  testing frames, while canonical anchors always provide
  the second best. Taking into account that convergence speed and
  relative cost are proportional, these results exemplify
  Theorem~\ref{th-anchoring-trace-comparison}. Specifically, they
  support both the greater computational complexity of applying fixed
  anchors and the superiority of anchor-free runs when the
  impact of irregularities on the learning process is limited.


\subsubsection{Evaluating the costs and benefits of using ${\mathcal H}_a$
  against ${\mathcal H}_r$}

\begin{figure*}[htbp]
\begin{center}
\begin{tabular}{cc}
\hspace*{-.5cm}
\includegraphics[width=0.52\textwidth]{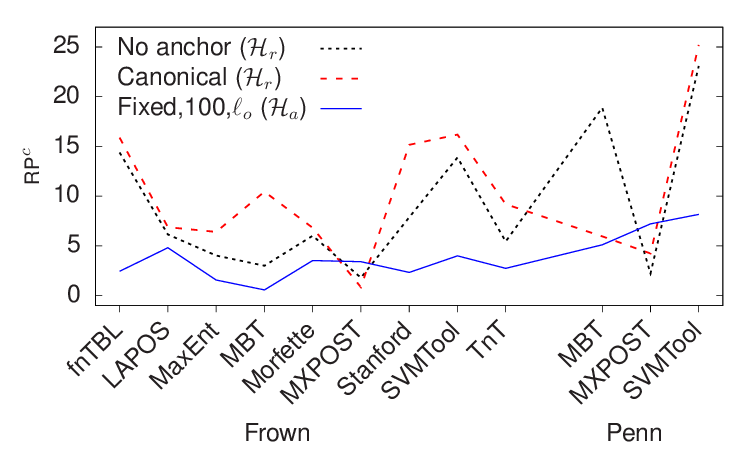} 
&
\hspace*{-.8cm}
\includegraphics[width=0.52\textwidth]{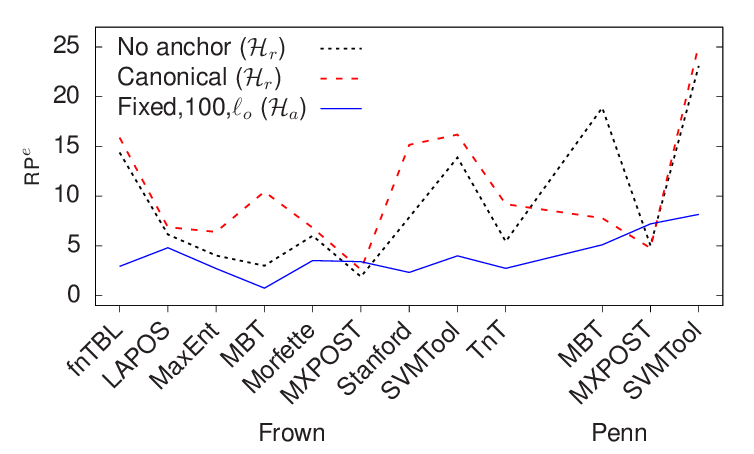}
\end{tabular}
\caption{$\mbox{\sc rp}^c$ and $\mbox{\sc rp}^e$ values for runs in
  ${\mathcal L}_{\mbox{\sc rp}}$.}
\label{fig-runs-RP}
\end{center}
\end{figure*}

Our metric is now $\mbox{\sc rp}^c$ (resp. $\mbox{\sc rp}^e$), whose
values are compiled in Table~\ref{table-runs-convergence-RP}
(resp. Table~\ref{table-runs-error-RP}) and illustrated in the
left-most (resp. right-most) diagram of Fig.~\ref{fig-runs-RP} for the
set ${\mathcal L}_{\mbox{\sc rp}}$ of significant runs for this issue,
as regards the treatment of convergence (resp. error) thresholds. Note
that most of these two tables are identical, just 22.2\% of the runs
have different values, which we have underlined so they can be easily
distinguished. The origin of this behaviour is the matching in most of
the runs of the values for $\mbox{\sc a}^c$ and $\mbox{\sc a}^e$.
This, in turn, is a consequence of the fact that when the impact of
irregularities in the learning process is limited, the maximum
divergence values considered for their calculation often occur at the
asymptotic level and, therefore, coincide.  

\begin{figure*}[htbp]
\begin{center}
\begin{tabular}{cc}
\hspace*{-.5cm}
\includegraphics[width=0.52\textwidth]{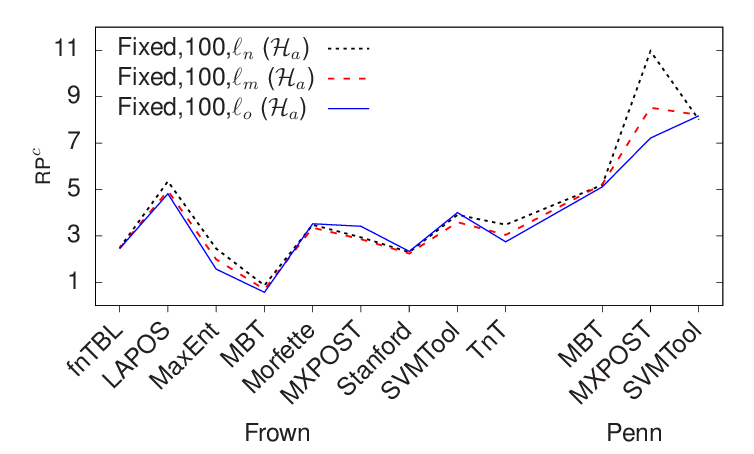} 
&
\hspace*{-.8cm}
\includegraphics[width=0.52\textwidth]{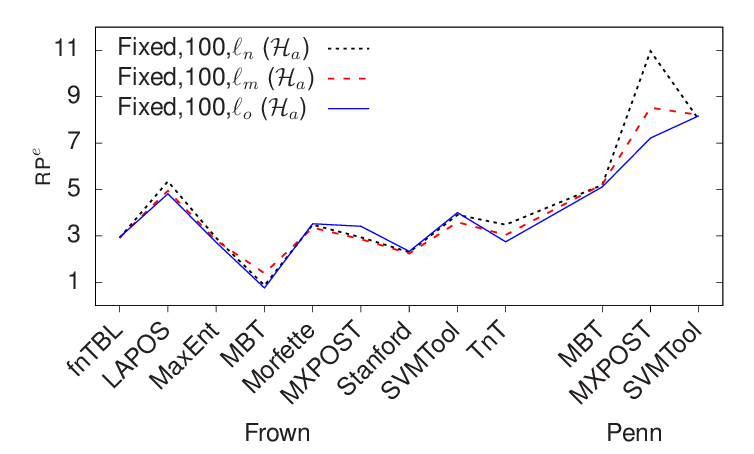}
\end{tabular}
\caption{$\mbox{\sc rp}^c$ and $\mbox{\sc rp}^e$ values for runs
  in ${\mathcal L}_{\mbox{\sc rp}_\ell}$ (look-ahead
  $\imath$, value $\beta=100$).}
\label{fig-runs-RP-fixed-anchors-absolute-thresholds}
\end{center}
\end{figure*}

In greater detail, anchor-free runs with ${\mathcal H}_r$ present the
(resp. second) best $\mbox{\sc rp}^c$ and $\mbox{\sc rp}^e$ values
in only one (resp. in ten) local testing frames, which corresponds to
{\sc mbt} on {\sc p}enn. Regarding the use of fixed anchors with
${\mathcal H}_a$, it turns out to be the (resp. second) best
choice in two (resp. in no) cases, {\sc mxpost} in both {\sc f}rown
and {\sc p}enn. In all other local testing frames, the canonical
anchoring with ${\mathcal H}_r$ achieves the (resp. second) best
results. Overall, as one would expect from its ability to adapt
dynamically to the evolution of the learning process, the best
performances come from the use of canonical anchors on ${\mathcal
  H}_r$. On the other hand, the strong static control imposed by the
fixed ones to ensure the completeness for ${\mathcal H}_a$
should relegate their use to situations where we have
no information about the magnitude of the irregularities and the
monotonicity -- increasing or decreasing -- of the learning trace
involved, or we simply need to ensure an absolute threshold.

\subsubsection{Stability of using ${\mathcal H}_a$ via fixed anchoring
               against look-ahead variations} Although the results
above illustrate the application of proximity conditions based on
absolute thresholds, they were obtained from runs with fixed anchors
and an optimal look-ahead $\ell_o$ resulting from a tuning
process. The aim therefore is now to determine the real impact of such
a process on run performance. In this context, we extend the set
$\Gamma$ of anchorings in each local testing frame
$\mathcal{L}^{\Lambda}_{\Gamma}[{\mathcal
    A}^\pi[\mathcal{D}^{\mathcal{K}}_{\sigma}],\tau]$ with the fixed
ones of value $\beta = 100$ and null look-ahead $\ell_n$, and also
$\ell_m := \ell_o/2$. As $\ell_o$ is assumed to provide the best
convergence speed for the proximity condition ${\mathcal H}_a$ in an
increasing setting sequence, $\ell_n$ should supply the lowest one and
$\ell_m$ intermediate ones. The monitoring of set ${\mathcal L}_\ell$,
which brings the significant runs for this issue, is compiled
(resp. illustrated) respectively for $\mbox{\sc rp}^c$ and $\mbox{\sc
  rp}^e$ in Tables~\ref{table-runs-augmented-convergence-RP} (resp. in
the left-most diagram of
Fig.~\ref{fig-runs-RP-fixed-anchors-absolute-thresholds})
and~\ref{table-runs-augmented-error-RP} (resp. in the right-most
diagram of Fig.~\ref{fig-runs-RP-fixed-anchors-absolute-thresholds}).
These tables also include the {\sc rc} scores, which are illustrated
separately in the right-most diagram of Fig.~\ref{fig-runs-RC}, and
most of their entries are identical as for runs in
${\mathcal L}_{\mbox{\sc rp}}$. Specifically, only 22.2\% of the
$\mbox{\sc rp}^c$ and $\mbox{\sc rp}^e$ values are different, and we
have again used underlined text to highlight them.

\begin{table*}[htbp]

\begin{center}
\caption{$\mbox{\sc rp}^c$ monitoring in ${\mathcal L}_{\mbox{\sc rp}_\ell}$}
\label{table-runs-augmented-convergence-RP}

\begin{scriptsize}
\begin{tabular}{@{\hspace{0pt}}l@{\hspace{3pt}}l@{\hspace{2pt}}l@{\hspace{2pt}}
                c@{\hspace{0pt}}c@{\hspace{3pt}}  
                c@{\hspace{0pt}}c@{\hspace{3pt}}  
                c@{\hspace{0pt}}c@{\hspace{3pt}}  
                c@{\hspace{0pt}}c@{\hspace{3pt}}  
                c@{\hspace{0pt}}c@{\hspace{3pt}}  
                c@{\hspace{0pt}}c@{\hspace{3pt}}  
                c@{\hspace{0pt}}c@{\hspace{3pt}}  
                c@{\hspace{0pt}}c@{\hspace{3pt}}  
                c@{\hspace{0pt}}c@{\hspace{3pt}}  
                c@{\hspace{0pt}}c@{\hspace{3pt}}  
                c@{\hspace{0pt}}c@{\hspace{3pt}}  
                c@{\hspace{0pt}}c@{\hspace{3pt}}  
                c@{\hspace{0pt}}c@{\hspace{3pt}}  
                c@{\hspace{0pt}}c@{\hspace{3pt}}  
                c@{\hspace{0pt}}c@{\hspace{3pt}}  
                c@{\hspace{0pt}}c@{\hspace{3pt}}  
                c@{\hspace{0pt}}c@{\hspace{3pt}}  
                c@{\hspace{0pt}}c@{\hspace{3pt}}  
                c@{\hspace{0pt}}c@{\hspace{3pt}}  
                c@{\hspace{0pt}}}                 

\hline
& & & \boldmath$\tau$ & & \multicolumn{13}{c}{{\bf Fixed} \boldmath$+\;{\mathcal H}_a$} & & \multicolumn{11}{c}{{\bf Fixed} \boldmath$+\;{\mathcal H}_a$} & & \multicolumn{11}{c}{{\bf Fixed} \boldmath$+\;{\mathcal H}_a$} \rule{0pt}{2.5ex} \\

\cline{1-2} \cline{4-4} \cline{6-18} \cline{20-30} \cline{32-42}

& & & & & \bf{\scshape put} & & \boldmath$\ell_o$ & & \bf{\scshape pl}evel & & \bf{\scshape cl}evel & & \boldmath$\mbox{\bf{\scshape a}}^c$ & & \bf{\scshape rc} & & \boldmath$\mbox{\bf{\scshape rp}}^c$ & & \boldmath$\ell_m$ & & \bf{\scshape pl}evel & & \bf{\scshape cl}evel & & \boldmath$\mbox{\bf{\scshape a}}^c$ & & \bf{\scshape rc} & & \boldmath$\mbox{\bf{\scshape rp}}^c$ & & \boldmath$\ell_n$ & & \bf{\scshape pl}evel & & \bf{\scshape cl}evel & & \boldmath$\mbox{\bf{\scshape a}}^c$ & & \bf{\scshape rc} & & \boldmath$\mbox{\bf{\scshape rp}}^c$ \rule{0pt}{2.5ex} \\

\cline{1-2} \cline{4-4} \cline{6-6} \cline{8-8} \cline{10-10} \cline{12-12} \cline{14-14} \cline{16-16} \cline{18-18} \cline{20-20} \cline{22-22} \cline{24-24} \cline{26-26} \cline{28-28} \cline{30-30} \cline{32-32} \cline{34-34} \cline{36-36} \cline{38-38} \cline{40-40} \cline{42-42}

\multirow{9}{*}{\begin{sideways}{\bf \textsc{f}rown}\end{sideways}} & fn\textsc{tbl} & & 4.26 & &  89.38 & &  6 & &  55 & &  100 & &  \underline{4.24} & &  1.72 & &  \underline{2.46} & &  3 & &  55 & &  101 & &  \underline{4.31} & &  1.74 & &  \underline{2.47} & & \bf 0 & & \bf 55 & & \bf 102 & & \bf \underline{4.37} & & \bf 1.76 & & \bf \underline{2.49}\rule{0pt}{2.75ex} \\
& \textsc{lapos} & & 3.47 & &  64.86 & &  18 & &  18 & &  80 & &  8.38 & &  1.74 & &  4.82 & &  9 & &  18 & &  82 & &  8.83 & &  1.78 & &  4.95 & & \bf 0 & & \bf 18 & & \bf 82 & & \bf 9.55 & & \bf 1.78 & & \bf 5.36\rule{0pt}{2.25ex} \\
& \textsc{m}ax\textsc{e}nt & & 4.42 & &  16.83 & &  52 & &  32 & &  94 & &  \underline{3.02} & &  1.92 & &  \underline{1.57} & &  26 & &  32 & &  98 & &  \underline{4.00} & &  2.00 & &  \underline{2.00} & & \bf 0 & & \bf 32 & & \bf 103 & & \bf \underline{5.17} & & \bf 2.10 & & \bf \underline{2.46}\rule{0pt}{2.25ex} \\
& \textsc{mbt} & & 4.78 & &  29.40 & &  43 & &  43 & &  98 & &  \underline{1.11} & &  1.92 & &  \underline{0.58} & &  22 & &  43 & &  102 & &  \underline{1.40} & &  2.00 & &  \underline{0.70} & & \bf 0 & & \bf 43 & & \bf 103 & & \bf 1.76 & & \bf 2.02 & & \bf 0.87\rule{0pt}{2.25ex} \\
& \textsc{m}orfette & & 3.67 & & \bf 69.71 & & \bf 19 & & \bf 20 & & \bf 95 & & \bf 6.98 & & \bf 1.98 & & \bf 3.53 & &  10 & &  20 & &  98 & &  6.85 & &  2.04 & &  3.35 & &  0 & &  20 & &  99 & &  7.18 & &  2.06 & &  3.48\rule{0pt}{2.25ex} \\
& \textsc{mxpost} & & 6.28 & & \bf 75.63 & & \bf 11 & & \bf 22 & & \bf 59 & & \bf 6.73 & & \bf 1.97 & & \bf 3.42 & &  6 & &  22 & &  68 & &  6.49 & &  2.27 & &  2.86 & &  0 & &  22 & &  72 & &  7.07 & &  2.40 & &  2.95\rule{0pt}{2.25ex} \\
& \textsc{s}tanford & & 4.31 & & \bf 78.29 & & \bf 12 & & \bf 24 & & \bf 82 & & \bf 5.33 & & \bf 2.28 & & \bf 2.34 & &  6 & &  24 & &  85 & &  5.30 & &  2.36 & &  2.24 & &  0 & &  24 & &  86 & &  5.50 & &  2.39 & &  2.30\rule{0pt}{2.25ex} \\
& \textsc{svmt}ool & & 3.82 & & \bf 76.02 & & \bf 10 & & \bf 41 & & \bf 92 & & \bf 7.09 & & \bf 1.77 & & \bf 4.01 & &  5 & &  41 & &  97 & &  6.71 & &  1.87 & &  3.60 & &  0 & &  41 & &  96 & &  7.22 & &  1.85 & &  3.91\rule{0pt}{2.25ex} \\
& \textsc{t}n\textsc{t} & & 3.69 & &  45.41 & &  32 & &  19 & &  86 & &  5.77 & &  2.10 & &  2.75 & &  16 & &  19 & &  90 & &  6.69 & &  2.20 & &  3.05 & & \bf 0 & & \bf 19 & & \bf 90 & & \bf 7.65 & & \bf 2.20 & & \bf 3.49\rule{0pt}{2.25ex} \\
\cline{1-2} \cline{4-4} \cline{6-6} \cline{8-8} \cline{10-10} \cline{12-12} \cline{14-14} \cline{16-16} \cline{18-18} \cline{20-20} \cline{22-22} \cline{24-24} \cline{26-26} \cline{28-28} \cline{30-30} \cline{32-32} \cline{34-34} \cline{36-36} \cline{38-38} \cline{40-40} \cline{42-42}
\multirow{3}{*}{\begin{sideways}{\bf \textsc{p}enn}\end{sideways}} & \textsc{mbt} & & 3.72 & &  39.50 & &  44 & &  15 & &  98 & &  12.85 & &  2.51 & &  5.11 & & \bf 22 & & \bf 15 & & \bf 99 & & \bf 13.26 & & \bf 2.54 & & \bf 5.22 & &  0 & &  15 & &  101 & &  13.47 & &  2.59 & &  5.20\rule{0pt}{2.75ex} \\
& \textsc{mxpost} & & 3.34 & &  18.73 & &  37 & &  17 & &  57 & &  14.70 & &  2.04 & &  7.22 & &  19 & &  17 & &  61 & &  18.58 & &  2.18 & &  8.53 & & \bf 0 & & \bf 17 & & \bf 65 & & \bf 25.46 & & \bf 2.32 & & \bf 10.97\rule{0pt}{2.25ex} \\
& \textsc{svmt}ool & & 2.64 & &  29.69 & &  47 & &  26 & &  87 & &  22.94 & &  2.81 & &  8.18 & & \bf 24 & & \bf 26 & & \bf 92 & & \bf 24.38 & & \bf 2.97 & & \bf 8.21 & &  0 & &  26 & &  97 & &  25.10 & &  3.13 & &  8.02\rule{0pt}{2.25ex} \\

\hline
\end{tabular}
\end{scriptsize}
\end{center}
\end{table*}

\begin{table*}[htbp]

\begin{center}
\caption{$\mbox{\sc rp}^e$ monitoring in ${\mathcal L}_{\mbox{\sc rp}_\ell}$}
\label{table-runs-augmented-error-RP}

\begin{scriptsize}
\begin{tabular}{@{\hspace{0pt}}l@{\hspace{3pt}}l@{\hspace{2pt}}l@{\hspace{2pt}}
                c@{\hspace{0pt}}c@{\hspace{3pt}}  
                c@{\hspace{0pt}}c@{\hspace{3pt}}  
                c@{\hspace{0pt}}c@{\hspace{3pt}}  
                c@{\hspace{0pt}}c@{\hspace{3pt}}  
                c@{\hspace{0pt}}c@{\hspace{3pt}}  
                c@{\hspace{0pt}}c@{\hspace{3pt}}  
                c@{\hspace{0pt}}c@{\hspace{3pt}}  
                c@{\hspace{0pt}}c@{\hspace{3pt}}  
                c@{\hspace{0pt}}c@{\hspace{3pt}}  
                c@{\hspace{0pt}}c@{\hspace{3pt}}  
                c@{\hspace{0pt}}c@{\hspace{3pt}}  
                c@{\hspace{0pt}}c@{\hspace{3pt}}  
                c@{\hspace{0pt}}c@{\hspace{3pt}}  
                c@{\hspace{0pt}}c@{\hspace{3pt}}  
                c@{\hspace{0pt}}c@{\hspace{3pt}}  
                c@{\hspace{0pt}}c@{\hspace{3pt}}  
                c@{\hspace{0pt}}c@{\hspace{3pt}}  
                c@{\hspace{0pt}}c@{\hspace{3pt}}  
                c@{\hspace{0pt}}c@{\hspace{3pt}}  
                c@{\hspace{0pt}}}                 

\hline
& & & \boldmath$\tau$ & & \multicolumn{13}{c}{{\bf Fixed} \boldmath$+\;{\mathcal H}_a$} & & \multicolumn{11}{c}{{\bf Fixed} \boldmath$+\;{\mathcal H}_a$} & & \multicolumn{11}{c}{{\bf Fixed} \boldmath$+\;{\mathcal H}_a$} \rule{0pt}{2.5ex} \\

\cline{1-2} \cline{4-4} \cline{6-18} \cline{20-30} \cline{32-42}

& & & & & \bf{\scshape put} & & \boldmath$\ell_o$ & & \bf{\scshape pl}evel & & \bf{\scshape cl}evel & & \boldmath$\mbox{\bf{\scshape a}}^c$ & & \bf{\scshape rc} & & \boldmath$\mbox{\bf{\scshape rp}}^c$ & & \boldmath$\ell_m$ & & \bf{\scshape pl}evel & & \bf{\scshape cl}evel & & \boldmath$\mbox{\bf{\scshape a}}^c$ & & \bf{\scshape rc} & & \boldmath$\mbox{\bf{\scshape rp}}^c$ & & \boldmath$\ell_n$ & & \bf{\scshape pl}evel & & \bf{\scshape cl}evel & & \boldmath$\mbox{\bf{\scshape a}}^c$ & & \bf{\scshape rc} & & \boldmath$\mbox{\bf{\scshape rp}}^c$ \rule{0pt}{2.5ex} \\

\cline{1-2} \cline{4-4} \cline{6-6} \cline{8-8} \cline{10-10} \cline{12-12} \cline{14-14} \cline{16-16} \cline{18-18} \cline{20-20} \cline{22-22} \cline{24-24} \cline{26-26} \cline{28-28} \cline{30-30} \cline{32-32} \cline{34-34} \cline{36-36} \cline{38-38} \cline{40-40} \cline{42-42}

\multirow{9}{*}{\begin{sideways}{\bf \textsc{f}rown}\end{sideways}} & fn\textsc{tbl} & & 4.26 & & \bf 89.38 & & \bf 6 & & \bf 55 & & \bf 100 & & \bf \underline{5.09} & & \bf 1.72 & & \bf \underline{2.95} & &  3 & &  55 & &  101 & &  \underline{5.10} & &  1.74 & &  \underline{2.93} & &  0 & &  55 & &  102 & &  \underline{5.13} & &  1.76 & &  \underline{2.92}\rule{0pt}{2.75ex} \\
& \textsc{lapos} & & 3.47 & &  64.86 & &  18 & &  18 & &  80 & &  8.38 & &  1.74 & &  4.82 & &  9 & &  18 & &  82 & &  8.83 & &  1.78 & &  4.95 & & \bf 0 & & \bf 18 & & \bf 82 & & \bf 9.55 & & \bf 1.78 & & \bf 5.36\rule{0pt}{2.25ex} \\
& \textsc{m}ax\textsc{e}nt & & 4.42 & &  16.83 & &  52 & &  32 & &  94 & &  \underline{5.23} & &  1.92 & &  \underline{2.73} & &  26 & &  32 & &  98 & &  \underline{5.62} & &  2.00 & &  \underline{2.81} & & \bf 0 & & \bf 32 & & \bf 103 & & \bf \underline{6.13} & & \bf 2.10 & & \bf \underline{2.92}\rule{0pt}{2.25ex} \\
& \textsc{mbt} & & 4.78 & &  29.40 & &  43 & &  43 & &  98 & &  \underline{1.46} & &  1.92 & &  \underline{0.76} & & \bf 22 & & \bf 43 & & \bf 102 & & \bf \underline{2.79} & & \bf 2.00 & & \bf \underline{1.40} & &  0 & &  43 & &  103 & &  1.76 & &  2.02 & &  0.87\rule{0pt}{2.25ex} \\
& \textsc{m}orfette & & 3.67 & & \bf 69.71 & & \bf 19 & & \bf 20 & & \bf 95 & & \bf 6.98 & & \bf 1.98 & & \bf 3.53 & &  10 & &  20 & &  98 & &  6.85 & &  2.04 & &  3.35 & &  0 & &  20 & &  99 & &  7.18 & &  2.06 & &  3.48\rule{0pt}{2.25ex} \\
& \textsc{mxpost} & & 6.28 & & \bf 75.63 & & \bf 11 & & \bf 22 & & \bf 59 & & \bf 6.73 & & \bf 1.97 & & \bf 3.42 & &  6 & &  22 & &  68 & &  6.49 & &  2.27 & &  2.86 & &  0 & &  22 & &  72 & &  7.07 & &  2.40 & &  2.95\rule{0pt}{2.25ex} \\
& \textsc{s}tanford & & 4.31 & & \bf 78.29 & & \bf 12 & & \bf 24 & & \bf 82 & & \bf 5.33 & & \bf 2.28 & & \bf 2.34 & &  6 & &  24 & &  85 & &  5.30 & &  2.36 & &  2.24 & &  0 & &  24 & &  86 & &  5.50 & &  2.39 & &  2.30\rule{0pt}{2.25ex} \\
& \textsc{svmt}ool & & 3.82 & & \bf 76.02 & & \bf 10 & & \bf 41 & & \bf 92 & & \bf 7.09 & & \bf 1.77 & & \bf 4.01 & &  5 & &  41 & &  97 & &  6.71 & &  1.87 & &  3.60 & &  0 & &  41 & &  96 & &  7.22 & &  1.85 & &  3.91\rule{0pt}{2.25ex} \\
& \textsc{t}n\textsc{t} & & 3.69 & &  45.41 & &  32 & &  19 & &  86 & &  5.77 & &  2.10 & &  2.75 & &  16 & &  19 & &  90 & &  6.69 & &  2.20 & &  3.05 & & \bf 0 & & \bf 19 & & \bf 90 & & \bf 7.65 & & \bf 2.20 & & \bf 3.49\rule{0pt}{2.25ex} \\
\cline{1-2} \cline{4-4} \cline{6-6} \cline{8-8} \cline{10-10} \cline{12-12} \cline{14-14} \cline{16-16} \cline{18-18} \cline{20-20} \cline{22-22} \cline{24-24} \cline{26-26} \cline{28-28} \cline{30-30} \cline{32-32} \cline{34-34} \cline{36-36} \cline{38-38} \cline{40-40} \cline{42-42}
\multirow{3}{*}{\begin{sideways}{\bf \textsc{p}enn}\end{sideways}} & \textsc{mbt} & & 3.72 & &  39.50 & &  44 & &  15 & &  98 & &  12.85 & &  2.51 & &  5.11 & & \bf 22 & & \bf 15 & & \bf 99 & & \bf 13.26 & & \bf 2.54 & & \bf 5.22 & &  0 & &  15 & &  101 & &  13.47 & &  2.59 & &  5.20\rule{0pt}{2.75ex} \\
& \textsc{mxpost} & & 3.34 & &  18.73 & &  37 & &  17 & &  57 & &  14.70 & &  2.04 & &  7.22 & &  19 & &  17 & &  61 & &  18.58 & &  2.18 & &  8.53 & & \bf 0 & & \bf 17 & & \bf 65 & & \bf 25.46 & & \bf 2.32 & & \bf 10.97\rule{0pt}{2.25ex} \\
& \textsc{svmt}ool & & 2.64 & &  29.69 & &  47 & &  26 & &  87 & &  22.94 & &  2.81 & &  8.18 & & \bf 24 & & \bf 26 & & \bf 92 & & \bf 24.38 & & \bf 2.97 & & \bf 8.21 & &  0 & &  26 & &  97 & &  25.10 & &  3.13 & &  8.02\rule{0pt}{2.25ex} \\

\hline
\end{tabular}
\end{scriptsize}
\end{center}
\end{table*} 

In short, we find that {\sc rc}{\footnotesize s} range from 1.72 for
fn{\sc tbl} on {\sc f}rown with a look-ahead $\ell_o$, to 3.13 for
{\sc svmt}ool on {\sc p}enn with $\ell_n$. In percentages, 36.11\% of
these values are less than 2, in an interval $[1, \infty)$ of possible
  costs. Analyzing each anchoring approach, this ratio grows to
  58.33\% when using $\ell_o$ look-aheads and drops to 25\%
  otherwise. The best score is for $\ell_o$ in all local testing
  frames, while $\ell_m$ provides the second best result in all cases
  but one, and $\ell_n$ in three. Note that $\ell_m$ and $\ell_n$
  obtain the same {\sc rc}{\footnotesize s} for {\sc lapos} and
  \textsc{t}n\textsc{t} in {\sc f}rown. This again exemplifies the
  conclusions of Theorem~\ref{th-anchoring-trace-comparison} about
  anchor categorization, this time regarding the use of look-aheads,
  but also illustrates the usefulness and validity of the concept of
  minimal look-ahead for a tentative {\sc put} as a mechanism for
  optimizing fixed anchorings.

  As for $\mbox{\sc rp}^c$ (resp. $\mbox{\sc rp}^e$), the optimal
  look-ahead $\ell_o$ gives the best values in four (resp. five)
  cases, while $\ell_m$ and $\ell_n$ do so in two (resp. three) and
  six (resp. four) ones, respectively.  Regarding the second best
  choice, it corresponds to $\ell_o$ in only one (resp. one) case and
  in six (resp. five) ones to $\ell_m$, while anchor-free runs reach
  that position five (resp. six) times. Overall, the best performances
  seem to correspond to the absence of look-ahead, followed by those
  associated with the use of the mean value $\ell_m$ and the optimum
  one $\ell_o$, although the differences are very small as can be seen
  in Fig.~\ref{fig-runs-RP-fixed-anchors-absolute-thresholds}. This
  suggests that the choice of look-ahead has a minor impact on
  the performance, thus allowing its use to be simplified, as it would
  no longer require a prior tuning phase.


\section{Conclusions}
\label{section-conclusions}

We have responded to the challenge of estimating absolute convergence
thresholds associated with the prediction of learning
processes as a means of reducing training effort and the need for
resources in the generation of {\sc ml}-based systems. The goal is to
get the most from a non-active adaptive sampling scheme used for that
purpose, by limiting its application in time to what is strictly
necessary, while avoiding the limitations of relative measures.

Our proposal proves its correctness with respect to its working
hypotheses. Namely, it determines the cycle from which we can ensure
that the threshold fixed by the user has been reached. Since this can
only be established in practice when the successive estimates for
accuracy are decreasing, the completeness of the technique is also
stated. To do so, we demonstrate that it is possible to redirect the
training dynamics in such a way that such a property verifies. This is
achieved by properly using the concept of anchor, a conservative
assessment of the final accuracy achievable by the learner, which is
calculated from a sufficiently representative sample interpreted as an
observation at the point of infinity for the calculation of the next
estimate. Furthermore, since the primary function of anchoring is to
compensate the irregularities in the learning process due to
deviations in the working hypotheses, the proposal shows a good degree
of robustness.

To reduce the slowdown caused on the pace of convergence by the use of
anchors, we introduce a parameterizable family of these structures,
categorizing them with respect to both the costs and the balance
between these and their benefits. The tests, taking the generation of
{\sc pos} taggers in {\sc nlp} as case study, corroborate our
expectations. In particular, although the consideration of absolute
thresholds applying our proposal entails greater computational cost,
it has demonstrated its reliability and practical aplicability, 
providing a stable and robust way to proceed when relative estimations
of the learning curve are not sufficient for the development of {\sc
ml} applications.


\section*{Acknowledgments}

\begin{small}
  Research partially funded by the Spanish Ministry of Economy and
  Competitiveness through projects TIN2017-85160-C2-1-R,
  TIN2017-85160-C2-2-R, PID2020-113230RB-C21 and PID2020-113230RB-C22,
  and by the Galician Regional Government under project ED431C
  2018/50.
\end{small}

\appendix

\section{Proofs for Subsection~\ref{subsection-completeness}}
\label{appendix-proofs}

\subsection{Proof of Theorem~\ref{th-sufficient-condition-for-decreasing-backbone}}

Since $\hat\rho_{i}(\infty) := \hat{\mathcal A}_i(\infty) -
\hat\alpha_{i}, \forall i > \omega$, the asymptotic
backbone $\{\hat\alpha_i\}_{i > \omega}$ is decreasing iff it verifies
that
\begin{equation}
\forall i > \omega, \; \hat\alpha_{i} \geq \hat\alpha_{i+1}
\Leftrightarrow \hat{\mathcal A}_i(\infty) - \hat{\mathcal A}_{i+1}(\infty) \geq \hat\rho_{i}(\infty) -
\hat\rho_{i+1}(\infty), \;
\end{equation}
\noindent which derives immediately from
Equation~\ref{equation-condition-anchors-evolution}. To now complete
the proof, we only need to demonstrate that $\{\hat\alpha_i\}_{i
> \omega}$ converges to $\alpha_{\dinfty{}}$. As $\hat{\mathcal
A}_i^\pi[{\mathcal D}^{\mathcal {K}}_{\sigma}]$ is a fitting curve for the values
\begin{equation}
\{[x_j, {\mathcal A}_{\dinfty{}}[{\mathcal D}](x_j)], \; x_j := \absd{\mathcal D_j} \}_{j=1}^{i}
\; \cup \; \{[\infty, \hat{\mathcal A}_{i}(\infty)]\}, \; \forall i > \omega
\end{equation}
\noindent we then have that
\begin{equation}
\hat{\mathcal A}_i(\infty)
\geq \hat\alpha_i := \lim \limits_{x \rightarrow \infty}\hat{\mathcal
  A}_i^\pi[{\mathcal D}^{\mathcal {K}}_{\sigma}](x), \; \forall i >
\omega
\end{equation}
\noindent If we also take into account that ${\mathcal A}_i^\pi[{\mathcal
    D}^{\mathcal {K}}_{\sigma}]$ is a curve fitting the set 
\begin{equation}
\{[x_j, {\mathcal A}_{\dinfty{}}[{\mathcal D}](x_j)], \; x_j := \absd{\mathcal D_j} \}_{j=1}^{i}
\; \cup \; \{[\infty, \alpha_i]\}, \; \forall i >
\omega
\end{equation}
\noindent with anchors verifying the
Equation~\ref{equation-condition-anchors-lower-bounded}, we then have
that
\begin{equation}
\hat{\mathcal A}_i(\infty)
\geq \hat\alpha_i := \lim \limits_{x \rightarrow \infty}\hat{\mathcal
  A}_i^\pi[{\mathcal D}^{\mathcal {K}}_{\sigma}](x) \geq \alpha_i, \;
\forall i > \wp
\end{equation}
\noindent from which
\begin{equation}
\lim \limits_{i \rightarrow \infty} \hat{\mathcal A}_i(\infty) \geq \lim \limits_{i
  \rightarrow \infty} \hat\alpha_i \geq \lim \limits_{i \rightarrow
  \infty} \alpha_i = \alpha_{\dinfty{}}
\end{equation}
\noindent Moreover, the impact of the singularity $\hat{\mathcal
  A}_\infty^\pi[{\mathcal D}^{\mathcal {K}}_{\sigma}](\infty)$ in the
generation of the learning trends $\{\hat{\mathcal A}_{i}[{\mathcal
    D}^{\mathcal {K}}_{\sigma}]\}_{i>\omega}$ decreases as the level
ascends. Namely, $\alpha_{\dinfty{}}$ is a supremum for
$\{\hat\alpha_i\}_{i > \omega}$ and $\lim \limits_{i \rightarrow
  \infty} \hat\alpha_i = \lim \limits_{i \rightarrow \infty} \alpha_i
= \alpha_{\dinfty{}}$, which proves the thesis.

\subsection{Proof of Theorem~\ref{th-canonical-anchoring-trace-decreasing}}

Let $\{\alpha_i\}_{i \in \mathbb{N}}$ be the backbone for the
reference $[{\mathcal A}^\pi[{\mathcal D}^{\mathcal
      {K}}_{\sigma}],\omega]$ of $\hat{\mathcal A}^\pi[{\mathcal
    D}^{\mathcal {K}}_{\sigma}]$ and $\wp$ the {\sc pl}evel of the
latter. By
Theorem~\ref{th-sufficient-condition-for-decreasing-backbone}, it is
enough to prove that its hypotheses verify. Focusing on 
Equation~\ref{equation-condition-anchors-lower-bounded}, let us first
assume $i=\omega+1$. Since by hypothesis $\{\alpha_i\}_{i \in
  \mathbb{N}}$ is decreasing, we conclude that
\begin{equation}
\hat{\mathcal A}_{\omega+1}(\infty) := \alpha_{\omega} \geq
\alpha_{\omega+1}
\end{equation}

\noindent Let us now assume that $i>\omega+1$, as $\{\alpha_i\}_{i \in
  \mathbb{N}}$ is decreasing,
Theorem~\ref{th-canonical-anchoring-trace} proves that
$\hat\alpha_{\omega + i} \geq \alpha_{\omega + i}, \forall i >
\omega$, from which we derive that
\begin{equation}
\hat{\mathcal A}_{i+1}(\infty) := \hat\alpha_{i} \geq \alpha_{i}
\geq \alpha_{i+1}
\end{equation}
\noindent and we can then affirm that $\hat{\mathcal
  A}_{i}(\infty) \geq \alpha_{i}, \; \forall i > \wp \geq \omega$,
completing the proof in this case. \\

\noindent Regarding the compliance of the condition referred in
Equation~\ref{equation-condition-anchors-evolution}, as
$\hat\rho_i(\infty) := \hat{\mathcal A}_{i}(\infty)
- \hat\alpha_i$, demonstrating the inequality
\begin{equation}
\hat{\mathcal A}_i(\infty) -
\hat{\mathcal A}_{i+1}(\infty) \geq \hat\rho_{i}(\infty) -
\hat\rho_{i+1}(\infty), \; \forall i > 0 + \omega =
\omega
\end{equation}
\noindent which is to prove $\hat\alpha_{i} \geq
\hat\alpha_{i+1}, \forall i > \omega$. As this was
stated in Theorem~\ref{th-canonical-anchoring-trace}, we conclude
the thesis.

\subsection{Proof of Theorem~\ref{th-fixed-anchoring-trace}}

By Theorem~\ref{th-sufficient-condition-for-decreasing-backbone}, as
$\{\hat{\mathcal A}_i^\beta(\infty)\}_{i > \omega}$ is convergent to
$\beta$, it is enough to prove that its hypotheses verify. To this
end, the condition in
Equation~\ref{equation-condition-anchors-lower-bounded} becomes
trivial because the {\sc pl}evel $\wp$ is the first level after the
working one $\omega$ from which the asymptotic backbone is below 100,
and therefore
\begin{equation}
\hat{\mathcal A}_i^\beta(\infty)
:= \beta \geq 100 \geq \alpha_{i}, \; \forall i > \wp
\end{equation}

\noindent With respect to the condition in
Equation~\ref{equation-condition-anchors-evolution}, given that
$\hat{\mathcal A}_i^{\stackrel{\beta}{\pi}}[{\mathcal D}^{\mathcal
{K}}_{\sigma}]$ is a curve fitting 
\begin{equation}
\{[x_j, {\mathcal A}_{\dinfty{}}[{\mathcal D}](x_j)], \; x_j := \absd{\mathcal D_j} \}_{j=1}^{i}
\; \cup \; \{[\infty, \beta]\}
\end{equation}
\noindent and the observations of the first set are lower or
equal than $100 \leq \beta$, we have that the asymptotic backbone
$\{\hat\alpha_i^\beta\}_{i>\omega}$ of $\hat{\mathcal
A}^{\stackrel{\beta}{\pi}}[{\mathcal D}^{\mathcal {K}}_{\sigma}]$ is
never greater than $\beta$ and therefore the residuals
$\{\hat\rho_{i}^\beta(\infty)\}_{i>\omega}$ are invariably positive or
null, because
\begin{equation}
\hat\rho_i^\beta(\infty) := \hat{\mathcal A}_{i}^\beta(\infty) - \hat\alpha_i^\beta := \beta - \hat\alpha_i^\beta
\end{equation}
\noindent As $\{[x_i, {\mathcal A}_{\dinfty{}}[{\mathcal
      D}](x_i)], \; x_i := \absd{\mathcal D_i}
\}_{i \in \mathbb{N}}$ is increasing, its impact on the generation of
the learning trends $\{\hat{\mathcal
  A}_{i}^{\stackrel{\beta}{\pi}}[{\mathcal D}^{\mathcal
  {K}}_{\sigma}]\}_{i>\omega}$ grows with each sample and
  therefore the absolute value of the residuals
  $\{\hat\rho_{i}^\beta(\infty)\}_{i>\omega}$ also. We then derive
  that:
\begin{equation}
\hat{\mathcal A}_i^\beta(\infty) - \hat{\mathcal
  A}_{i+1}^\beta(\infty) = 0 \geq \hat\rho_{i}^\beta(\infty)
- \hat\rho_{i+1}^\beta(\infty),
\; \forall i > \omega
\end{equation}
and Equation~\ref{equation-condition-anchors-evolution}
verifies. 

\subsection{Proof of Theorem~\ref{th-completeness-traces}}

By Theorem~\ref{th-fixed-anchoring-trace}, as ${\mathcal
  A}^\pi[{\mathcal D}^{\mathcal {K}}_{\sigma}]$ is a learning trace
with fixed anchor, its asymptotic backbone $\{\alpha_i\}_{i > \omega}$
is decreasing and the thesis is proved applying 
the same reasoning used in Theorem~\ref{th-correctness-trace}.

\subsection{Proof of Theorem~\ref{th-fixed-anchoring-trace-with-look-ahead}}

Following
Theorem~\ref{th-sufficient-condition-for-decreasing-backbone}, as
$\{\hat{\mathcal A}_i^{\beta,\ell}(\infty)\}_{i > \omega}$ converges
to $\hat{\mathcal A}_{\omega + \ell}^{\beta,\ell}(\infty)$, it is
enough to prove that its hypotheses verify. With regard to the
condition expressed in
Equation~\ref{equation-condition-anchors-lower-bounded}, let us assume
$\{\hat{\mathcal A}_i^{\stackrel{\beta}{\pi}}[{\mathcal
D}^{\mathcal {K}}_{\sigma}]\}_{i > \omega}$ the fixed anchoring
learning trace of value $\beta$ for ${\mathcal A}^\pi[{\mathcal
D}^{\mathcal {K}}_{\sigma}]$. As its asymptotic backbone
$\{\hat\alpha_i^\beta\}_{i > \omega}$ is decreasing and $\wp \geq
\omega$, we have that, $\forall i \geq
\wp + \ell + 1$
\begin{equation} 
\hat\alpha_{\wp + \ell}^\beta \geq
\hat\alpha_{i}^\beta := \lim \limits_{x \rightarrow \infty}
\hat{\mathcal A}_{i}^{\stackrel{\beta}{\pi}}[{\mathcal D}^{\mathcal
    {K}}_{\sigma}](x)
\end{equation} 
where $\hat{\mathcal A}_{i}^{\stackrel{\beta}{\pi}}[{\mathcal
D}^{\mathcal {K}}_{\sigma}]$ is a curve fitting the set of values
\begin{equation}
\{[x_j, {\mathcal A}_{\dinfty{}}[{\mathcal D}](x_j)]\}_{j=1}^{i} \; \cup \;
\{[\infty, \hat{\mathcal A}_{i}^{\beta}(\infty)]\} = 
\{[x_j, {\mathcal A}_{\dinfty{}}[{\mathcal D}](x_j)]\}_{j=1}^{i} \; \cup \;
\{[\infty, \beta]\}, \; x_j := \absd{\mathcal
  D_j}
\end{equation}
\noindent Since ${\mathcal A}_{i}^\pi[{\mathcal D}^{\mathcal
    {K}}_{\sigma}]$ is a curve fitting the first set $\{[x_j,
  {\mathcal A}_{\dinfty{}}[{\mathcal D}](x_j)], \; x_j :=
\absd{\mathcal D_j}\}_{j=1}^{i}$ and by hypothesis $\beta \geq 100
\geq {\mathcal A}_{\dinfty{}}[{\mathcal D}](x_j), \forall j \in
\mathbb{N}$, we deduce that $\hat\alpha_{i}^\beta \geq \alpha_{i}$,
with $\{\alpha_i\}_{i \in \mathbb{N}}$ the asymptotic backbone for the
reference $[{\mathcal A}^\pi[{\mathcal D}^{\mathcal
      {K}}_{\sigma}],\omega]$. Taking also into account that $\lim
\limits_{x \rightarrow \infty} \hat{\mathcal
  A}_{i}^{\stackrel{\beta}{\pi}}[{\mathcal D}^{\mathcal
    {K}}_{\sigma}](x) := \hat\alpha_{i}^\beta$ and $\lim \limits_{x
  \rightarrow \infty} {\mathcal A}_{i}^\pi[{\mathcal D}^{\mathcal
    {K}}_{\sigma}](x) := \alpha_{i}$, it verifies that
\begin{equation}
\hat{\mathcal A}_{i}^{\beta,\ell}(\infty) :=
\hat\alpha_{\wp + \ell}^{\beta,\ell} :=
\hat\alpha_{\wp + \ell}^\beta \geq
\hat\alpha_{i}^\beta \geq \alpha_{i}, \; \forall i \geq
\wp + \ell + 1
\end{equation}
\noindent Furthermore, since $\wp$ is
precisely the first level after the working one
$\omega$ from which the asymptotic backbone is
below 100, we deduce that 
\begin{equation}
\hat{\mathcal A}_i^{\beta,\ell}(\infty) := \beta \geq 100 \geq
\alpha_{i}, \; \forall \; \wp + \ell + 1 > i >
\omega
\end{equation}
\noindent and therefore $\hat{\mathcal
  A}_i^{\beta,\ell}(\infty) \geq \alpha_{i}, \forall i >
\omega$, thus matching the relationship
described in
Equation~\ref{equation-condition-anchors-lower-bounded}. \\

\noindent With respect to the condition in
Equation~\ref{equation-condition-anchors-evolution}, we study it
separately in each interval of definition for the
asymptotic backbone $\{\hat\alpha_i^{\beta,\ell}\}_{i > \omega}$. Let
us first assume $\wp + \ell + 1 > i > \omega$. As in this case
\begin{equation}
\hat{\mathcal A}_i^{\beta,\ell}(\infty) := \hat{\mathcal
  A}_i^{\beta}(\infty) := \beta \geq 100
\end{equation}
\noindent we can apply the same reasoning used in
Theorem~\ref{th-fixed-anchoring-trace} to prove that
$\{\hat\alpha_i^{\beta,\ell}\}_{i > \omega}$ is decreasing in that
interval. \\

\noindent Let us now consider $i > \wp + \ell + 1$. Then, $\hat{\mathcal
  A}_i^{\stackrel{\beta,\ell}{\pi}}[{\mathcal D}^{\mathcal
    {K}}_{\sigma}]$ is here a curve fitting the collection of values
\begin{equation}
\{[x_j, {\mathcal A}_{\dinfty{}}[{\mathcal D}](x_j)]\}_{j=1}^{i}
\; \cup \; \{[\infty, \hat{\mathcal
    A}_i^{\beta,\ell}(\infty)]\} = \{[x_j, {\mathcal A}_{\dinfty{}}[{\mathcal D}](x_j)]\}_{j=1}^{i}
\; \cup \; \{[\infty, \hat\alpha_{\wp + \ell}^{\beta,\ell}]\}
\end{equation}
\noindent with $x_j := \absd{\mathcal D_j}$. Given that the
observations of the first set are always lower or equal than
\begin{equation}
\hat{\mathcal
  A}_i^{\beta,\ell}(\infty) := \hat\alpha_{\wp +
    \ell}^{\beta,\ell} := \hat\alpha_{\wp +
    \ell}^\beta
\end{equation}
\noindent because Theorem~\ref{th-fixed-anchoring-trace} states that
$\{\hat\alpha_i^\beta\}_{i > \omega}$ is decreasing, the subsequence
$\{\hat\alpha_i^{\beta,\ell}\}_{i>\wp+\ell+1}$ of the asymptotic
backbone of $\hat{\mathcal A}^{\stackrel{\beta,\ell}{\pi}}[{\mathcal
    D}^{\mathcal {K}}_{\sigma}]$ is never greater than
$\hat\alpha_{\wp + \ell}^{\beta,\ell}$ and therefore the sequence
of its associated residuals
$\{\hat\rho_{i}^{\beta,\ell}(\infty)\}_{i>\wp+\ell+1}$ is invariably
positive or null, because
\begin{equation}
\hat\rho_i^{\beta,\ell}(\infty) := \hat{\mathcal
  A}_{i}^{\beta,\ell}(\infty) - \hat\alpha_i^{\beta,\ell} :=
\hat{\mathcal A}_{i}^{\beta,\ell}(\infty) - \lim \limits_{x \rightarrow
  \infty} \hat{\mathcal A}_i^{\stackrel{\beta,\ell}{\pi}}[{\mathcal
    D}^{\mathcal {K}}_{\sigma}](x)
\end{equation}
\noindent As
$\{[x_i, {\mathcal A}_{\dinfty{}}[{\mathcal D}](x_i)], \; x_i := \absd{\mathcal
    D_i} \}_{i \in \mathbb{N}}$ is increasing, its impact on the
    generation of the learning trends $\{\hat{\mathcal
    A}_{i}^{\stackrel{\beta,\ell}{\pi}}[{\mathcal D}^{\mathcal
    {K}}_{\sigma}]\}_{i>\wp+\ell+1}$ grows with each sample and
    therefore the absolute value of the residuals
    $\{\hat\rho_{i}^{\beta,\ell}(\infty)\}_{i > \wp+\ell+1}$ also. We
    then derive that:
\begin{equation}
\hat{\mathcal A}_i^{\beta,\ell}(\infty) - \hat{\mathcal
  A}_{i+1}^{\beta,\ell}(\infty) =
0 \geq \hat\rho_{i}^{\beta,\ell}(\infty)
- \hat\rho_{i+1}^{\beta,\ell}(\infty), \; \forall i >
\wp + \ell + 1
\end{equation}
and Equation~\ref{equation-condition-anchors-evolution} also 
verifies in the latter interval. \\

\noindent That way, what is lacking to match the relationship
in Equation~\ref{equation-condition-anchors-evolution} on
its full application domain is to demonstrate that
\begin{equation}
\hat{\mathcal A}_{\wp + \ell}^{\beta,\ell}(\infty) - \hat{\mathcal
  A}_{\wp + \ell+1}^{\beta,\ell}(\infty) \geq
\hat\rho_{\wp + \ell}^{\beta,\ell}(\infty)
- \hat\rho_{\wp + \ell +1}^{\beta,\ell}(\infty)
\end{equation}
\noindent which is equivalent to prove that $\hat\alpha_{\wp + \ell}^{\beta,\ell} \geq \hat\alpha_{\wp + \ell+1}^{\beta,\ell}$, because
$\hat\rho_i^{\beta,\ell}(\infty) := \hat{\mathcal
A}_{i}^{\beta,\ell}(\infty) -
\hat\alpha_i^{\beta,\ell}$. \\

\noindent From Theorem~\ref{th-fixed-anchoring-trace},
$\{\hat\alpha_{i}^\beta\}_{i > \omega}$ converges decreasingly to the
final accuracy $\alpha_{\dinfty{}} :=
\lim \limits_{x \rightarrow \infty} {\mathcal A}_{\dinfty}[{\mathcal
    D}](x)$ attained from the training process. As ${\mathcal
A}_{\dinfty{}}[{\mathcal D}]$ is increasing, we infer that
\begin{equation}
{\mathcal A}_{\dinfty{}}[{\mathcal D}](x_i) \leq \alpha_{\dinfty{}} \leq
\hat\alpha_{\wp + \ell}^\beta :=
\hat\alpha_{\wp + \ell}^{\beta,\ell}, \; x_i := \absd{\mathcal D_i}
\end{equation}
\noindent Furthermore, $\hat\alpha_{\wp +
  \ell+1}^{\beta,\ell} := \lim \limits_{x \rightarrow \infty}
\hat{\mathcal A}_{\wp +
  \ell+1}^{\stackrel{\beta,\ell}{\pi}}[{\mathcal D}^{\mathcal
    {K}}_{\sigma}](x)$, where $\hat{\mathcal
  A}_{\wp + \ell+1}^{\stackrel{\beta,\ell}{\pi}}[{\mathcal
    D}^{\mathcal {K}}_{\sigma}]$ is a fitting curve for the set
\begin{equation}
\{[x_j, {\mathcal A}_{\dinfty{}}[{\mathcal D}](x_j)]\}_{j=1}^{\wp + \ell} \; \cup \;
\{[\infty, \hat{\mathcal A}_{\wp +
    \ell}^{\beta,\ell}(\infty)]\} = \{[x_j, {\mathcal
      A}_{\dinfty{}}[{\mathcal D}](x_j)]\}_{j=1}^{\wp + \ell} \; \cup \;
\{[\infty, \hat\alpha_{\wp + \ell -1}^{\beta,\ell}]\}
\end{equation}
\noindent with $x_j := \absd{\mathcal D_j}$. As we have established
that the ordinates of these values are below $\hat\alpha_{\wp +
  \ell+1}^{\beta,\ell}$, we derive what we were looking for
\begin{equation}
\hat\alpha_{\wp + \ell}^{\beta,\ell} \geq 
\lim \limits_{x \rightarrow \infty} \hat{\mathcal
  A}_{\wp + \ell+1}^{\stackrel{\beta,\ell}{\pi}}[{\mathcal D}^{\mathcal
    {K}}_{\sigma}](x) := \hat\alpha_{\wp + \ell+1}^{\beta,\ell}
\end{equation}
\noindent thus terminating the proof. 

\subsection{Proof of Theorem~\ref{th-anchoring-trace-comparison}}

\noindent We first address the
Equations~\ref{eq-speed-comparison-for-standard-and-canonical-asymptotic-backbones}
and~\ref{eq-speed-comparison-for-anchoring-asymptotic-backbones}, when
no hypothesis regarding the monotony of the asymptotic backbone
$\{\alpha_i\}_{i \in \mathbb{N}}$ associated to the reference is
established. Given that the former has already been
stated~\citep{VilaresDarribaRibadas16}, we focus on the second one,
starting from its left-most inequality. \\

\noindent Let us assume $\jmath > \imath$, $\eta > \beta$ and $i \geq
\wp + \jmath$, then
$\hat\alpha_{i}^{\beta,\imath} := \lim \limits_{x \rightarrow \infty}
\hat{\mathcal A}_{i}^{\stackrel{\beta,\imath}{\pi}}[{\mathcal
    D}^{\mathcal {K}}_{\sigma}](x)$, with $\hat{\mathcal
  A}_{i}^{\stackrel{\beta,\imath}{\pi}}[{\mathcal D}^{\mathcal
    {K}}_{\sigma}]$ a fitting curve for the values
\begin{equation}
\{[x_j, {\mathcal A}_{\dinfty{}}[{\mathcal D}^{\mathcal
      {K}}_{\sigma}](x_j)]\}_{j=1}^{i}
\; \cup \; \{[\infty, \hat{\mathcal A}_{i}^{\beta,\imath}(\infty)]\}
= \{[x_j, {\mathcal A}_{\dinfty{}}[{\mathcal D}^{\mathcal
      {K}}_{\sigma}](x_j)]\}_{j=1}^{i} \; \cup \;
\{[\infty, \hat\alpha_{\imath}^\beta]\}
\end{equation}
\noindent with $x_j := \absd{\mathcal D_j}$. Similarly,
$\hat\alpha_{i}^{\beta,\jmath} := \lim \limits_{x \rightarrow \infty}
\hat{\mathcal A}_{i}^{\stackrel{\beta,\jmath}{\pi}}[{\mathcal
    D}^{\mathcal {K}}_{\sigma}](x)$, with $\hat{\mathcal
  A}_{i}^{\stackrel{\beta,\jmath}{\pi}}[{\mathcal D}^{\mathcal
    {K}}_{\sigma}]$ a curve fitting 
\begin{equation}
\{[x_j, {\mathcal A}_{\dinfty{}}[{\mathcal D}](x_j)], \; x_j := \absd{\mathcal D_j}\}_{j=1}^{i}
\; \cup \; \{[\infty, \hat\alpha_{\jmath}^{\beta}]\}
\end{equation}
\noindent where, as by Theorem~\ref{th-fixed-anchoring-trace} the
sequence $\{\hat\alpha_i^{\beta}\}_{i >
  \omega}$ converges decreasingly to
$\alpha_\infty = \alpha_{\dinfty{}} \geq 0$, we conclude that
$\hat\alpha_\imath^\beta > \hat\alpha_\jmath^\beta$. As
Theorem~\ref{th-fixed-anchoring-trace-with-look-ahead} guarantees
that $\{\hat\alpha_i^{\beta,\imath}\}_{i >
  \omega}$ and
$\{\hat\alpha_i^{\beta,\jmath}\}_{i > \omega}$
also converge decreasingly to $\alpha_\infty = \alpha_{\dinfty{}} \geq
0$, and $\wp \geq
\omega$, we can conclude that
$\abs{\hat\alpha_i^{\beta,\jmath} - \alpha_{\dinfty{}}} \leq
\abs{\hat\alpha_i^{\beta,\imath} - \alpha_{\dinfty{}}}$. We thereby
demonstrate the inequality in question. \\

\noindent On the other hand, $\hat\alpha_{i}^{\eta,\imath} := \lim
\limits_{x \rightarrow \infty} \hat{\mathcal
  A}_{i}^{\stackrel{\eta,\imath}{\pi}}[{\mathcal D}^{\mathcal
    {K}}_{\sigma}](x)$, with $\hat{\mathcal
  A}_{i}^{\stackrel{\eta,\imath}{\pi}}[{\mathcal D}^{\mathcal {K}}_{\sigma}]$
a fitting curve for the values
\begin{equation}
\{[x_j, {\mathcal A}_{\dinfty{}}[{\mathcal D}](x_j)]\}_{j=1}^{i}
\; \cup \; \{[\infty, \hat{\mathcal A}_{i}^{\eta,\imath}(\infty)]\} = 
\{[x_j, {\mathcal A}_{\dinfty{}}[{\mathcal D}](x_j)]\}_{j=1}^{i} \; \cup \;
\{[\infty, \hat\alpha_{\imath}^\eta]\}
\end{equation}
\noindent where $x_j := \absd{\mathcal D_j}$ and $\hat\alpha_{i}^{\eta} := \lim
\limits_{x \rightarrow \infty} \hat{\mathcal
  A}_{i}^{\stackrel{\eta}{\pi}}[{\mathcal D}^{\mathcal
    {K}}_{\sigma}](x)$, with $\hat{\mathcal
  A}_{i}^{\stackrel{\eta}{\pi}}[{\mathcal D}^{\mathcal {K}}_{\sigma}]$
a fitting curve for the values
\begin{equation}
\{[x_j, {\mathcal A}_{\dinfty{}}[{\mathcal D}](x_j)]\}_{j=1}^{i}
\; \cup \; \{[\infty, \hat{\mathcal
    A}_{i}^{\eta}(\infty)]\} =
\{[x_j, {\mathcal A}_{\dinfty{}}[{\mathcal D}](x_j)]\}_{j=1}^{i} \; \cup \;
\{[\infty, \eta]\}, \; x_j := \absd{\mathcal D_j}
\end{equation}
\noindent Similarly, $\hat\alpha_{i}^{\beta} := \lim
\limits_{x \rightarrow \infty} \hat{\mathcal
  A}_{i}^{\stackrel{\beta}{\pi}}[{\mathcal D}^{\mathcal
    {K}}_{\sigma}](x)$, with $\hat{\mathcal
  A}_{i}^{\stackrel{\beta}{\pi}}[{\mathcal D}^{\mathcal {K}}_{\sigma}]$
a fitting curve for the values
\begin{equation}
\{[x_j, {\mathcal A}_{\dinfty{}}[{\mathcal D}](x_j)], \; x_j := \absd{\mathcal
  D_j}\}_{j=1}^{i} \; \cup \;
\{[\infty, \beta]\}
\end{equation}
\noindent Since $\eta > \beta$, we then have that
$\hat\alpha_\imath^\eta \leq \hat\alpha_\imath^\beta$, and therefore
$\hat\alpha_\imath^{\eta,\imath} \leq
\hat\alpha_\imath^{\beta,\imath}$. Accordingly, we also conclude
that $\abs{\hat\alpha_i^{\beta,\imath} - \alpha_{\dinfty{}}} \leq
\abs{\hat\alpha_i^{\eta,\imath} - \alpha_{\dinfty{}}}$, because
$\{\hat\alpha_i^{\eta,\imath}\}_{i > \omega}$
and $\{\hat\alpha_i^{\beta,\imath}\}_{i >
  \omega}$ are both positive definite and
$\wp \geq \omega$. \\

\noindent Finally, we demonstrate the 
Equation~\ref{eq-speed-comparison-for-decreasing-asymptotic-backbones},
when the asymptotic backbone $\{\alpha_i\}_{i \in \mathbb{N}}$ of the
reference is decreasing. On the basis of the result previously stated
for the generic case, and taking into account that by definition
$\omega \leq \wp \leq 100 \leq \beta$ and $\hat{\mathcal
A}^{\stackrel{\beta,0}{\pi}}[{\mathcal D}^{\mathcal {K}}_{\sigma}]
= \hat{\mathcal A}^{\stackrel{\beta}{\pi}}[{\mathcal D}^{\mathcal
{K}}_{\sigma}]$, it is sufficient to establish that
\begin{equation}
\abs{\hat\alpha_i - \alpha_{\dinfty{}}} \leq \abs{\hat\alpha_i^\beta -
  \alpha_{\dinfty{}}}, \; \forall i > \wp
\end{equation}
\noindent where $\hat\alpha_{i} := \lim \limits_{x \rightarrow \infty}
\hat{\mathcal A}_{i}^{\pi}[{\mathcal D}^{\mathcal
    {K}}_{\sigma}](x)$, with $\hat{\mathcal
  A}_{i}^{\pi}[{\mathcal D}^{\mathcal {K}}_{\sigma}]$ a fitting curve for the values
\begin{equation}
\{[x_j, {\mathcal A}_{\dinfty{}}[{\mathcal D}](x_j)]\}_{j=1}^{i}
\; \cup \; \{[\infty, \hat{\mathcal A}_{i}(\infty)]\}
=
\{[x_j, {\mathcal A}_{\dinfty{}}[{\mathcal D}](x_j)]\}_{j=1}^{i} \; \cup \;
\{[\infty, \hat\alpha_{i-1}]\}
\end{equation}
\noindent with $x_j := \absd{\mathcal D_j}$. Furthermore,
$\hat\alpha_{i}^\beta := \lim \limits_{x \rightarrow \infty}
\hat{\mathcal A}_{i}^{\stackrel{\beta}{\pi}}[{\mathcal D}^{\mathcal
    {K}}_{\sigma}](x)$, with $\hat{\mathcal
  A}_{i}^{\stackrel{\beta}{\pi}}[{\mathcal D}^{\mathcal
    {K}}_{\sigma}]$ a curve fitting 
\begin{equation}
\{[x_j, {\mathcal A}_{\dinfty{}}[{\mathcal D}](x_j)]\}_{j=1}^{i}
\; \cup \; \{[\infty, \hat{\mathcal A}_{i}^{\beta}(\infty)]\}
=
\{[x_j, {\mathcal A}_{\dinfty{}}[{\mathcal D}](x_j)]\}_{j=1}^{i} \; \cup \;
\{[\infty, \beta]\}, \; x_j := \absd{\mathcal D_j}
\end{equation}
\noindent So, as $\{\hat\alpha_i\}_{i >
  \omega}$ is positive definite, for stating
the desired relation it is enough to prove that $\hat\alpha_{i-1}
\leq \beta$, which is trivial because $\hat\alpha_{i-1} \leq
100 \leq \beta$. This completes the proof.

\begin{footnotesize}

\end{footnotesize}

\end{document}